\documentclass[sigconf]{acmart}

\usepackage{multirow}

\usepackage{multicol}
\usepackage{graphicx}
\usepackage{balance}
\usepackage{color}
\usepackage{hyperref}
\hypersetup{
    colorlinks=true,  
    urlcolor=magenta,
}

\AtBeginDocument{%
  \providecommand\BibTeX{{%
    \normalfont B\kern-0.5em{\scshape i\kern-0.25em b}\kern-0.8em\TeX}}}

\copyrightyear{2023}
\acmYear{2023}
\setcopyright{acmlicensed}\acmConference[MM '23]{Proceedings of the 31st ACM International Conference on Multimedia}{October 29-November 3, 2023}{Ottawa, ON, Canada}
\acmBooktitle{Proceedings of the 31st ACM International Conference on Multimedia (MM '23), October 29-November 3, 2023, Ottawa, ON, Canada}
\acmPrice{15.00}
\acmDOI{10.1145/3581783.3611983}
\acmISBN{979-8-4007-0108-5/23/10}

\begin{document}

\title{Recurrent Multi-scale Transformer for High-Resolution Salient Object Detection}




\author{Xinhao Deng}
\affiliation{%
  \institution{Dalian University of Technology}
  \city{Dalian}
  \country{China}
  }
\email{dengxh@mail.dlut.edu.cn}
\orcid{0000-0003-1131-4710}

\author{Pingping Zhang}
\authornotemark[1]
\affiliation{%
  \institution{Dalian University of Technology}
  \city{Dalian}
  \country{China}
  }
\email{zhpp@dlut.edu.cn}
\orcid{0000-0003-1206-1444}

\author{Wei Liu}
\affiliation{%
  \institution{Shanghai Jiaotong University}
  \city{Shanghai}
  \country{China}
  }
\email{weiliucv@sjtu.edu.cn}
\orcid{0000-0001-6351-9019}

\author{Huchuan Lu}
\affiliation{%
  \institution{Dalian University of Technology}
  \city{Dalian}
  \country{China}
  }
\email{lhchuan@dlut.edu.cn}
\orcid{0000-0002-6668-9758}


\begin{abstract}
    \it Salient Object Detection (SOD) aims to identify and segment the most conspicuous objects in an image or video. As an important pre-processing step, it has many potential applications in multimedia and vision tasks. With the advance of imaging devices, SOD with high-resolution images is of great demand, recently. However, traditional SOD methods are largely limited to low-resolution images, making them difficult to adapt to the development of \textbf{High-Resolution SOD (HRSOD)}. Although some HRSOD methods emerge, there are no large enough datasets for training and evaluating. Besides, current HRSOD methods generally produce incomplete object regions and irregular object boundaries. To address above issues, in this work, we first propose a new \textbf{HRS10K} dataset, which contains 10,500 high-quality annotated images at 2K-8K resolution. As far as we know, it is the largest dataset for the HRSOD task, which will significantly help future works in training and evaluating models. Furthermore, to improve the HRSOD performance, we propose a novel \textbf{Recurrent Multi-scale Transformer (RMFormer)}, which recurrently utilizes shared Transformers and multi-scale refinement architectures. Thus, high-resolution saliency maps can be generated with the guidance of lower-resolution predictions. Extensive experiments on both high-resolution and low-resolution benchmarks show the effectiveness and superiority of the proposed framework. The source code and dataset are released at: \href{https://github.com/DrowsyMon/RMFormer}{\textcolor{magenta}{https://github.com/DrowsyMon/RMFormer}}.

\end{abstract}





\maketitle
\vspace{-2mm}
\section{Introduction}
As an important pre-processing step, Salient Object Detection (SOD) aims to identify and segment the most conspicuous objects in an image or video \cite{borji2019salient, wang2021salient}. It has shown great potential for many computer vision and multimedia tasks, such as object recognition \cite{ren2013region}, image segmentation \cite{wang2017}, visual tracking \cite{zhang2020non}. Recently, with the advance of imaging devices, SOD with high-resolution images \cite{zeng2019towards} is of great demand and has draw the attention of many researches. However, most of existing SOD methods are designed for typical low-resolution images (e.g., 256 $\times$ 256, 384 $\times$ 384). When performing on high-resolution images, these methods usually produce blurry object boundaries and miss key object regions. As a consequence, they are not suitable to the High-Resolution Salient Object Detection (HRSOD) task, which needs a higher boundary quality and computation efficiency.

\begin{figure}[!t]
    \centering
    \includegraphics[width=1\linewidth]{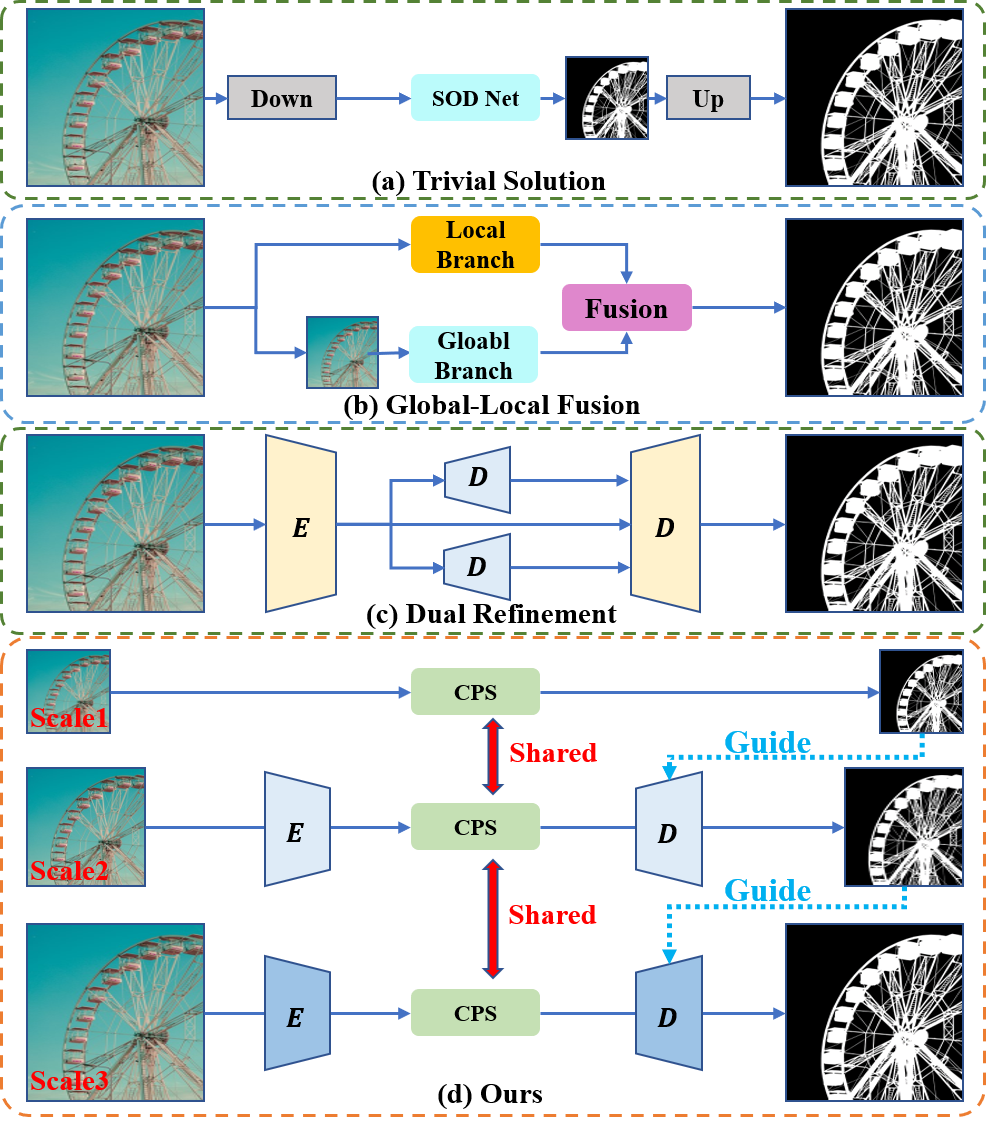}
    \vspace{-2mm}
    \caption{Comparisons of different HRSOD methods. 
    (a) directly utilizes low-resolution SOD methods with down-sampled high-resolution images and resizes the SOD prediction. (b) integrates a global branch for low-resolution input and a local branch for high-resolution one, and fuses them for the final prediction. (c) predicts the high-resolution saliency map with two complementary decoders in an end-to-end manner. (d) Our method introduces a recurrent structure with the prediction refinement in multiple stages.}
    \vspace{-2mm}
    \label{fig:motivation}
\end{figure}

Some researchers have already noticed above facts and tried to solve these challenges in HRSOD. As shown in Figure \ref{fig:motivation} (a), the trivial solution is that one can directly down-sample high-resolution images and utilize low-resolution SOD methods to generate SOD predictions. However, due to the lack of high-resolution information, there are incomplete object regions and irregular object boundaries. To address these issues, as shown in Figure \ref{fig:motivation} (b), Zeng \textit{et al.} \cite{zeng2019towards} propose the first elegant solution to HRSOD, which integrates a global branch for low-resolution images and a local branch for high-resolution patches, and fuse the two kinds of features for high-resolution saliency maps. Similarly, Tang \textit{et al.} \cite{tang2021disentangled} propose to disentangle the HRSOD task into a low-resolution saliency classification and a high-resolution detail refinement. However, they generally introduce the multi-stage training and additional feature fusions. Thus, the final performance can be greatly decreased by the error accumulation. As a replacement, some researchers try to build a structure to take in high-resolution images and directly predict the high-resolution saliency maps. For example, as shown in Figure \ref{fig:motivation} (c), Zhang \textit{et al.} \cite{zhang2021looking} propose the first end-to-end learnable framework with a shared feature extractor and two effective refinement heads. In addition, Kim \textit{et al.} \cite{kim2022revisiting} utilize an image pyramid structure and ensemble multiple SOD results with image blending for HRSOD. These methods show better results than previous ones. However, they encounter the complexity of designing a framework that can effectively extract contextual and detail features.

Meanwhile, the lack of large-scale datasets also limits the development of HRSOD methods. In fact, Zeng \textit{et al.} \cite{zeng2019towards} introduce the first HRSOD dataset. However, it only contains 1,610 training images and 400 test images. Then, Xie \textit{et al.} \cite{xie2022pyramid} propose the UHRSD dataset, which contains 4,932 training images and 988 test images. However, both of them have a small amount of images, given the complexity of the information contained in high-resolution images. More importantly, we find that there are some salient objects (such as roads, buildings and landforms) with significant semantic differences collected by HRSOD and UHRSD datasets. This fact leads to the performance bias for specific datasets when using different training settings \cite{xie2022pyramid, kim2022revisiting}. Thus, it needs new HRSOD datasets to realize a fair comparison for training and evaluating models.

To address above issues, in this work, we first propose a new HRSOD dataset, which contains 10,500 high-quality annotated images at 2K-8K resolution. As far as we know, it is the largest dataset for the HRSOD task, which will significantly help future works in training and evaluating models. Furthermore, to improve the HRSOD performance, we propose a novel Recurrent Multi-scale Transformer (RMFormer), which recurrently utilizes shared Transformers and multi-scale refinement architectures. As shown in Figure \ref{fig:motivation} (d), high-resolution saliency maps can be generated with the guidance of lower-resolution predictions. Besides, we design an Image Guided Encoder (IGE) and a Dual-flow Guided Decoder (DGD) to boost multi-scale representations. Then, a Pixel-wise Refiner (PR) is proposed to enhance the boundary predictions to gain better results. Extensive experiments on both high-resolution and low-resolution SOD benchmarks show the effectiveness and superiority of the proposed framework.

Our main contributions can be summarized as follows:
\begin{itemize}
    \item We provide a new large-scale HRSOD dataset. It contains 10,500 high-quality and high-resolution images with manually annotated fine-grained masks. As far as we know, it is the largest HRSOD dataset at present.
    \item We propose a novel Recurrent Multi-scale Transformer (RMFormer). It can recurrently utilize shared Transformers and multi-scale refinements to obtain better HRSOD results.
    \item Extensive experiments on both existing SOD datasets and ours demonstrate that our framework performs better than most state-of-the-art methods.
\end{itemize}
\vspace{-2mm}
\section{Related Work}
\subsection{SOD with Low-Resolution Images}
Traditional SOD methods \cite{itti1998model,cheng2014global,yan2013hierarchical} mainly use low-level features. These methods are effective in simple scenes but have unstable performance when facing challenging cases. More detail analysis on these SOD methods can be found in \cite{borji2019salient}. Recently, with the advances of deep neural networks, various SOD methods have been proposed to improve performances with low-resolution images \cite{wang2017}. Most of them have well-designed deep structures for robust feature extraction \cite{zhang2017amulet}, or using additional prior knowledge to improve the structural detail \cite{zhao2019egnet}. However, these methods are not suitable for the HRSOD task, since they cannot directly take high-resolution images as inputs, and can hardly capture fine-grained information in object details.
\subsection{SOD with High-Resolution Images}
To solve aforementioned issues, some researchers have shifted their efforts to the HRSOD task. For example, Zeng \textit{et al.} \cite{zeng2019towards} contribute the first HRSOD dataset, and they also design a baseline method by jointly learning global semantics and local details. After that, Zhang \textit{et al.} \cite{zhang2021looking} propose a shared feature extractor and two effective refinement heads. By decoupling the detail and context information, one head utilizes a global-aware feature pyramid for detail information. The other head combines convolutional blocks and upsamplings for contextual information. Similarly, Tang \textit{et al.} \cite{tang2021disentangled} disentangle the HRSOD task into a low-resolution saliency classification and a high-resolution detail refinement. By introducing uncertainty into the training process, their method can address the annotation errors in edge pixels. Furthermore, Xie \textit{et al.}\cite{xie2022pyramid} propose an one-stage learning framework to extract features from different resolution images independently and then graft the features for final predictions. Besides, they transfer the features from CNNs and Transformers \cite{dosovitskiy2020image}, which allows CNNs inherit global information. Very recently, Kim \textit{et al.} \cite{kim2022revisiting} utilize an image pyramid structure and ensemble multiple SOD results for HRSOD predictions. With a pyramid-based image blending, they can use low-resolution images to produce high-quality HRSOD predictions. In this work, we adopt a new learning paradigm, and propose a Recurrent Multi-scale Transformer (RMFormer) to obtain better HRSOD results.
\subsection{High-Resolution Salient Object Detection Datasets}
SOD datasets significantly boost the development of existing SOD methods. In fact, there are many low-resolution SOD datasets, such as ECSSD \cite{yan2013hierarchical}, DUTS \cite{wang2017}, DUT-OMRON \cite{yang2013saliency}, HKU-IS \cite{li2015visual}, PASCAL-S \cite{li2014secrets}, etc. We refer the readers to \cite{wang2017} for more details. However, low-resolution images in these datasets make current SOD methods tend to ignore the fine-grained details in high-resolution images. Besides, with the advances of imaging and display devices, such as smart phone, 4K/8K television, there are great demands for high-resolution image processing, including HRSOD. To support these demands, Zeng \textit{et al.} \cite{zeng2019towards} propose the first HRSOD dataset, which contains 1,610 training images and 400 test images. Recently, Xie \textit{et al.} \cite{xie2022pyramid} propose the UHRSD dataset, which contains 4,932 images for training and 988 images for testing. However, when compared with the widely-used low-resolution SOD dataset for model training \cite{wang2017}, both of these two datasets are too small. In fact, even when mixing the training images (6,542 images), it is still not sufficient. More importantly, we find that there is a large performance bias when training models with different HRSOD datasets. The main reason is that the selection methods of salient objects in these datasets are very different. Thus, there is an urgent need for new HRSOD datasets to realize a fair comparison for training and evaluating models.
\section{Our HRS10K Dataset}
To address above issues and further promote the development of HRSOD, we contribute a new HRSOD dataset named HRS10K, which contains a total of 10,500 images, dividing 8,400 images for training and 2,100 images for testing. 
\subsection{Image Collection}
All the high-resolution images are collected from public websites (https://unsplash.com/ and https://pixabay.com/) with the license of all use. We first search images with MS COCO subjects \cite{lin2014microsoft}, such as person (with different sizes and poses), animal (dog, cat, horse, sheep, etc.), vehicle (airplane, cars, ship, etc.), street object (signs, building), and other objects (toys, status, food, etc.). This procedure grabs about 14,000 images. Then, we weed out the images with inconsistent salient objects. Finally, the resulting 10,500 high-quality images are manually annotated by ten well-trained persons.
\begin{figure}[!t]
    \centering
    \includegraphics[width=0.99\linewidth]{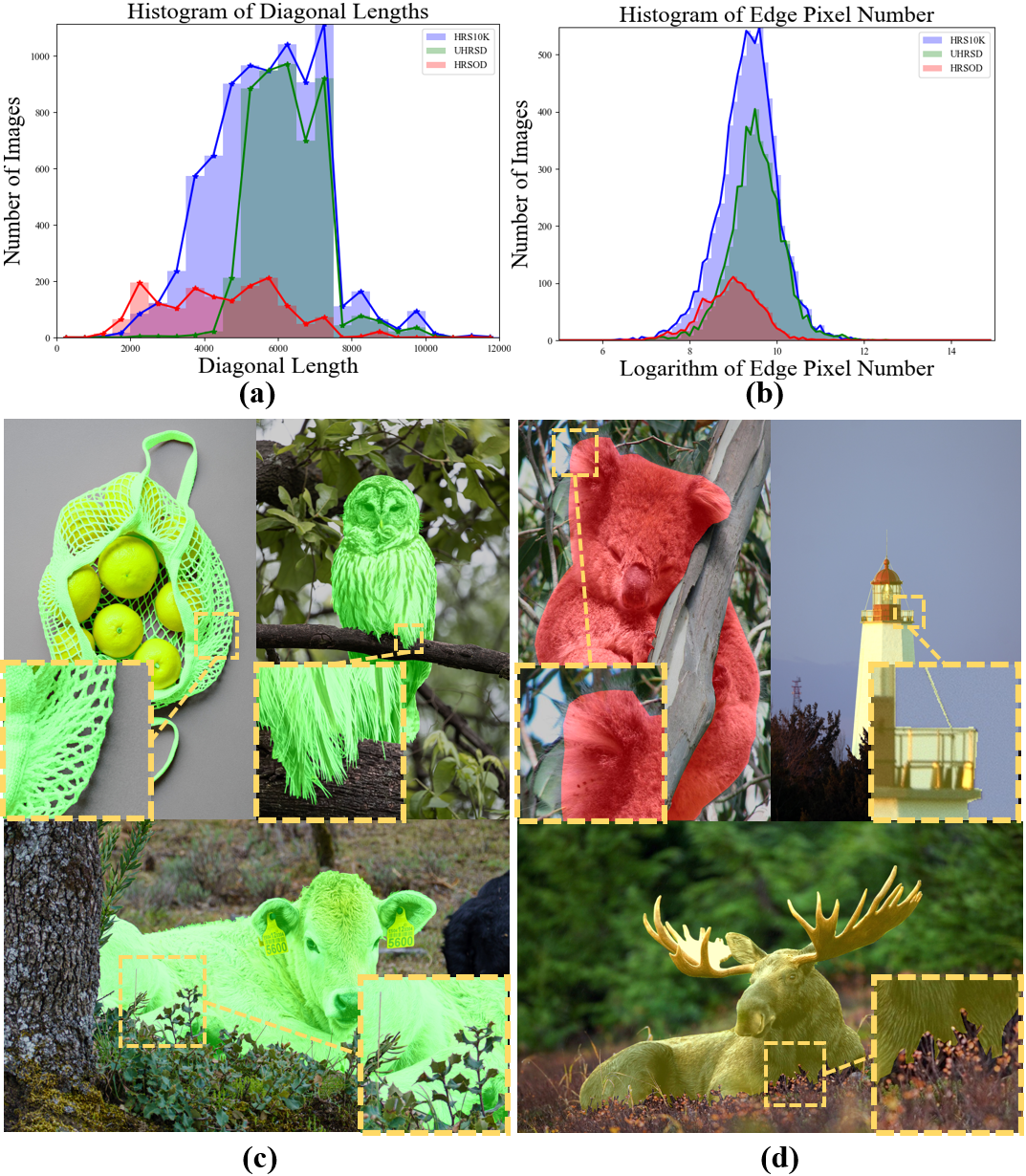}
    \vspace{-4mm}
    \caption{Statistics and visual comparisons of different high-resolution SOD datasets. (a) Histogram of images' diagonal lengths, which reflects the resolution of datasets. (b) Histogram of edge pixel numbers, which shows the complexity of annotated salient objects. (c) Samples from our HRS10K dataset, where labels are shown in green. (d) Samples from HRSOD and UHRSD, where labels are shown in red and yellow. Best view by zooming in.}
    \label{fig:dataset}
    \vspace{-4mm}
\end{figure}
\subsection{Annotation Collection}
For the annotation, we adopt a back-to-back strategy. The procedure is as follows: 1) First, each image will be annotated with five independent persons. 2) Then, two different persons will check the annotated mask for quality control. 3) The final annotated mask is generated with the maximum consistency. In our dataset, all the images are carefully annotated with precise object boundaries, as shown in Figure \ref{fig:dataset}. To avoid the redundant annotation, we utilize a pre-trained MobileNet \cite{howard2017mobilenets} to generate a feature vector for each image. Then, we calculate the feature similarity to filter out extremely similar images in our dataset.
\begin{figure*}[!t]
    \centering
    \includegraphics[width=1\linewidth]{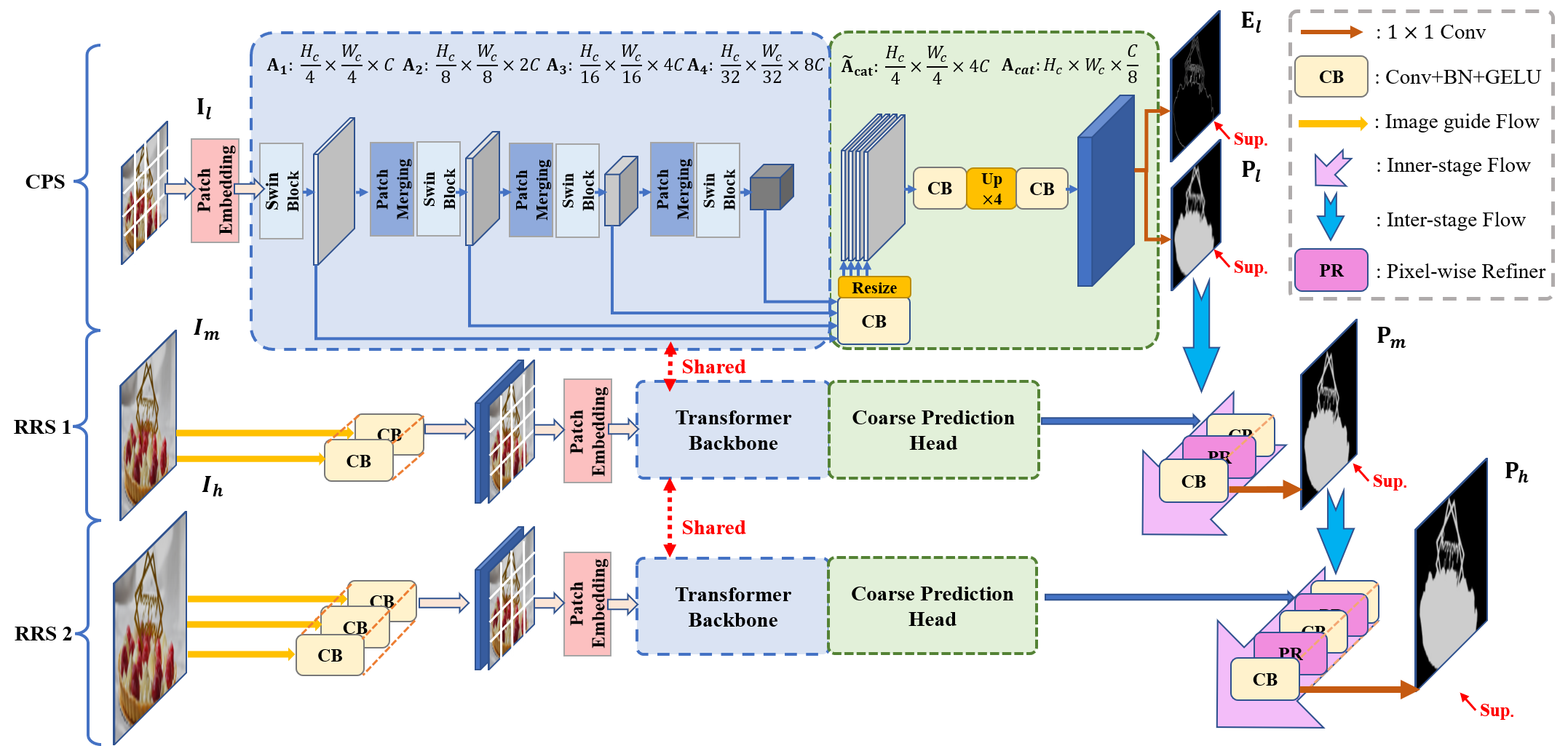}
    \caption{The overall architecture of our proposed RMFormer. A total of three prediction stages are designed. A Coarse Prediction Stage (CPS) takes the resized low-resolution image and generates a coarse prediction. The later two Recurrent Refinement Stages (RRS) take higher resolution inputs, using an dual-flow as refinement guidance. All the stages share the same Transformer structure and parameters.}
    \label{fig:main}
\end{figure*}
\subsection{Dataset Analysis}
Our proposed dataset not only has substantial images, but also has high-quality masks. Figure \ref{fig:dataset} (a) shows the statistics of images’ diagonal lengths. One can see that our dataset has a broad range with a minimal side is more than 1,600, which is larger than the previous HRSOD and UHRSD datasets. Besides, as shown in Figure \ref{fig:dataset} (b), our dataset has more edge pixels in images. This means that our dataset has more challenging salient objects with complex boundaries. Figure \ref{fig:dataset} (c) and (d) show the label comparison of our HRS10K (green), HRSOD (red) and UHRSD (yellow). It can be seen that our dataset shows more accurate details (e.g, fluffy fur or edges shaded by grass). Besides, our dataset includes rarely-appeared object subjects. Thus, it can relieve the performance bias and avoid the discrepancy to previous datasets. To our best knowledge, our dataset is currently the largest HRSOD dataset.
\section{Our Method}
In this section, we present a novel Recurrent Multi-Scale Transformer (RMFormer) to effectively handle HRSOD for better performances. The whole framework consists of three stages: one Coarse Prediction Stage (CPS) and two Recurrent Refinement Stages (RRS1 and RRS2). The CPS takes a low-resolution image as input and generates a coarse saliency map. Then, two RRS recurrently utilize shared Transformers with multi-scale refinement architectures to improve the high-resolution predictions. Each RRS has an Image Guided Encoder, a shared CPS and a Dual-flow Guided Decoder to generate high-resolution predictions. A Pixel-wise Refiner (PR) is proposed for the boundary refinement. These modules are elaborated in the following sections.
\subsection{Coarse Prediction Stage}
In this stage, we adopt a common structure to generate the coarse saliency prediction with low-resolution images. Since Transformers have achieved expressive results in many vision tasks \cite{liu2021swin, carion2020end, xie2021segformer}, we choose the Swin-B \cite{liu2021swin} pretrained on ImageNet as our backbone. As shown in the CPS part of Figure \ref{fig:main}, for the high-resolution image \(\textbf{I} \in \mathbb{R}^{H \times W \times 3}\), we first resize it to \(\textbf{I}_{l} \in \mathbb{R}^{H_{l} \times W_{l} \times 3}\). Here \(H\) and \(W\) are the height and width of high-resolution image, respectively. While \(H_{l}\) and \(W_{l}\) are the low-resolution ones. Then, a patch embedding layer splits and projects the image \(\textbf{I}_{l}\) to generate the patch-wise feature \(\textbf{A}_{0} \in \mathbb{R}^{\frac{H_{l}}{4} \times \frac{W_{l}}{4} \times C}\),
 \begin{gather}
     \textbf{A}_{0} = \psi(\textbf{I}_{l}),
 \end{gather}
where \(C\) refers to the number of embedding channels and \(\psi\) refers to the patch embedding layer.

After that, the Swin Transformer generates four feature maps with different resolutions,
 \begin{gather}
     \textbf{A}_{i} = Sblk_{i}(\textbf{A}_{i-1}), \quad i=1,\ldots,4,
 \end{gather}
where \(Sblk_{i}\) donates the \(i\)-th Swin blocks with \(\textbf{A}_{i} \in \mathbb{R}^{\frac{H_{l}}{2^{1+i}} \times \frac{W_{l}}{2^{1+i}} \times 2^{1+i}C} \). 

To improve the feature representation ability, we design a coarse prediction head to generate the coarse prediction \(\textbf{P}_{l}\) and coarse edge prediction \(\textbf{E}_{l}\). We follow the decoder design of Segformer \cite{xie2021segformer}. For each feature \(\textbf{A}_{i}\), we first use a Convolutional Block (CB) to reduce the channel, then upsample them to obtain \(\tilde{\textbf{A}}_{i}\), 
 \begin{gather}
    \tilde{\textbf{A}}_{i} = Up(f(\textbf{A}_{i})), \quad i=1,\ldots,4,
 \end{gather}
 where \(Up\) donates the upsampling operation, and \(f\) donates the Convolutional Blocks (CB), which contains a \(3 \times 3\) convolutional layer, a Batch Normalization Layer (BN) and a GELU activation function \cite{hendrycks2016gaussian}. The obtained features \(\tilde{\textbf{A}}_{i} \in \mathbb{R}^{\frac{H_{l}}{4} \times \frac{W_{l}}{4} \times C}\) have the same shape of \(\textbf{A}_{0}\). Then we concatenate \(\tilde{\textbf{A}}_{i}\) to obtain the feature map \(\tilde{\textbf{A}}_{cat} \in \mathbb{R}^{\frac{H_{l}}{4} \times \frac{W_{l}}{4} \times 4C}\). The feature's channel is shrunk to \(\frac{C}{8}\) by another CB. These operations can be formulated as:
 \begin{gather}
     \tilde{\textbf{A}}_{cat} = f([\tilde{\textbf{A}}_{1}, \tilde{\textbf{A}}_{2}, \tilde{\textbf{A}}_{3}, \tilde{\textbf{A}}_{4}]),\\
     \textbf{S}_{cat} = f(\Gamma(\tilde{\textbf{A}}_{cat})),
 \end{gather}
where \(\Gamma\) is 4 \(\times\) up-sampling, and \([\cdot]\) is the concatenation operation. Then, two 1\(\times\)1 convolutional layers (\(\Phi_{1}\), \(\Phi_{2}\)) are used to obtain the coarse region prediction \(\textbf{P}_{l}\) and coarse edge prediction \(\textbf{E}_{l}\),
 \begin{gather}
     \textbf{P}_{l} = \sigma(\Phi_{1}(\textbf{A}_{cat})), \quad 
     \textbf{E}_{l} = \sigma(\Phi_{2}(\textbf{A}_{cat})),
 \end{gather}
where \(\sigma\) is the Sigmoid activation function. In this way, multi-scale features can be obtained to produce coarse predictions, which are used as strong guidance for prediction refinement. 
\begin{figure}[!t]
    \centering
    \includegraphics[width=1\linewidth]{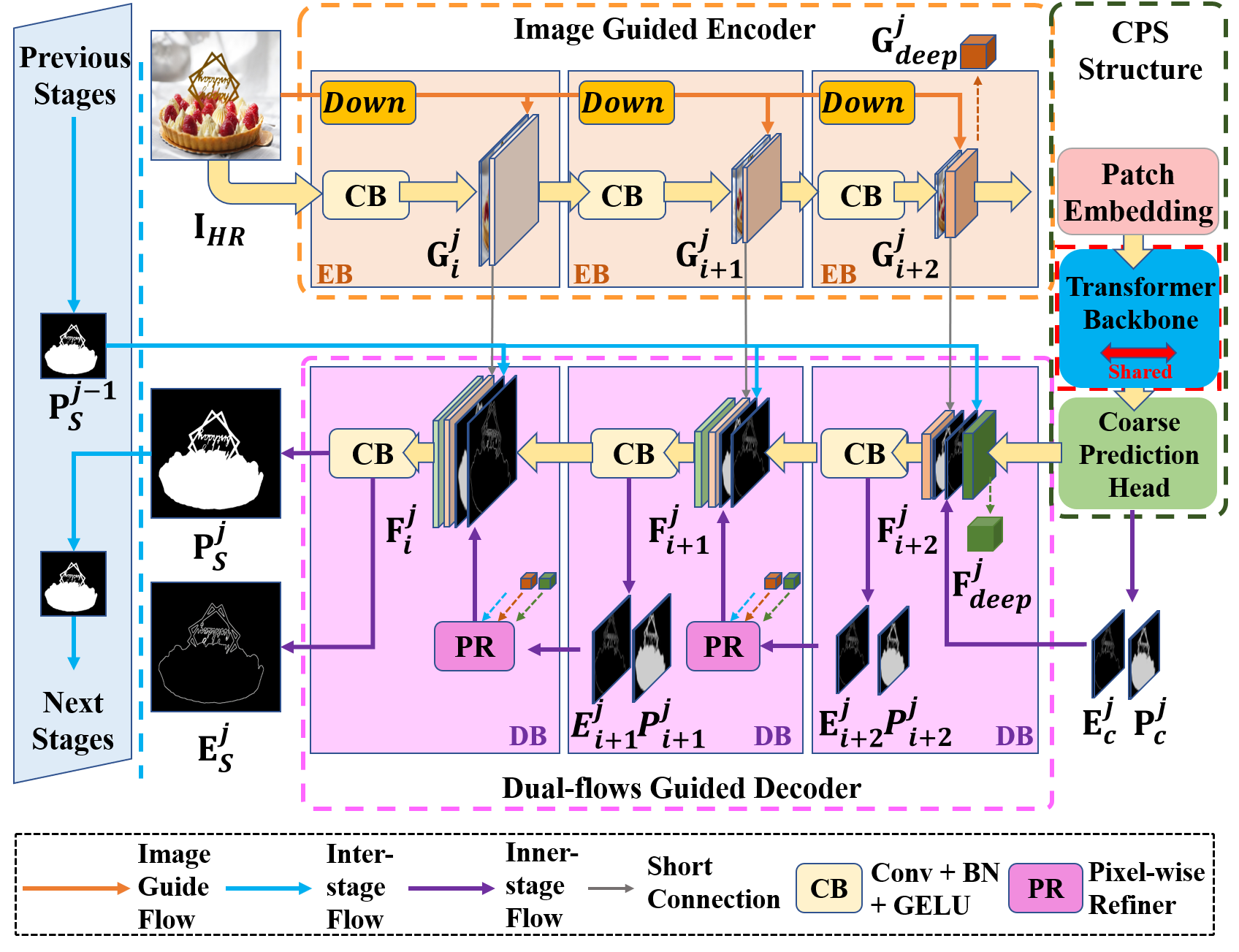}
    \caption{The detailed structure of our proposed Recurrent Refinement Stage (RRS).}
    \label{fig:submain}
    \vspace{-4mm}
\end{figure}
\subsection{Recurrent Refinement Stage}
An observation is that the coarse SOD prediction is able to locate the salient region. However, it needs to restore the missing details for HR prediction. Thus, the key of the HRSOD task is to extract and utilize the high-resolution information in the image. Therefore, we design Recurrent Refinement Stages (RSS) to refine the SOD prediction in a recurrent way. The structure of RRS is shown in Figure \ref{fig:submain}. Each RSS includes an Image Guided Encoder (IGE), a shared CPS and a Dual-flow Guided Decoder (DGD). IGE is used to capture high-resolution information. The CPS is used recurrently, to constrain the saliency information to be consistent in different stages. Then, the DGD is used for restoring the high-resolution details to obtain HR predictions. A PR module performs prediction refinement in the DGD.

\textbf{Image Guided Encoder}. As shown in the top of Figure \ref{fig:submain}, the IGE is composed of several Encoder Blocks (EB). Each EB takes the previous feature to go through a CB, then concatenates it with the resized input image,
 \begin{gather}
     \textbf{G}^{j}_{i+1} = [f_{i+1}(\textbf{G}^{j}_{i}), rz(\textbf{I}_{HR})],
 \end{gather}
where \(\textbf{G}^{j}_{i} \in \mathbb{R}^{\frac{H}{2^{(i-1)}} \times \frac{W}{2^{(i-1)}} \times C}\) indicates the encoder features for the \(i\)-th EB in the \(j\)-th RRS. Then, feature \(\textbf{G}^{j}_{i+1}\) is down-sampled to feed into the next EB. We use several EBs to obtain the feature \(\textbf{G}^{j}_{deep}\), whose shape meets the one used for CPS.

Then, the low-resolution images \(\textbf{I}_{l}\) and the last encoder features \(\textbf{G}^{j}_{deep}\) are concatenated, followed by a new patch embedding layer \(\psi^{'}\) is used to project them to patch-wise features \(\textbf{A}^{j}_{0}\) as, 
 \begin{gather}
     \textbf{A}^{j}_{0} = \psi^{'}([\textbf{I}_{l}, \textbf{G}^{j}_{deep}]),
 \end{gather}
where \(\textbf{A}^{j}_{0}\) has the same shape as \(\textbf{A}_{0}\). Then, it is used for the remaining recurrently-used CPS to generate the saliency feature \(\textbf{F}^{j}_{deep}\).

\textbf{Dual-flow Guided Decoder}. 
As shown in the bottom of Figure \ref{fig:submain}, the decoder is composed of several Decoder Blocks (DB). Each DB takes the previous DB's feature \(\textbf{F}^{j}_{i+1}\), and adds it with the corresponding encoder feature \(\textbf{G}^{j}_{i+1}\) as the basic feature. As the light-blue arrows shown in Figure \ref{fig:submain}, DB also takes the stage prediction \(\textbf{P}^{j-1}_{S}\) from the previous stage as strong prior knowledge, which can be seen as an inter-stage flow. Besides, DB takes the \(\textbf{E}^{j}_{i}\) from the last DB or CPS to gain the inner-stage flow, which is shown as purple arrows in Figure \ref{fig:submain}. Thus, operations at the \(i\)-th DB can be formulated as,
 \begin{gather}
     \textbf{F}^{j}_{i} = f_{i}([\textbf{F}^{j}_{i+1} + \textbf{G}^{j}_{i+1}, \textbf{E}^{j}_{i+1}, \textbf{P}^{j}_{S}]),\\
     \textbf{P}^{j}_{i} = \sigma(\Phi(\textbf{F}^{j}_{i})), \quad
     \textbf{E}^{j}_{i} = \sigma(\Phi(\textbf{F}^{j}_{i})).    
 \end{gather}
where \(f_{i}\) refers to the CB in the \(i\)-th DB. For the \((i-1)\)-th DB, we upsample \(\textbf{E}^{j}_{i}\) for higher resolution prediction results. Under the dual-flow guidance, the decoder can restore the HR detail and produce HR prediction.

\begin{figure}[!t]
    \centering
    \includegraphics[width=1\linewidth]{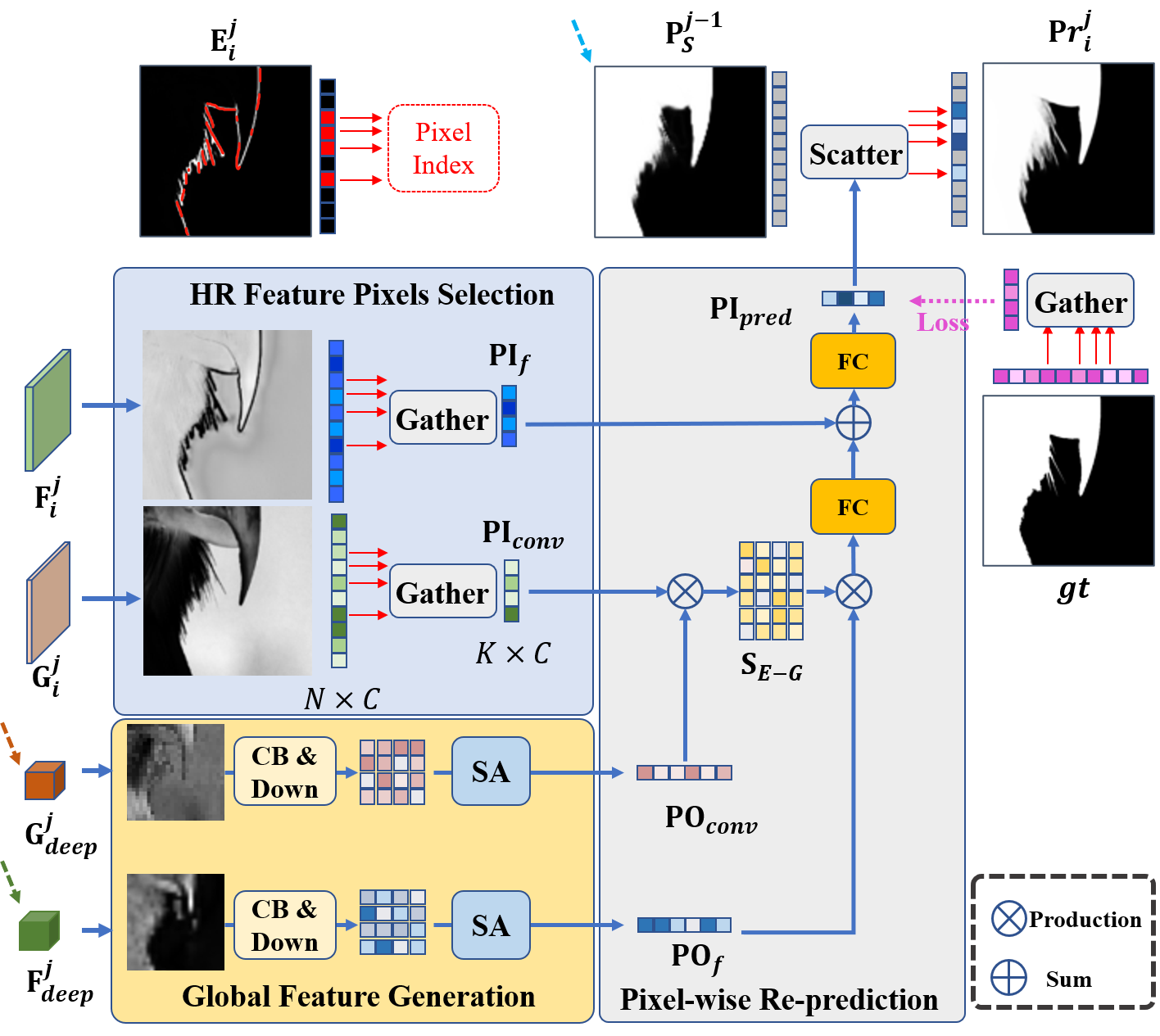}
    \caption{The detailed structure of Pixel-wise Refiner (PR).}
    \label{fig:PWR}
    \vspace{-4mm}
\end{figure}

\textbf{Pixel-wise Refiner}. 
It can be observed that if boundary areas are wrongly predicted in the low-resolution, it is hard to correct them in later high-resolution refinements. One plausible explanation could be that while deep contextual information filters out noises, it may also hinder the ability to capture high-resolution details. Thus, we propose a Pixel-wise Refiner (PR), as shown in Figure \ref{fig:PWR}. The PR consists of three units, i.e., \textit{HR Feature Pixel Selection}, \textit{Global Feature Generation} and \textit{Pixel-wise Re-prediction}.

\textit{HR Feature Pixel Selection} takes the encoder feature \(\textbf{G}^{j}_{i}\) and decoder feature \(\textbf{F}^{j}_{i}\)  as inputs (Both of them are flattened to \(\mathbb{R}^{N \times C}\)). Then, selected feature pixels \(\textbf{PI}_{conv} \in \mathbb{R}^{K \times C} \) and \(\textbf{PI}_{f}\in \mathbb{R}^{K \times C}\) are
 \begin{gather}
    \textbf{PI}_{conv} = Gather(\textbf{G}^{j}_{i}, \theta_{e}), \quad 
    \textbf{PI}_{f} = Gather(\textbf{F}^{j}_{i}, \theta_{e}),
 \end{gather}
where \(\theta_{e}\) refers to the position index of top-\(K\) pixels' in the edge prediction \(\textbf{E}^{j}_{i}\). \(Gather\) operation uses \(\theta_{e}\) to select pixels in \(\textbf{G}^{j}_{i}\) and \(\textbf{F}^{j}_{i}\) at corresponding positions.

\textit{Global Feature Generation} takes the encoder feature \(\textbf{G}^{j}_{deep}\) and initial saliency feature \(\textbf{F}^{j}_{deep}\) as inputs. Then, they are down-sampled and go through self-attention operations \cite{dosovitskiy2020image} to obtain an enhanced global features,
 \begin{gather}
     \textbf{PO}_{conv} = SA(f_{t}(\textbf{G}^{j}_{deep}, s=2)) \quad t=1,\ldots,3,\\
     \textbf{PO}_{f} = SA(f_{t}(\textbf{F}^{j}_{deep}, s=2)) \quad t=1,\ldots,3,
 \end{gather}
where \(f_{t}\) is the \(t\)-th CB with stride \(s\) and\(SA\) donates the self-attention operation. \(\textbf{PO}_{conv}\) and \(\textbf{PO}_{f}\) are enhanced global features.

\textit{Pixel-wise Re-prediction} uses selected feature pixels \(\textbf{PI}\) and global features \(\textbf{PO}\) to make re-prediction for selected pixels. As shown in the right part of Figure \ref{fig:PWR}. Firstly, we perform a matrix multiplication with \(\textbf{PI}_{conv}\) and \(\textbf{PO}_{conv}\), which is formulated as,
 \begin{gather}
     \textbf{S}_{E-G} = \sigma(W_{q}(\textbf{PI}_{conv}) \times W_{k}(\textbf{PO}_{conv})),
 \end{gather}
where \(W_{q}\) and \(W_{k}\) are linear projections. \(\textbf{S}_{E-G}\) is the edge-global similarity matrix. It can be used to project the information in global feature \(\textbf{PO}_{f}\) to each of the selected pixels. Formally,
 \begin{gather}
     \textbf{PI}_{temp} = \textbf{S}_{E-G} \times W_{v}(\textbf{PO}_{f}),
 \end{gather}
where \(W_{v}\) is a linear projection and \(\textbf{PI}_{temp}\) is the selected feature pixels which are enhanced by global information.

Finally, we can obtain the pixel-wise re-prediction \(\textbf{PI}_{pred}\) for selected pixels with two fully-connected layers. The process can be expressed as:
 \begin{gather}
     \widetilde{\textbf{PI}}_{pred} = W_{fc1}(\textbf{PI}_{temp}) + \textbf{PI}_{f},\\
     \textbf{PI}_{pred} = \sigma(W_{fc2}(\widetilde{\textbf{PI}}_{pred})),
 \end{gather}
 where \(W_{fc1}\) and \(W_{fc2}\) are fully-connected layers. Then, we can scatter \(\textbf{PI}_{pred}\) back to the prediction map \(\textbf{P}^{j-1}_{S}\) by using the same index \(\theta_{e}\) , to obtain the edge enhanced prediction map \(\textbf{Pr}^{j}_{i}\). 

We note that PR takes the feature pixels around the object boundary and uses deep features to make a re-prediction of each pixel. It is helpful for correcting the pixels which are wrongly predicted at low-resolution. Besides, the HR details can be restored more easily.
\subsection{Loss Functions}
There are two commonly-used loss functions in SOD task, namely BCE loss \cite{de2005tutorial} and IoU loss \cite{mattyus2017deeproadmapper}. Here we donate \(\mathcal{L}_{bce}\) as the BCE loss and \(\mathcal{L}_{iou}\) as the IoU loss. Each CB block in the decoder of RRS outputs a saliency prediction \(P_{i}\) and an edge prediction \(E_{i}\). Both of them are supervised by the BCE and IoU losses with resized ground truths. The loss for one RRS can be expressed as:
 \begin{gather}
     \mathcal{L}_{out} = \mathcal{L}_{bce} + \mathcal{L}_{iou},\\
     \mathcal{L}_{RRS} = \sum_{i=1} (\mathcal{L}_{out}(P_{i}, P_{gt}) + \mathcal{L}_{out}(E_{i}, E_{gt}))
 \end{gather}
where \(P_{gt}\) and \(E_{gt}\) represents the saliency and edge ground truth.

Finally, we add the loss of CPS \(\mathcal{L}_{CPS}\) and two RRS \(\mathcal{L}_{RRS}\) to get the total loss \(\mathcal{L}_{total}\), which can be represented as:
 \begin{gather}
     \mathcal{L}_{total} = \mathcal{L}_{CPS} + \mathcal{L}_{RRS1} + \mathcal{L}_{RRS2}.
 \end{gather}
\begin{table*}[h!t]
    \renewcommand{\arraystretch}{1.5}
    \setlength{\tabcolsep}{3pt}
    \centering %
    \caption{Quantitative comparisons with state-of-the-art SOD methods. The best two results are shown in \textcolor{red}{red} and \textcolor{blue}{blue}. DH: trained on DUTS-TR and HRSOD, UH: trained on HRSOD and UHRSD, KUH: trained on HRSOD, UHRSD and HRS10K.} %
    \vspace{-2mm}
    \resizebox{\linewidth}{!}{
    \begin{tabular}{l | ccccc| ccccc| ccccc | ccccc| cccc| cccc} %
        \hline
        \multicolumn{1}{l|}{\multirow{2}*{Method}} & \multicolumn{5}{c|}{DAVIS-S} 
                                                    & \multicolumn{5}{c|}{HRSOD-TE}
                                                    & \multicolumn{5}{c|}{UHRSD-TE}
                                                    & \multicolumn{5}{c|}{HRS10K-TE}
                                                    & \multicolumn{4}{c|}{DUTS-TE}
                                                    & \multicolumn{4}{c}{DUT-OMRON}\\

        \cline{2-29}
        \multicolumn{1}{l|}{} &   MAE&    \(F^{max}_{\beta}\)&   \(E_{\xi}\)&   \(S_{m}\)&  mBA&  
                                MAE&    \(F^{max}_{\beta}\)&   \(E_{\xi}\)&   \(S_{m}\)&  mBA&
                                MAE&    \(F^{max}_{\beta}\)&   \(E_{\xi}\)&   \(S_{m}\)&  mBA&
                                MAE&    \(F^{max}_{\beta}\)&   \(E_{\xi}\)&   \(S_{m}\)&  mBA&
                                MAE&    \(F^{max}_{\beta}\)&   \(E_{\xi}\)&   \(S_{m}\)&
                                MAE&    \(F^{max}_{\beta}\)&   \(E_{\xi}\)&   \(S_{m}\)\\

        \hline  

        BASNet&   
        .038&  .839&   .913&   .880&  .668&
        .038&  .861&   .919&   .891&  .663&
        .053&  .886&   .915&   .883&  .686&
        .057&  .865&   .901&   .868&  .662&
        .047&  .838&   .902&   .866&
        .056&  .779&   .871&   .836\\

        PoolNet&   
        .022&  .904&   .969&   .919&  .650&
        .041&  .867&   .924&   .895&  .646&
        .052&  .886&   .922&   .888&  .666&
        .055&  .874&   .913&   .879&  .648&
        .043&  .843&   .911&   .874&
        .057&  .764&   .866&   .831\\

        EGNet&   
        .023&  .904&   .965&   .922&  .663&
        .038&  .878&   .930&   .899&  .654&
        .049&  .897&   .927&   .895&  .680&
        .056&  .877&   .912&   .881&  .660&
        .042&  .850&   .916&   .877&
        .055&  .773&   .874&   .838\\

        SCRN&   
        .024&  .894&   .957&   .912&  .643&
        .034&  .882&   .935&   .904&  .642&
        .048&  .893&   .928&   .895&  .658&
        .053&  .882&   .918&   .886&  .639&
        .039&  .863&   .924&   .885&
        .055&  .773&   .875&   .836\\

        F3Net&   
        .020&  .909&   .968&   .913&  .666&
        .035&  .880&   .929&   .897&  .661&
        .045&  .890&   .924&   .891&  .684&
        .051&  .573&   .910&   .880&  .663&
        .035&  .871&   .926&   .886&
        .052&  .778&   .871&   .838\\

        MINet&   
        .024&  .881&   .956&   .904&  .648&
        .034&  .880&   .934&   .900&  .645&
        .043&  .892&   .930&   .894&  .660&
        .048&  .882&   .919&   .885&  .664&
        .036&  .864&   .926&   .884&
        .055&  .768&   .869&   .833\\

        LDF&   
        .019&  .904&   .964&   .921&  .667&
        .031&  .889&   .936&   .905&  .663&
        .047&  .889&   .922&   .889&  .683&
        .050&  .883&   .917&   .884&  .662&
        .033&  .876&   .929&   .892&
        .051&  .782&   .869&   .839\\

        GateNet&   
        .024&  .903&   .963&   .915&  .651&
        .032&  .893&   .943&   .911&  .650&
        .048&  .894&   .929&   .895&  .665&
        .052&  .883&   .917&   .886&  .647&
        .038&  .876&   .932&   .890&
        .054&  .782&   .877&   .840\\

        PFSNet&   
        .019&  .910&   .963&   .922&  .688&
        .032&  .895&   .940&   .907&  .674&
        .042&  .901&   .932&   .899&  .701&
        .049&  .884&   .915&   .884&  .674&
        .035&  .879&   .929&   .892&
        .054&  .790&   .877&   .842\\

        CTDNet&   
        .021&  .906&   .967&   .911&  .641&
        .033&  .886&   .940&   .898&  .637&
        .050&  .887&   .922&   .883&  .654&
        .055&  .870&   .908&   .871&  .633&
        .037&  .868&   .928&   .883&
        .052&  .784&   .876&   .837\\

        \hline  
        HRSOD-DH&   
        .026&  .899&   .955&   .897&  .623&
        .030&  .939&   .896&   .897&  .623&  
        -&  -&   -&   -&  -&  
        -&  -&   -&   -&  -&
        .050&  .835&   .885&   .894&
        .065&  .743&   .831&   .762\\
        
        DHQSD-DH&   
        .013&  .932&   .981&   .937&  .714&
        .024&  .906&   .952&   .918&  .690&  
        .039&  .911&   .935&   .900&  .716&
        .043&  .902&   .928&   .893&  .694&
        .031&  .900&   .919&   .894&
        .045&  .820&   .873&   .836\\
                
        PGNet-DH&
        .012&  .953&   .986&   .949&  .730&
        .020&  .931&   .966&   .936&  .726&  
        .036&  .915&   .939&   .910&  .746&  
        .042&  .902&   .930&   .901&  .721&
        .028&  \textcolor{blue}{.919}&   .925&   .912&
        .046&  .835&   .887&   .858\\ 

        PGNet-UH&   
        .012&  .950&   .987&   .940&  .723&
        .020&  .938&   .967&   .937&  .721&  
        .025&  .943&   .963&   .937&  .764&  
        .034&  .923&   .948&   .920&  .728&
        .036&  .857&   .930&   .873&
        .058&  .759&   .869&   .797\\ 

        PGNet-KUH&   
        {.009}&  \textcolor{blue}{.962}&   \textcolor{blue}{.991}&   \textcolor{blue}{.959}&  \textcolor{blue}{.737}&
        \textcolor{red}{.018}&  \textcolor{blue}{.946}&   \textcolor{red}{.972}&    \textcolor{blue}{.945}&  {.732}&  
        .023&  .948&   .965&   .941&  .769&  
        \textcolor{blue}{.024}&  \textcolor{blue}{.947}&   \textcolor{blue}{.965}&   \textcolor{blue}{.938}&  \textcolor{blue}{.745}&
        .038&  .871&   .897&   .859&
        .058&  .772&   .884&   .786\\

        \hline 

        Ours-DH&   
        {.009}&  {.953}&   {.987}&   {.952}&  {.717}&
        {.019}&  {.933}&   {.965}&   {.940}&  {.716}&  
        {.027}&  {.938}&   {.957}&   {.930}&  {.744}&  
        {.035}&  {.923}&   {.944}&   {.916}&  {.716}&
        \textcolor{red}{.022}&  \textcolor{red}{.921}&   \textcolor{red}{.959}&   \textcolor{red}{.951}&
        \textcolor{red}{.040}&  \textcolor{red}{.842}&   \textcolor{blue}{.914}&   \textcolor{red}{.876}\\ 

        Ours-UH&   
        \textcolor{blue}{.008}&  {.961}&   \textcolor{blue}{.991}&  {.958}&  {.736}&
        {.019}&  {.939}&   {.968}&   {.941}&  \textcolor{blue}{.736}&  
        \textcolor{blue}{.019}&  \textcolor{blue}{.955}&   \textcolor{blue}{.971}&   \textcolor{blue}{.946}&  \textcolor{blue}{.784}&  
        {.028}&  {.936}&   {.956}&   {.929}&  \textcolor{blue}{.745}&
        \textcolor{blue}{.024}&  {.917}&   \textcolor{blue}{.957}&   \textcolor{blue}{.918}&
        \textcolor{blue}{.041}&  \textcolor{red}{.844}&   \textcolor{blue}{.914}&   \textcolor{blue}{.874}\\ 

        Ours-KUH&   
        \textcolor{red}{.007}&  \textcolor{red}{.966}&   \textcolor{red}{.992}&   \textcolor{red}{.961}&  \textcolor{red}{.747}&
        \textcolor{red}{.018}&  \textcolor{red}{.948}&   \textcolor{blue}{.971}&   \textcolor{red}{.948}&  \textcolor{red}{.748}&  
        \textcolor{red}{.018}&  \textcolor{red}{.956}&   \textcolor{red}{.972}&   \textcolor{red}{.948}&  \textcolor{red}{.794}&  
        \textcolor{red}{.021}&  \textcolor{red}{.953}&   \textcolor{red}{.970}&   \textcolor{red}{.943}&  \textcolor{red}{.766}&
        {.026}&  {.915}&   {.955}&   {.917}&
        {.042}&  {.841}&   \textcolor{red}{.916}&   {.873}\\ 
        \hline 
    \end{tabular}
    }
    \label{tab:hrresult}
\end{table*}
\section{Experiments}
\subsection{Datasets and Metrics}
\textbf{Datasets}. The proposed method is trained and evaluated on our proposed new HRS10K dataset (8,400 images for training and 2,100 for testing). We also use other currently available high-resolution salient detection datasets. HRSOD is the first high-resolution dataset, containing 1,610 images for training and 400 images for testing. UHRSD is a larger dataset containing 4,932 images for training and 988 images for testing. For further evaluation, we also use the DAVIS-S dataset \cite{perazzi2016benchmark} 
which includes 92 images.

Following the work in \cite{xie2022pyramid}, we use the UH setting, which mixes the HRSOD and UHRSD to train our model for a fair comparison. We also show the results on two LR benchmarks, i.e., DUTS-TE \cite{wang2017}, DUT-OMRON \cite{yang2013saliency}. The results on ECSSD \cite{yan2013hierarchical} and HKU-IS \cite{li2015visual} can be found in the appendix.

\textbf{Evaluation Metrics}. A total of five metrics are used to evaluate the performance of all methods. Mean Absolute Error (MAE) is used to evaluate the mean average error of saliency maps, which computes the averaged L1-norm distance between predictions and ground truth maps. Max F-measure (\(F^{max}_{\beta}\)) is a weighted score of the precision and recall, where \(\beta\) is set to 0.3 as suggest in \cite{borji2015salient}. Structural similarity measure (\(S_{m}\)) and E-measure (\(E_{\xi}\)) are adopted \cite{xie2022pyramid} to better evaluate the spatial structure similarities. We also follow \cite{cheng2020cascadepsp} to evaluate the mean boundary accuracy (mBA).

\textbf{Implementation Details}. Our experiments are conducted on one RTX 3090 GPU with Pytorch \cite{paszke2017automatic}. We use the Swin-B pretrained on the ImageNet-22K as our backbone. Our method is trained end-to-end by using the Stochastic Gradient Descent (SGD). We set the learning rate to 0.001 for the backbone and 0.01 for other parts. The input image is resized to \(1536\times1536\) as HR input and \(384\times384\) for the Swin backbone. The maximum epoch number is set to 32 and the batchsize is set to 3. We adopt the widely-used data augmentation methods, i.e., random flip and random crop. Finally, we use the cosine annealing as the learning rate decay scheduler.

\subsection{Comparison with State-of-the-arts}
We compared our method with three high-resolution SOD methods (HRSOD \cite{zeng2019towards}, DHQSOD \cite{tang2021disentangled}, PGNet \cite{xie2022pyramid}) and ten low-resolution SOD methods (BASNet \cite{qin2019basnet}, PoolNet \cite{liu2019simple}, EGNet \cite{zhao2019egnet}, SCRN \cite{wu2019stacked}, F3Net \cite{wei2020f3net}, MINet \cite{pang2020multi}, LDF \cite{wei2020label}, GateNet \cite{zhao2020suppress}, PFSNet \cite{ma2021pyramidal}, CTDNet \cite{zhao2021complementary}). We use the implementation with recommended parameter settings or the provided saliency maps by corresponding authors for evaluation. The evaluation toolbox is the same as \cite{fan2018SOC}.

\textbf{Quantitative Evaluation}. We train our model in three training settings. Following the same training schedule as \cite{xie2022pyramid}, Ours-DH mixes DUTS and HRSOD. Ours-UH mixes HRSOD and UHRSD. Ours-KUH mixes HRSOD, UHRSD and HRS10K datasets, exploring the full potential of our method and dataset. As shown in Table \ref{tab:hrresult}, our method improves all five metrics on both HR and LR SOD datasets, which proves our design's effectiveness. When using the KUH training setting, the result shows a great performance boost on the mBA metric. It can be attributed that our proposed dataset has more accurate high-resolution labels. Besides, from the results of PGNet-UH and PGNet-KUH, one can see that the KUH training setting shows better generalization on HRSOD and DAVIS-S datasets. This is due to that our HRS10K dataset has taken the previous HRSOD and UHRSD subjects into consideration, and is consistent with their semantics when we annotate. The results also prove that our HRS10K dataset has reduced some discrepancies between previous HRSOD and UHRSD datasets.
\begin{figure*}[!t]
\resizebox{1\textwidth}{!}
{
\renewcommand\arraystretch{0.1}
\begin{tabular}{@{}c@{}c@{}c@{}c@{}c@{}c@{}c@{}c@{}c@{}c@{}c@{}c@{}c@{}c@{}c}

\vspace{0.5mm}
\includegraphics[width=0.1\linewidth,height=1.6cm]{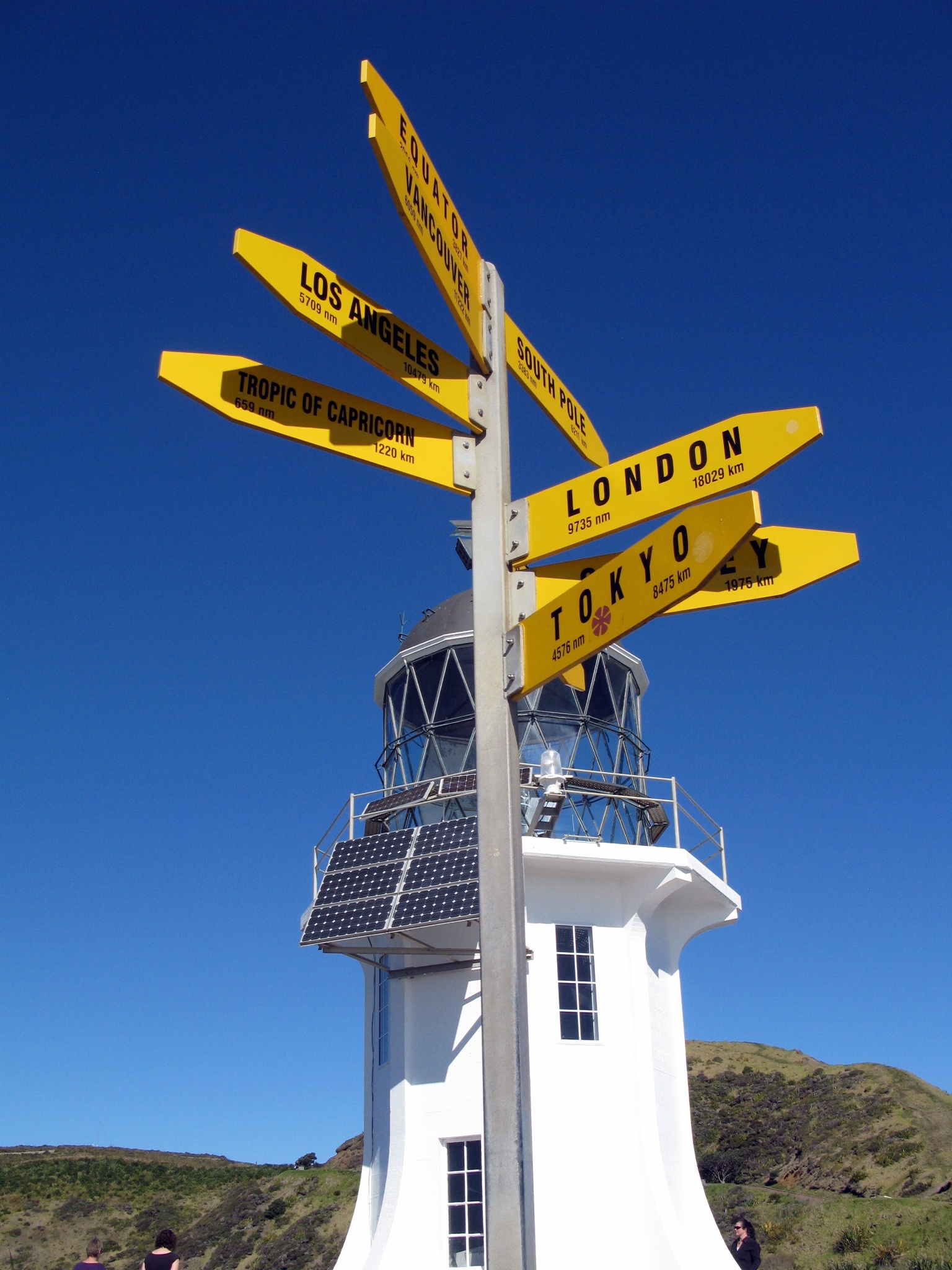}\ &
\includegraphics[width=0.1\linewidth,height=1.6cm]{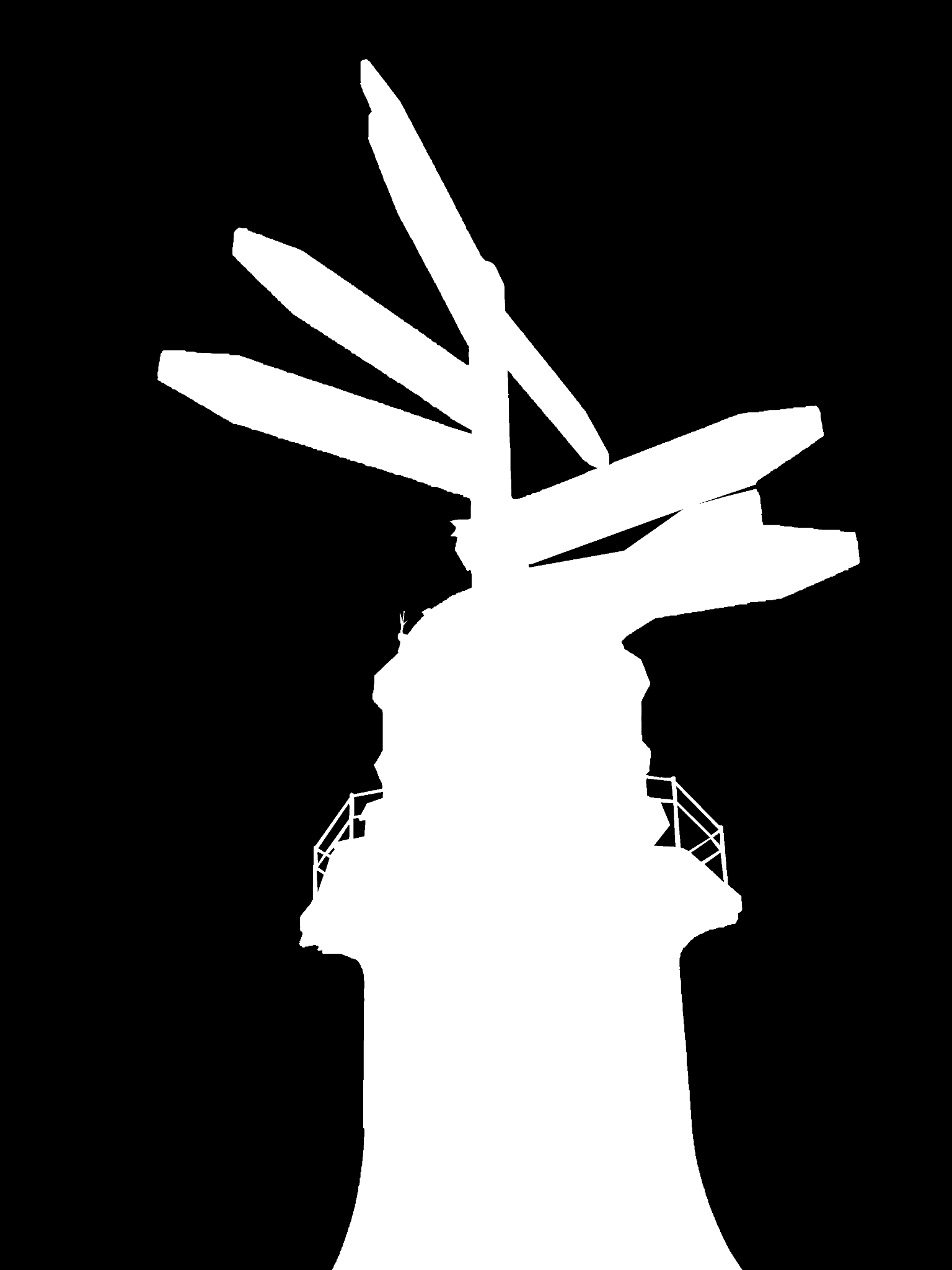}\ &
\includegraphics[width=0.1\linewidth,height=1.6cm]{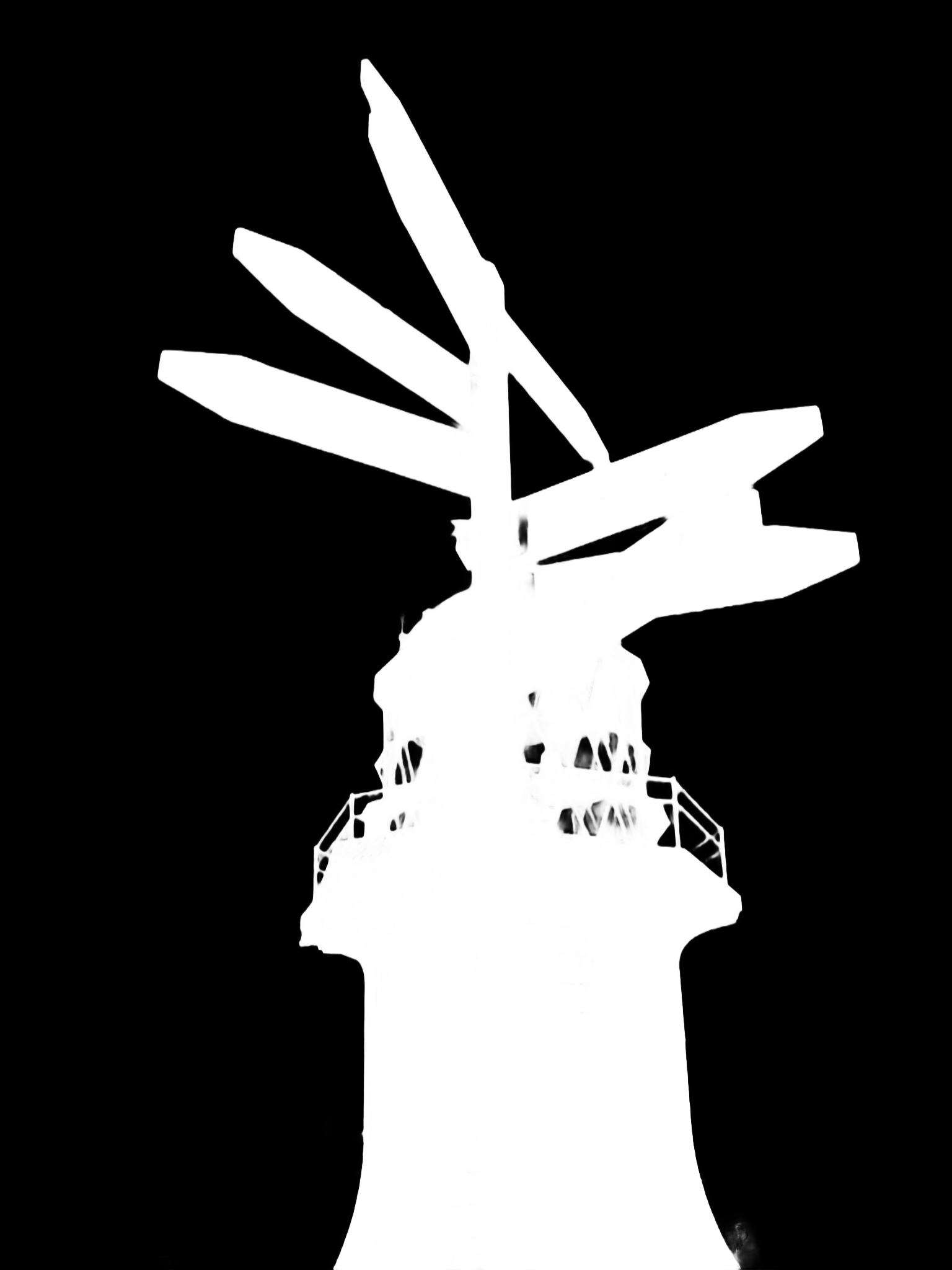}\ &
\includegraphics[width=0.1\linewidth,height=1.6cm]{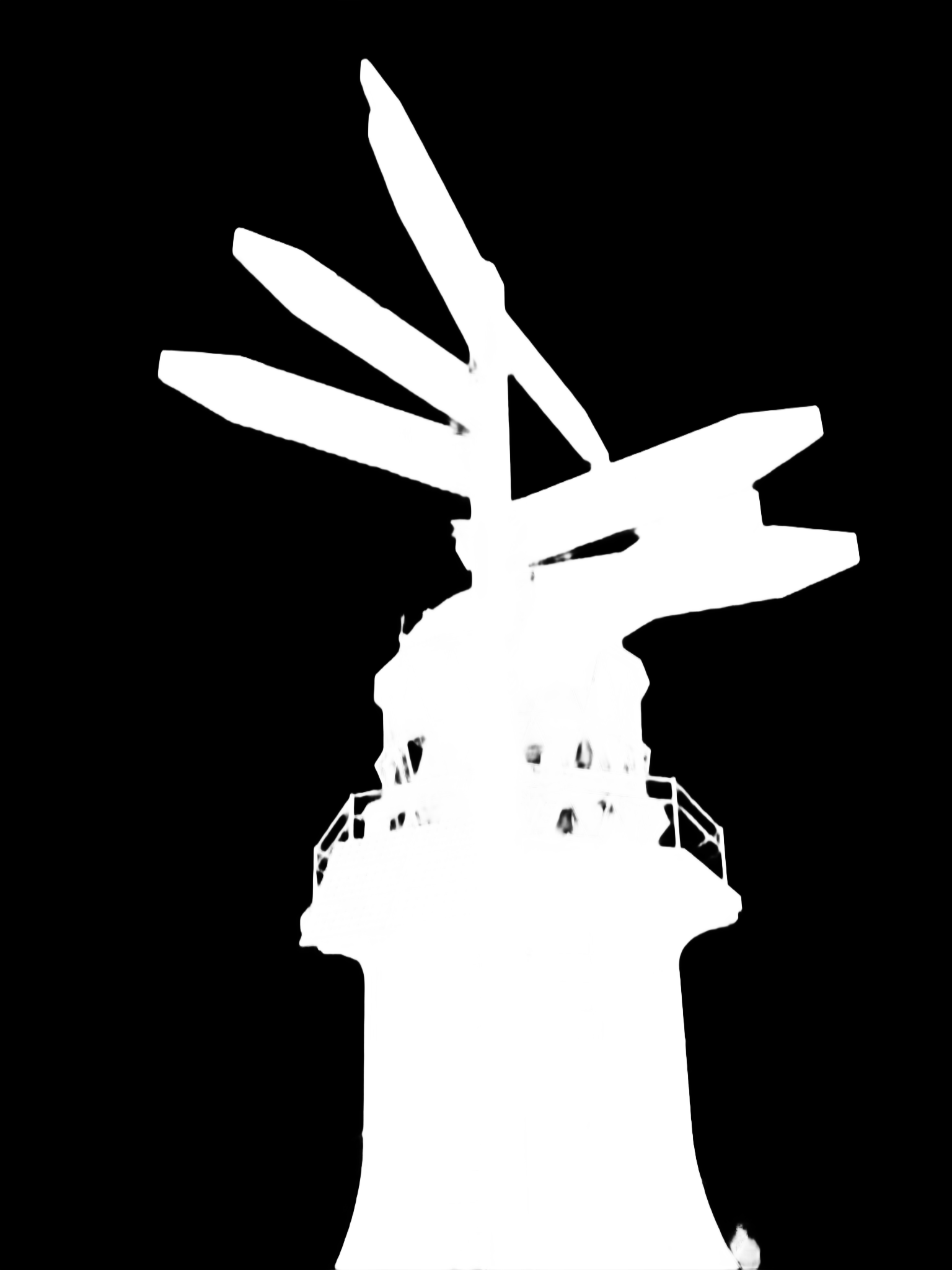}\ &
\includegraphics[width=0.1\linewidth,height=1.6cm]{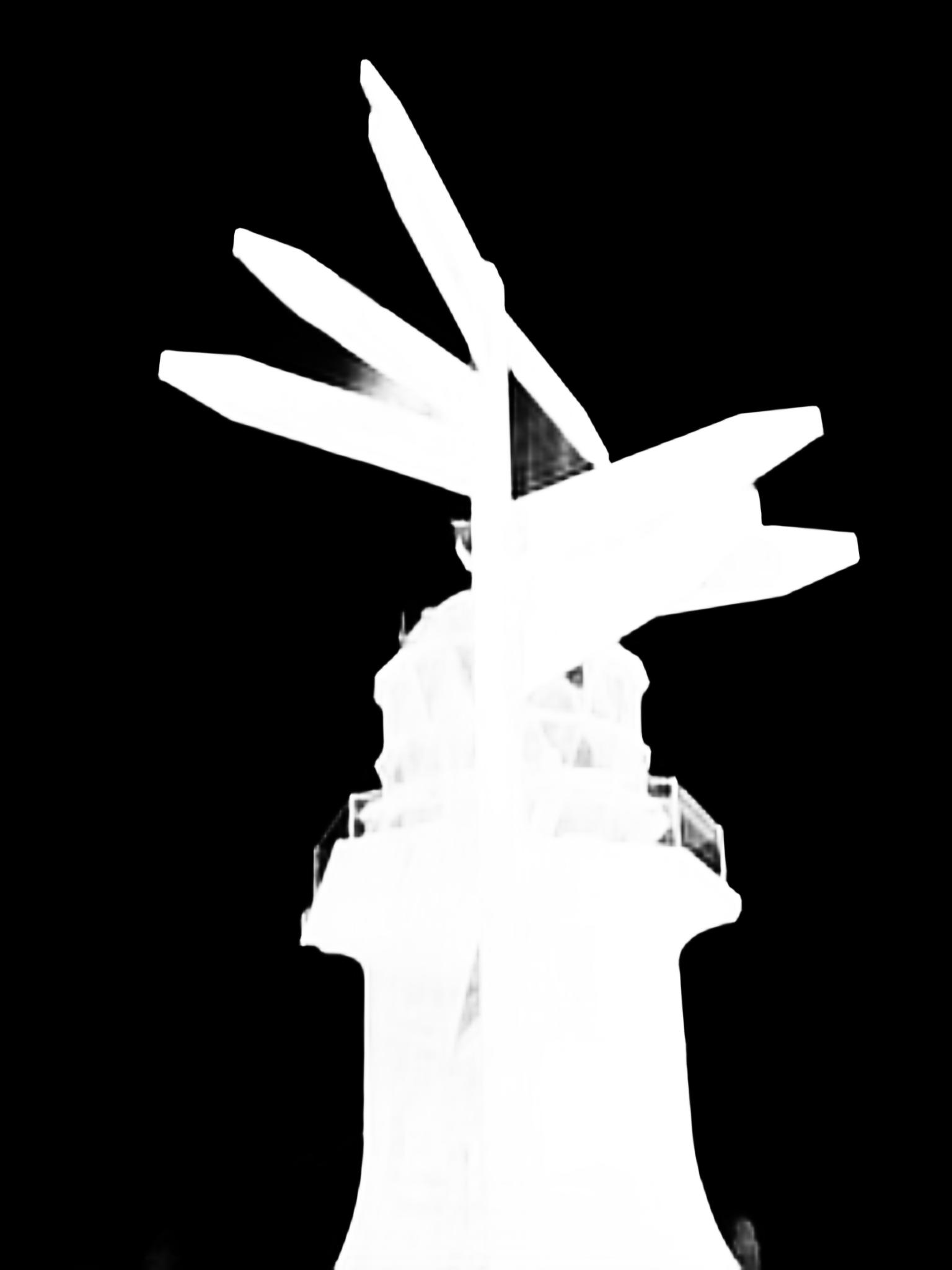}\ &
\includegraphics[width=0.1\linewidth,height=1.6cm]{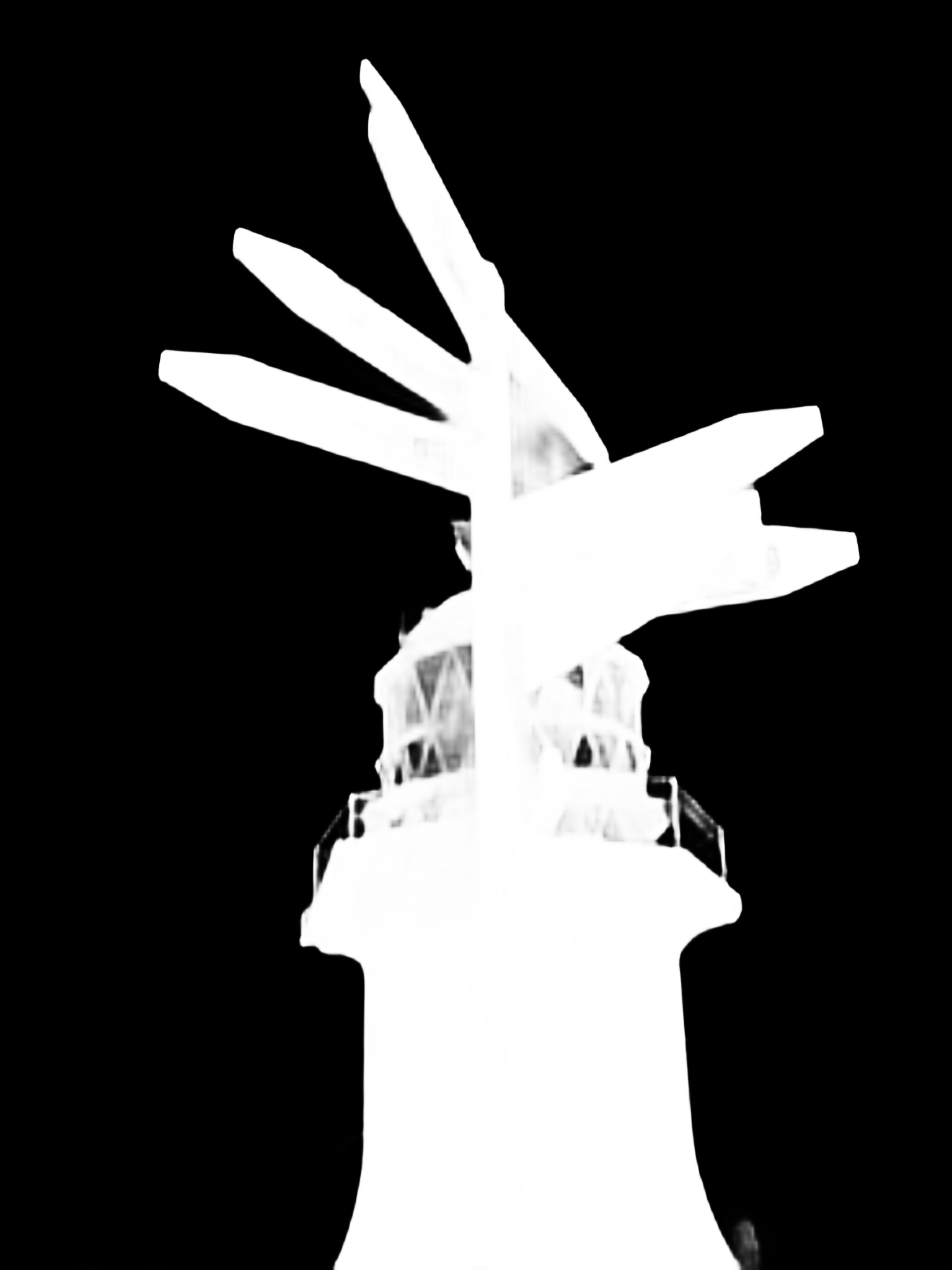}\ &
\includegraphics[width=0.1\linewidth,height=1.6cm]{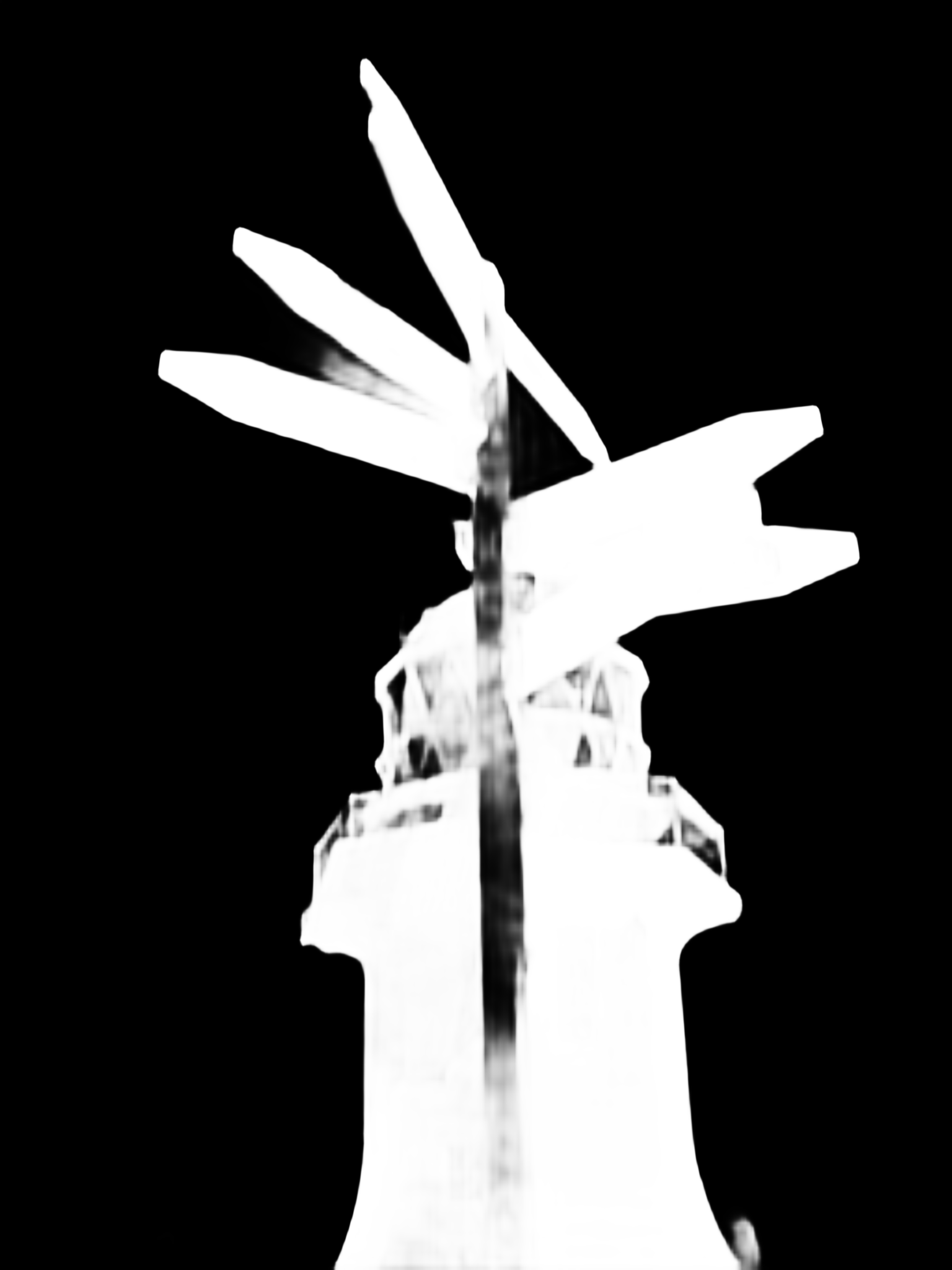}\ &
\includegraphics[width=0.1\linewidth,height=1.6cm]{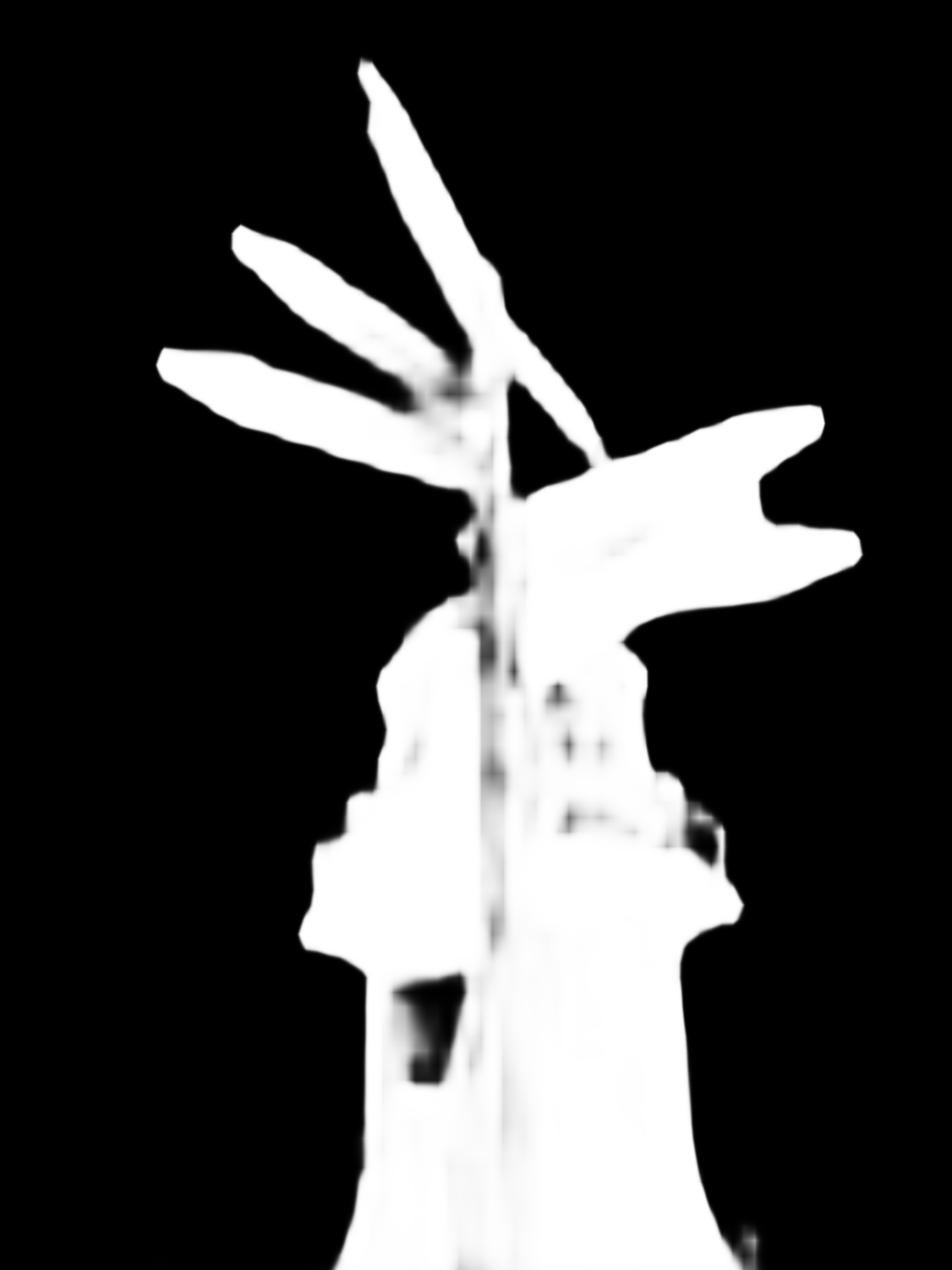}\ &
\includegraphics[width=0.1\linewidth,height=1.6cm]{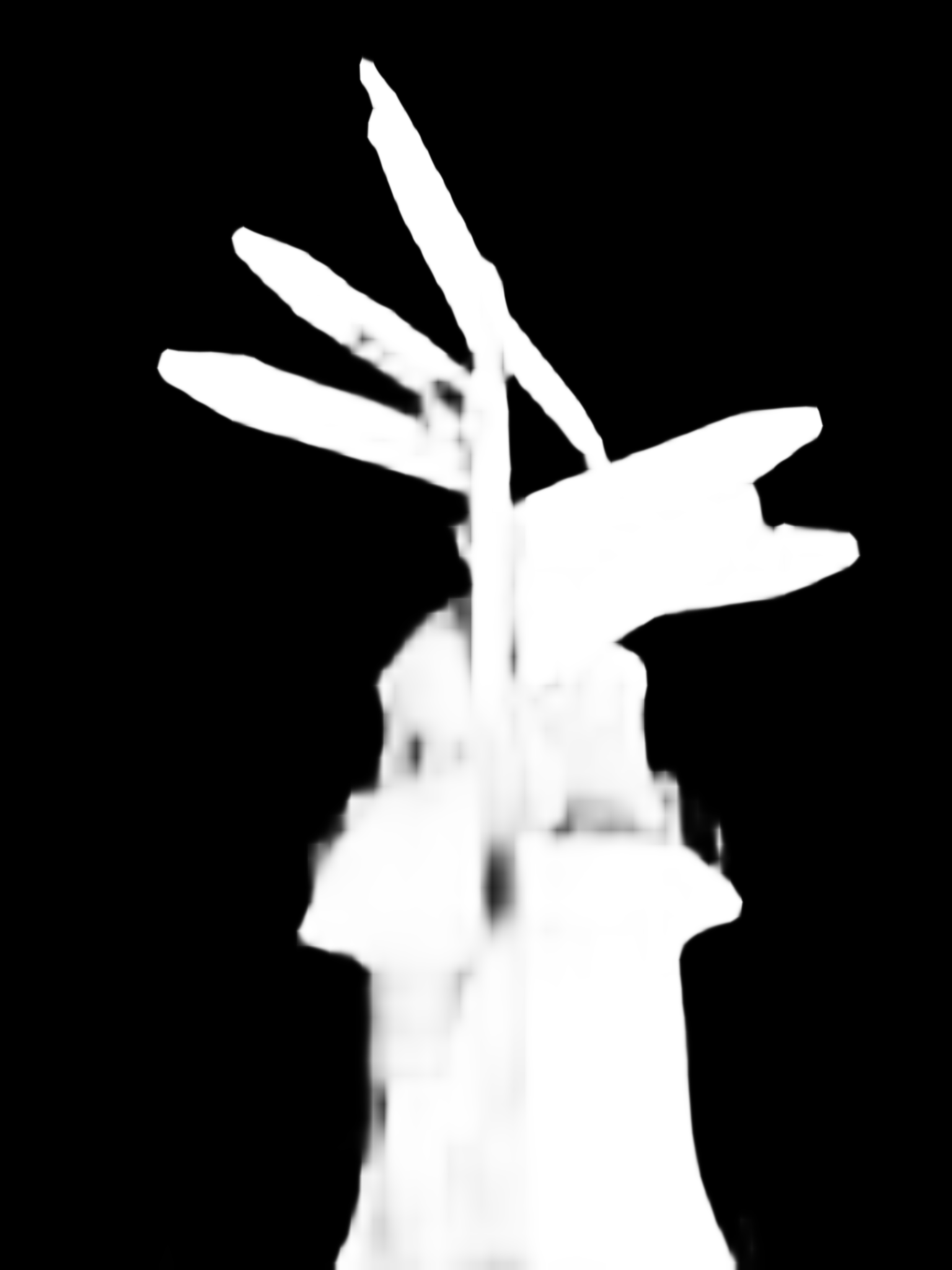}\ &
\includegraphics[width=0.1\linewidth,height=1.6cm]{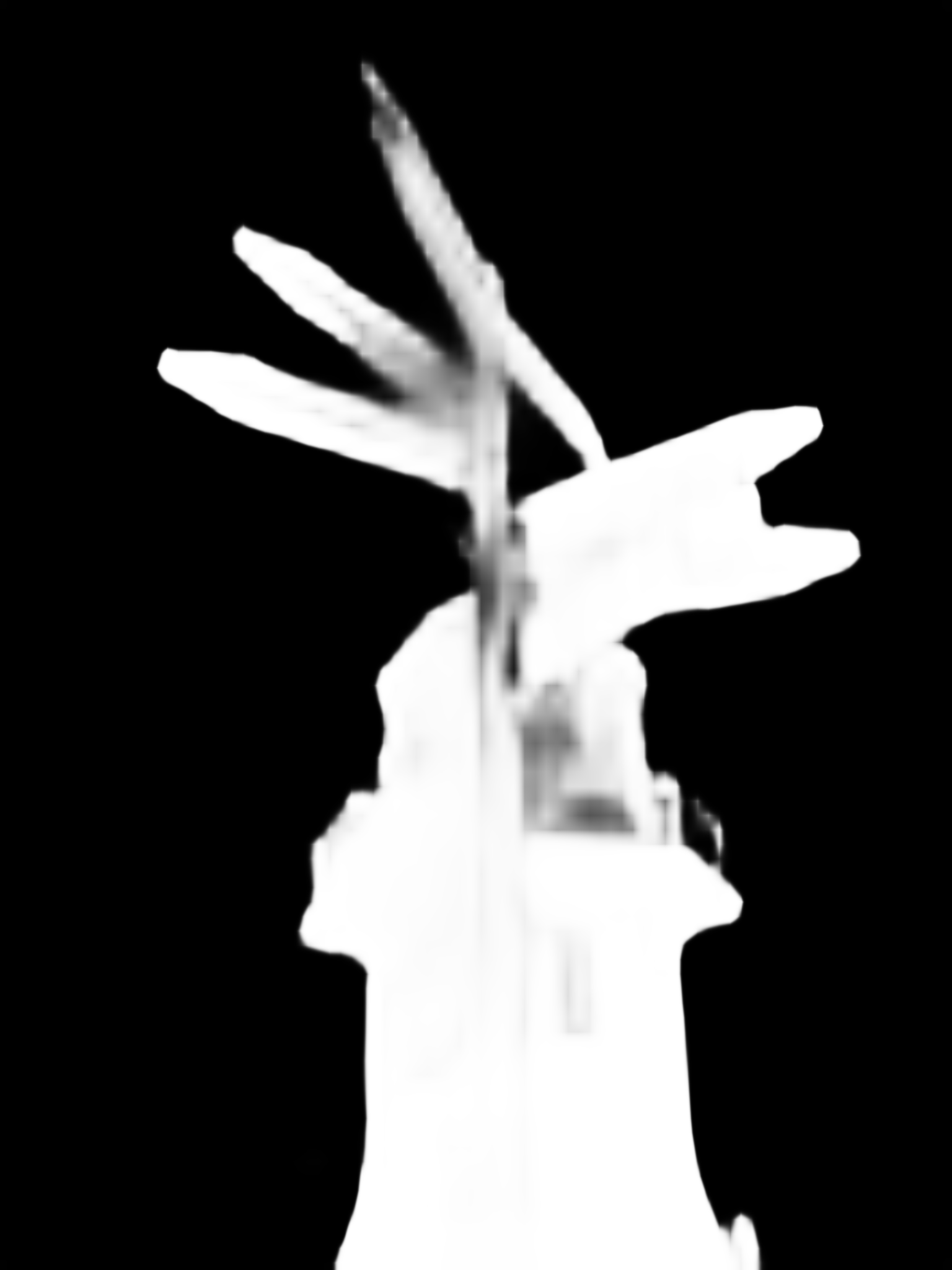}\ \\

\vspace{0.5mm}
\includegraphics[width=0.1\linewidth,height=1.7cm]{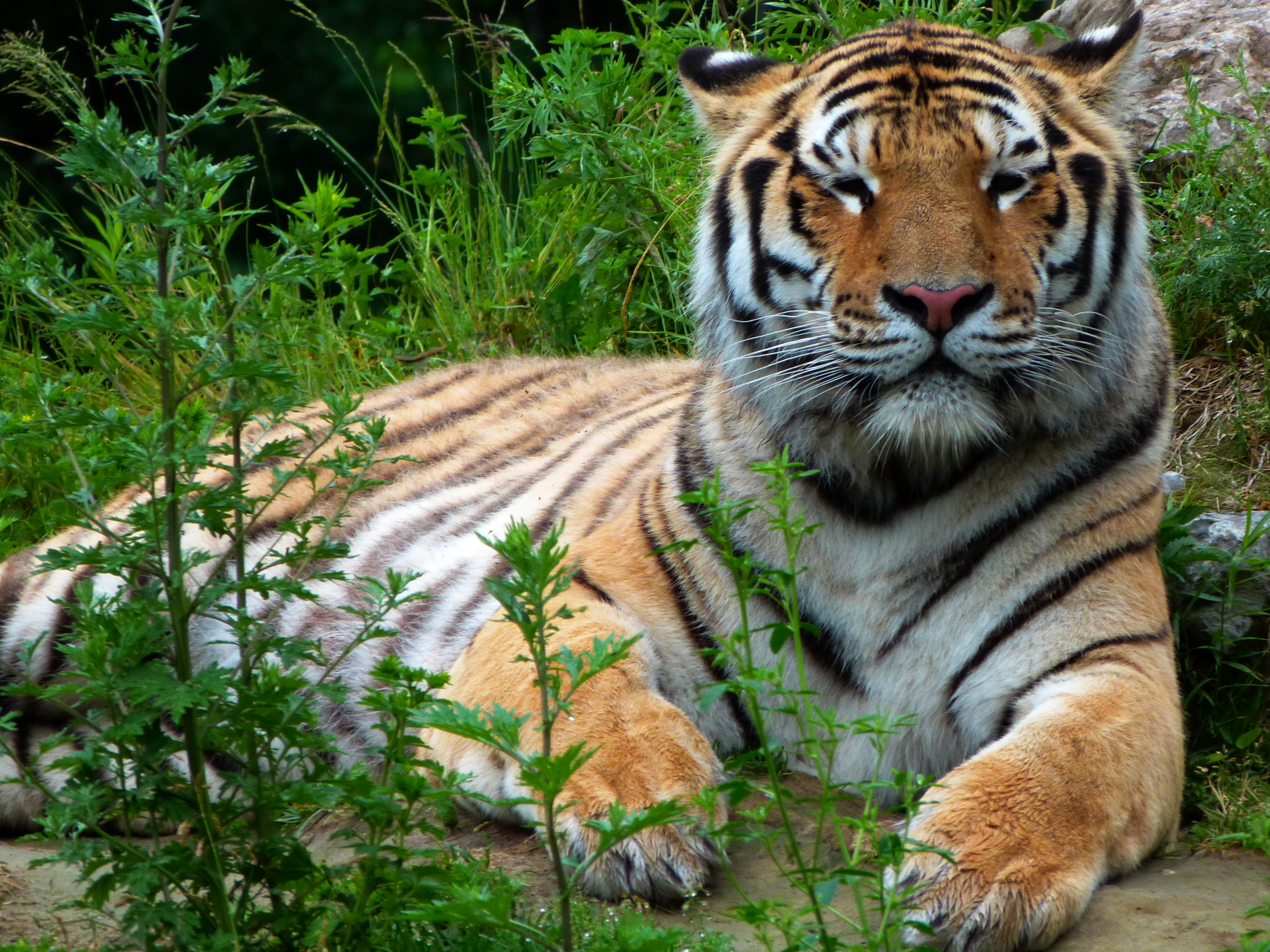}\ &
\includegraphics[width=0.1\linewidth,height=1.7cm]{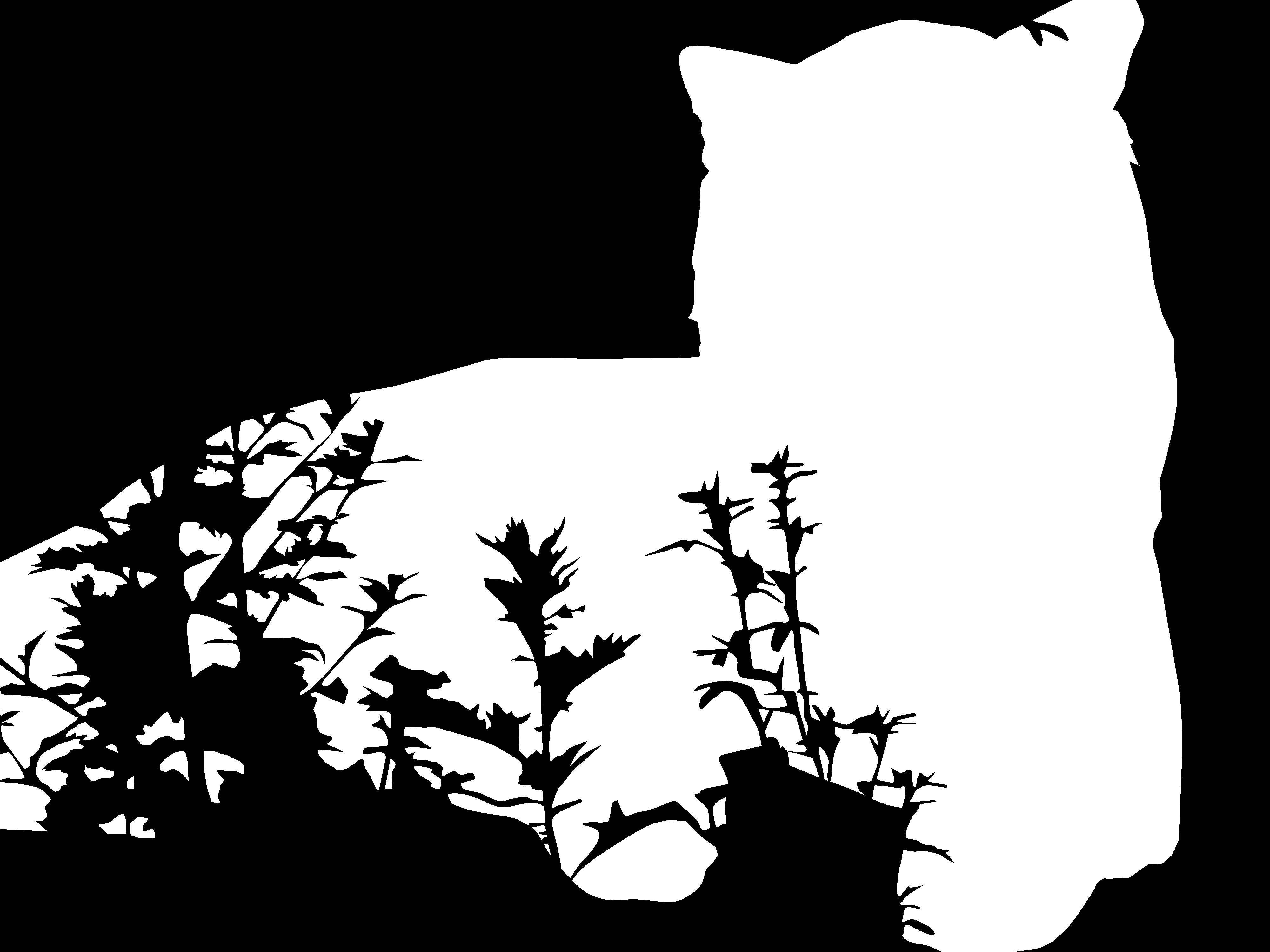}\ &
\includegraphics[width=0.1\linewidth,height=1.7cm]{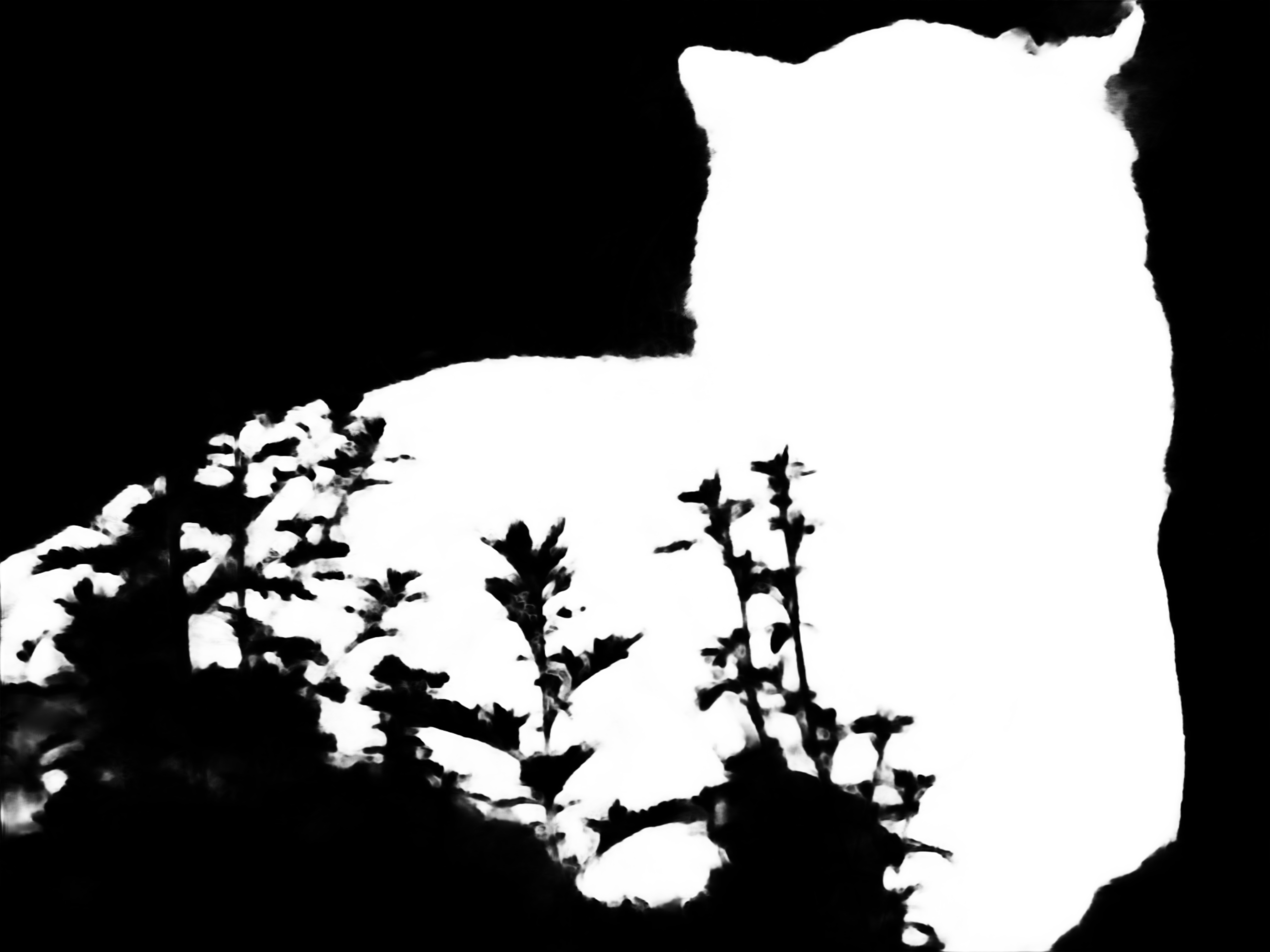}\ &
\includegraphics[width=0.1\linewidth,height=1.7cm]{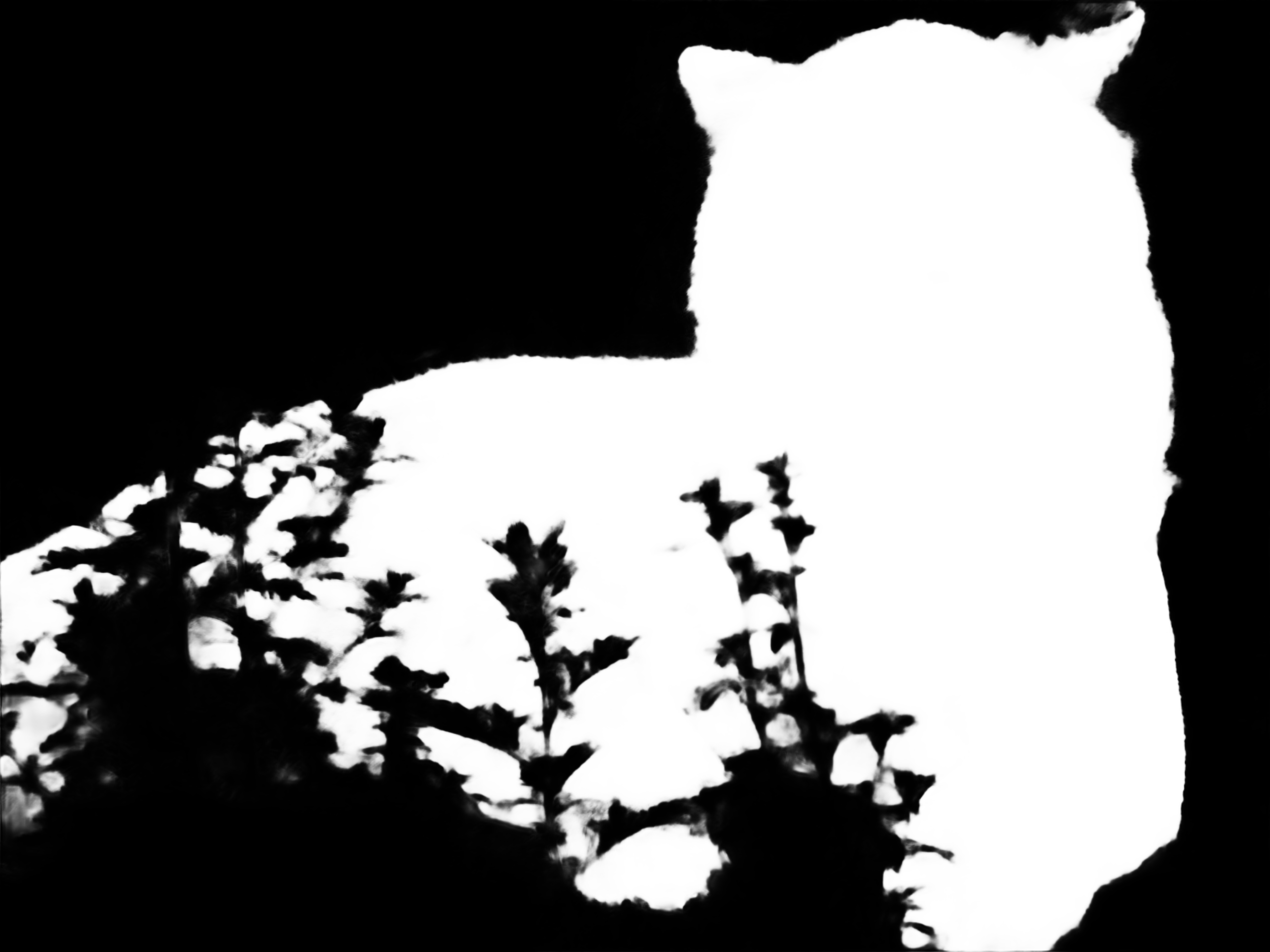}\ &
\includegraphics[width=0.1\linewidth,height=1.7cm]{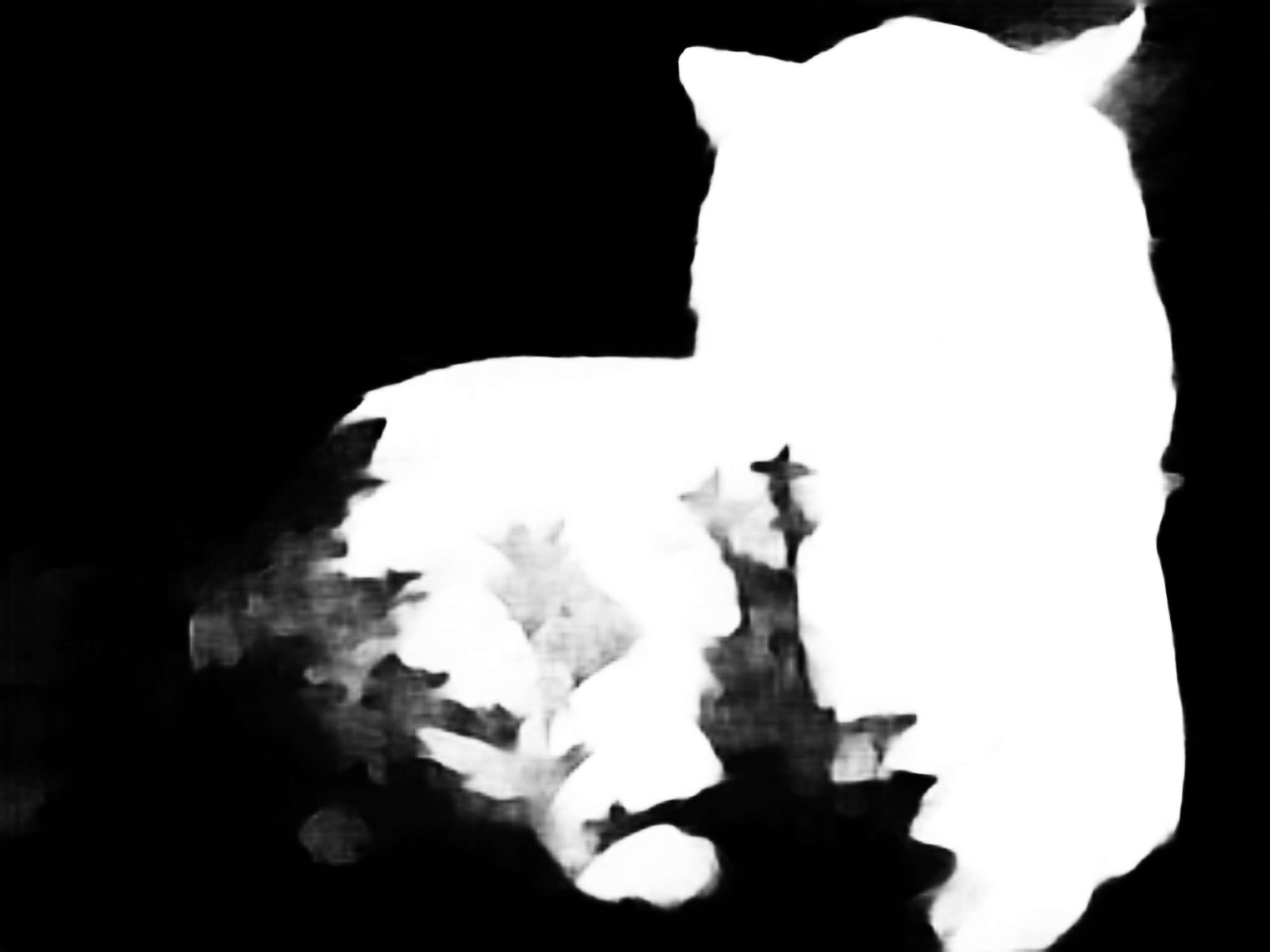}\ &
\includegraphics[width=0.1\linewidth,height=1.7cm]{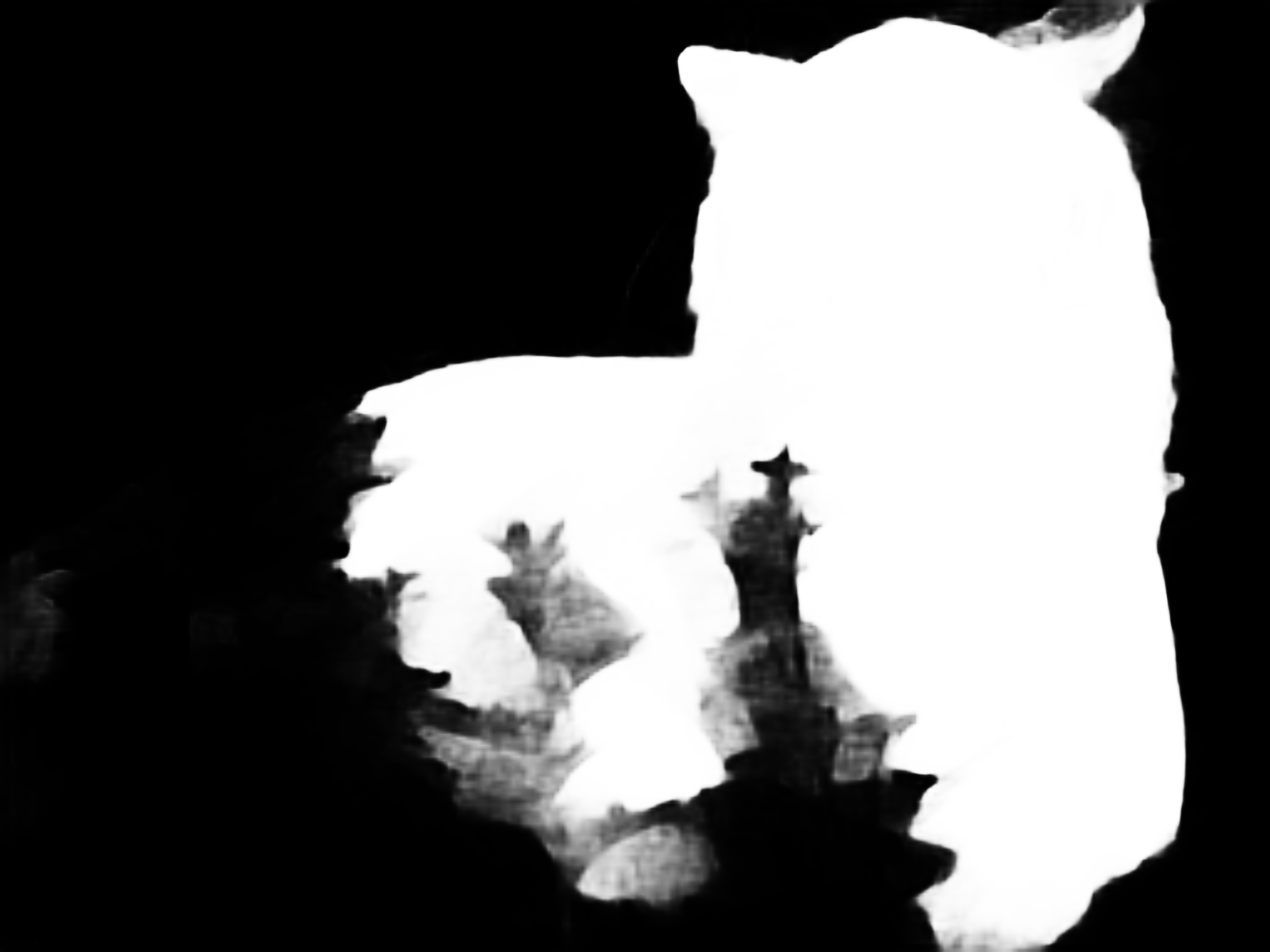}\ &
\includegraphics[width=0.1\linewidth,height=1.7cm]{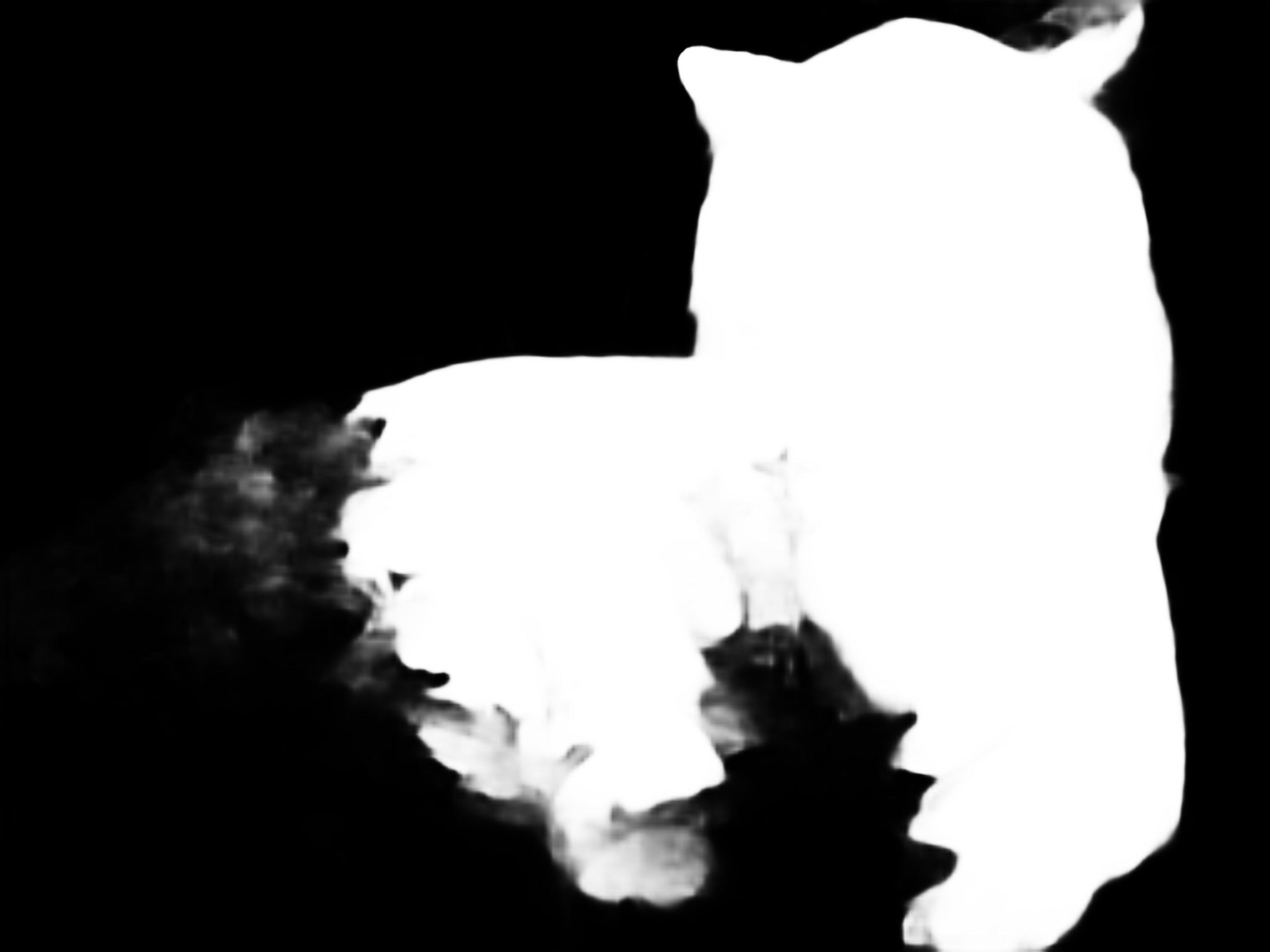}\ &
\includegraphics[width=0.1\linewidth,height=1.7cm]{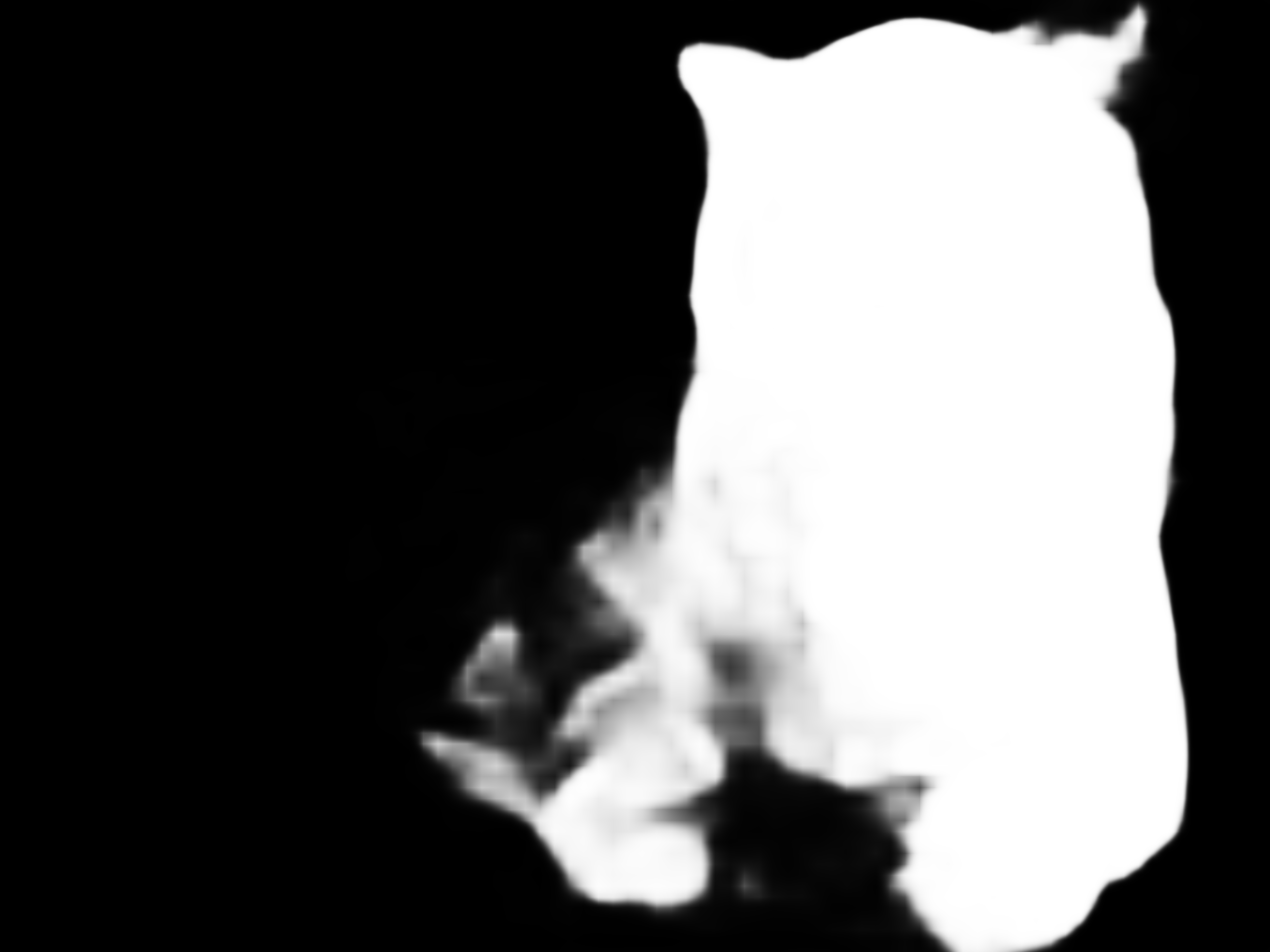}\ &
\includegraphics[width=0.1\linewidth,height=1.7cm]{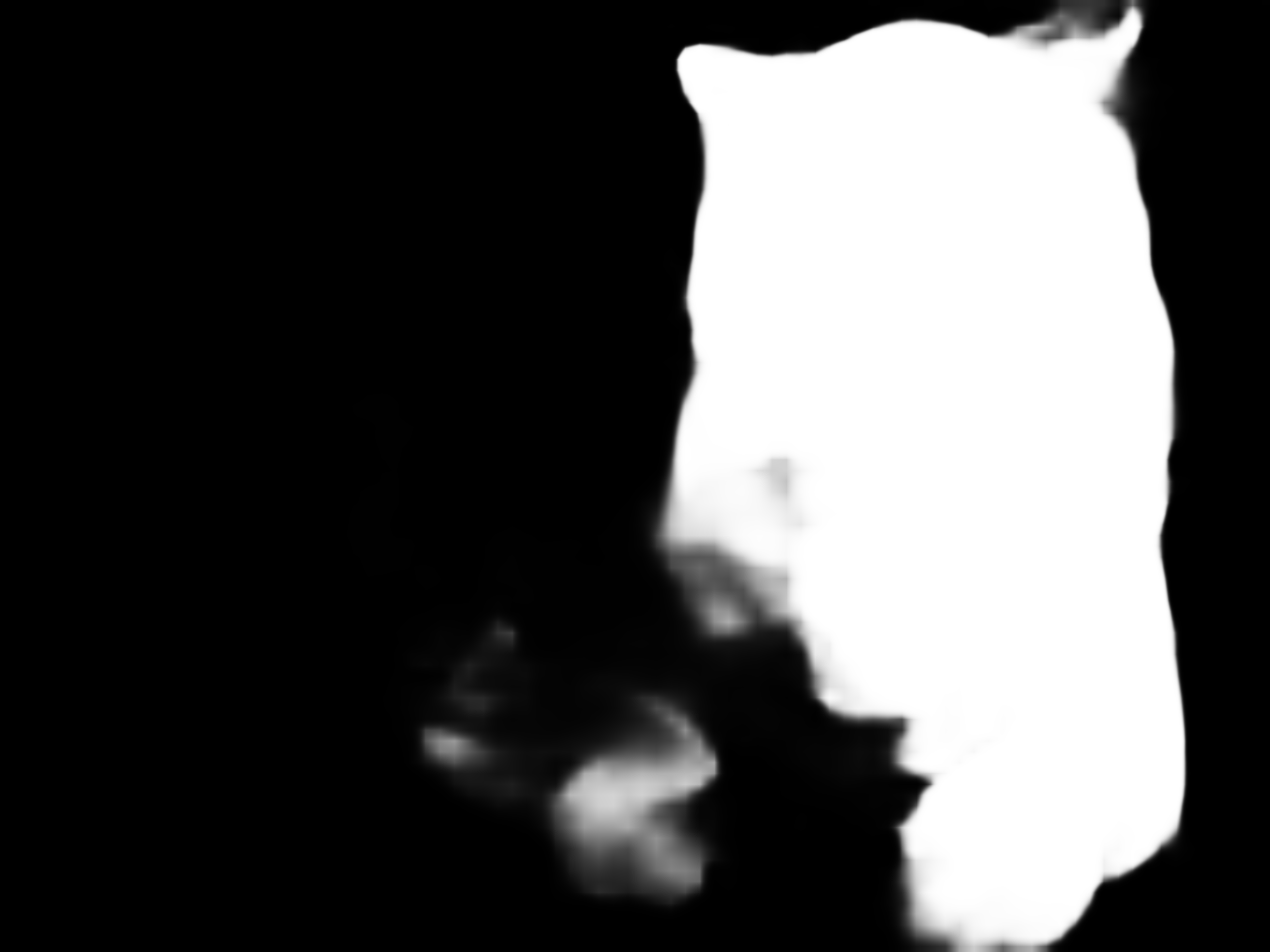}\ &
\includegraphics[width=0.1\linewidth,height=1.7cm]{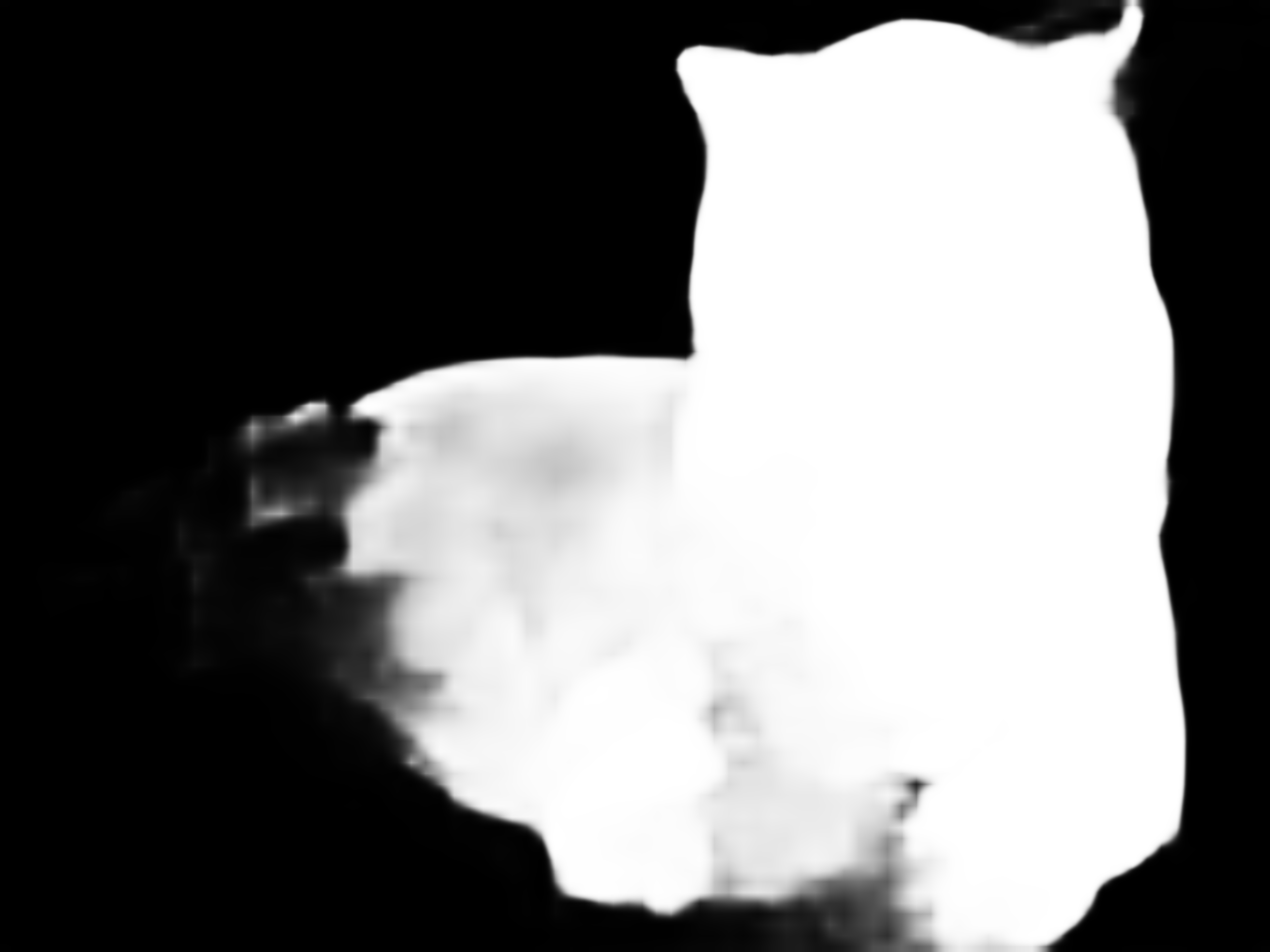}\ \\

\vspace{0.5mm}
\includegraphics[width=0.1\linewidth,height=1.7cm]{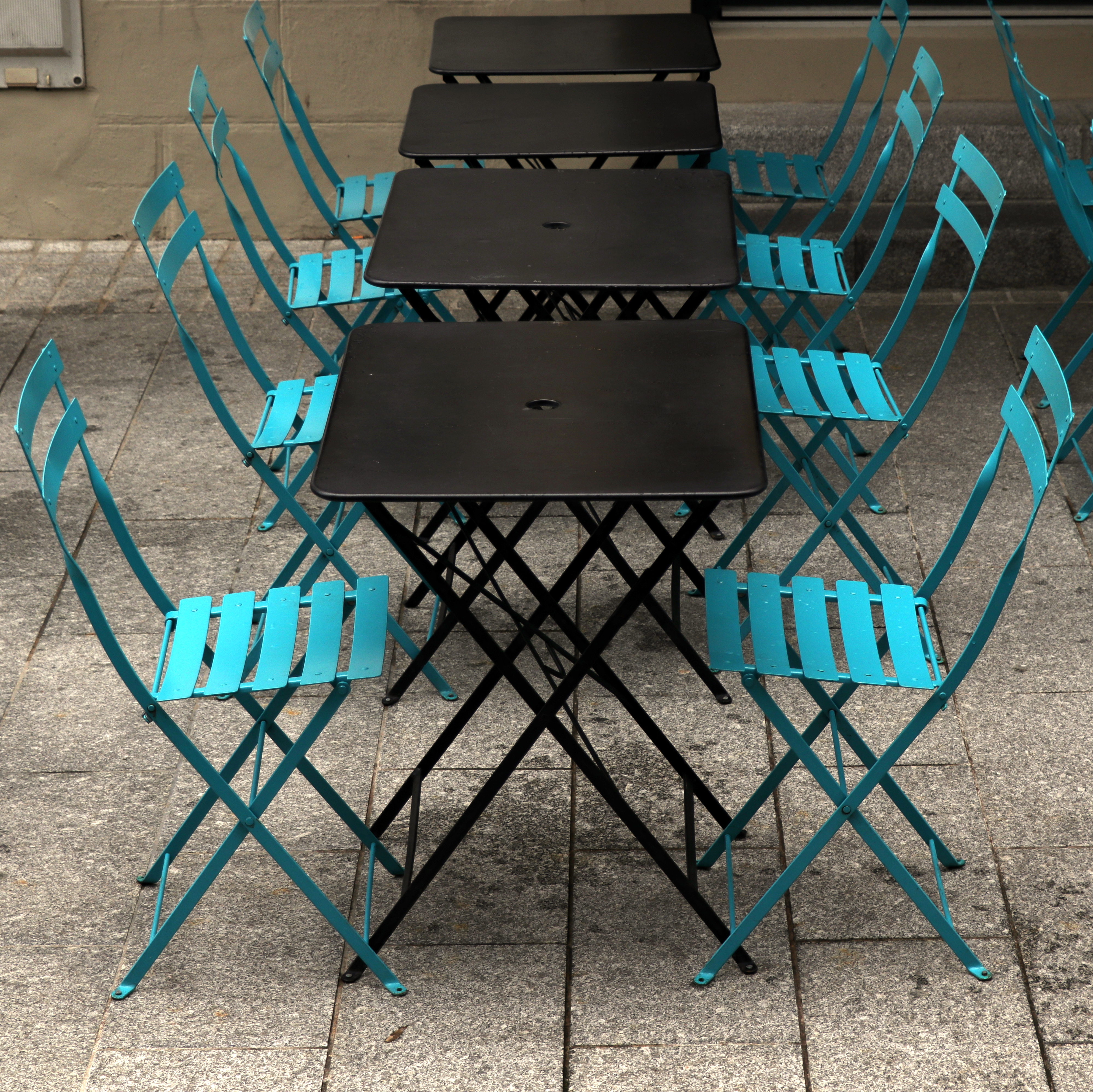}\ &
\includegraphics[width=0.1\linewidth,height=1.7cm]{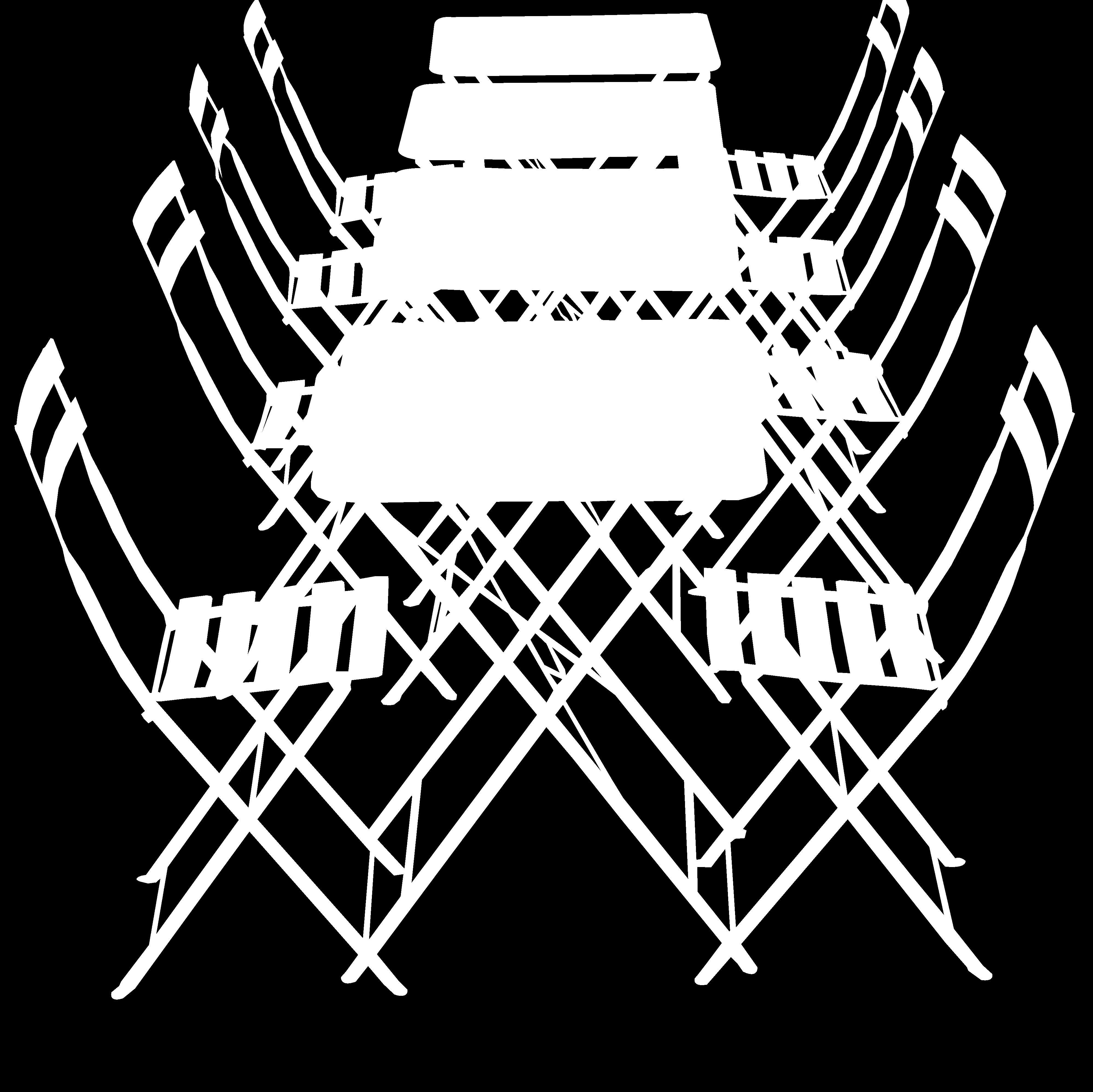}\ &
\includegraphics[width=0.1\linewidth,height=1.7cm]{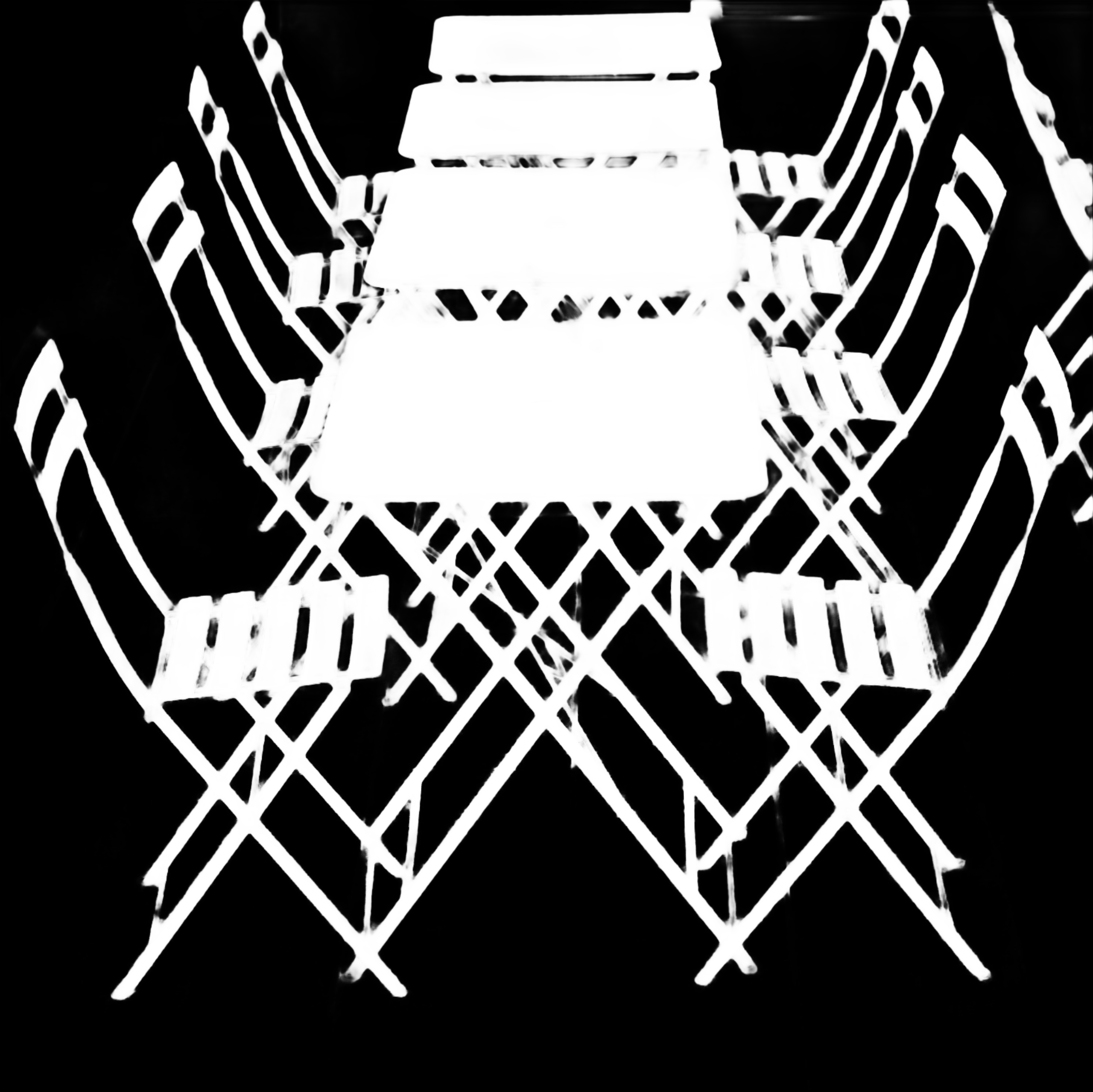}\ &
\includegraphics[width=0.1\linewidth,height=1.7cm]{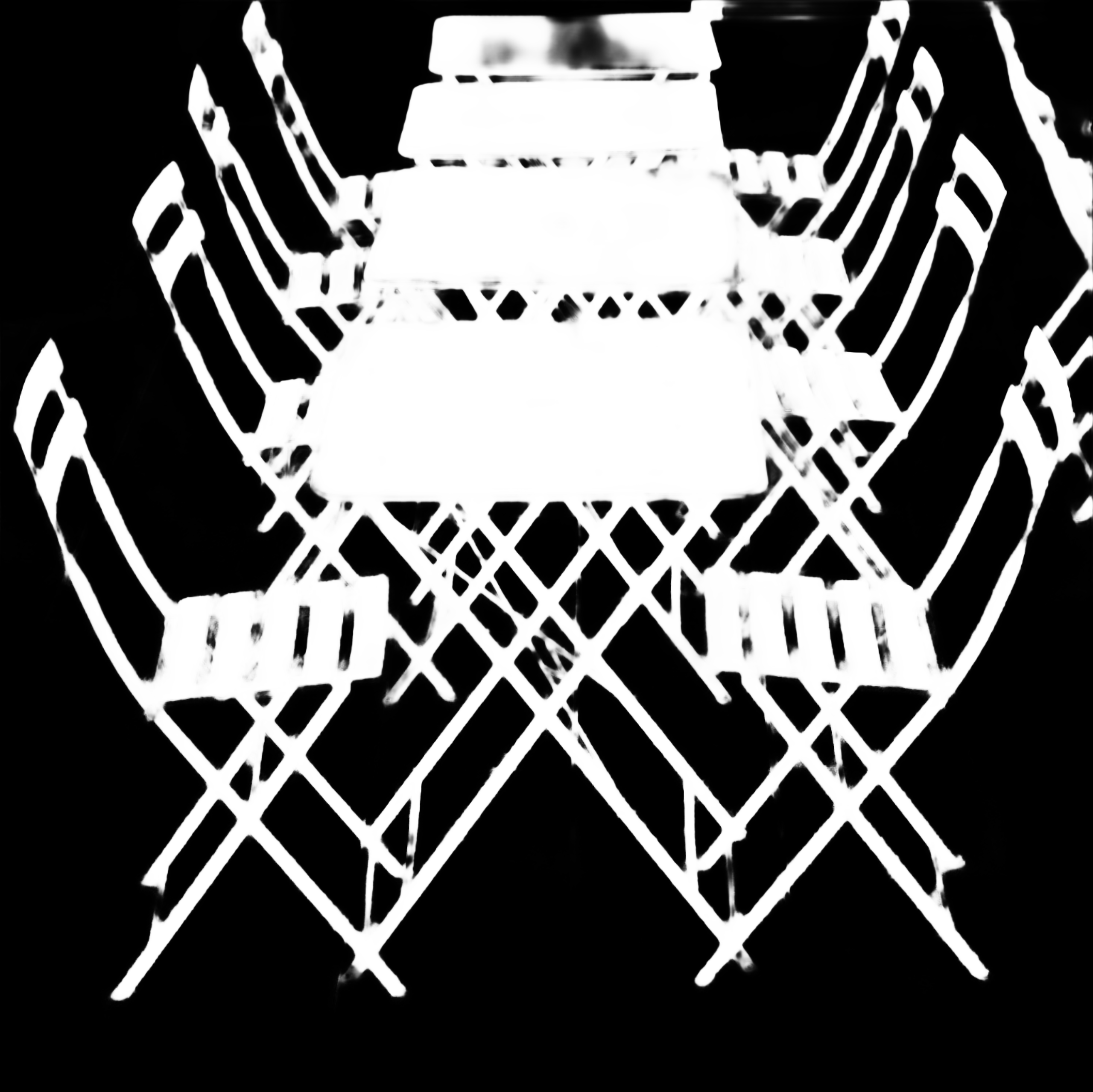}\ &
\includegraphics[width=0.1\linewidth,height=1.7cm]{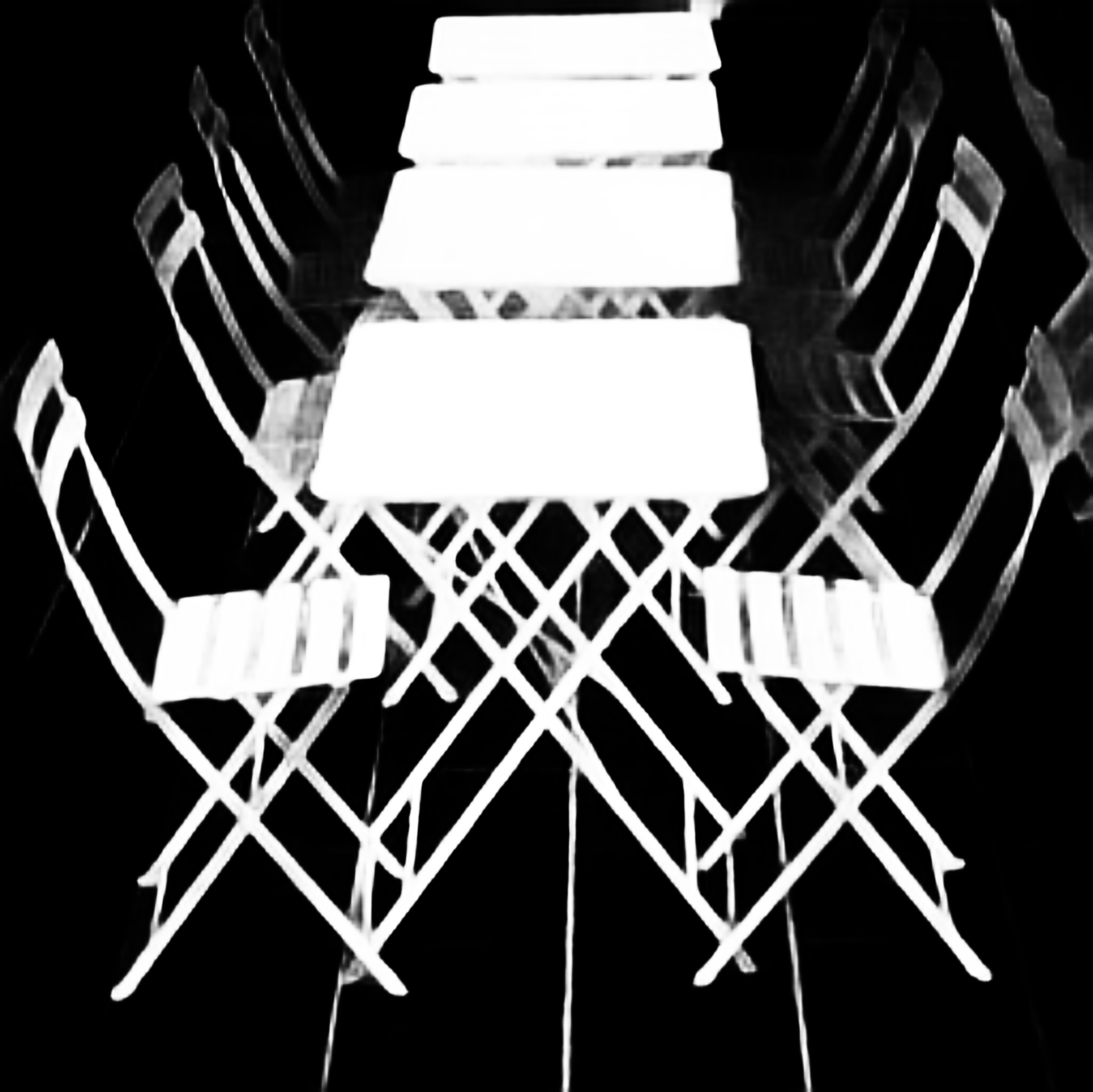}\ &
\includegraphics[width=0.1\linewidth,height=1.7cm]{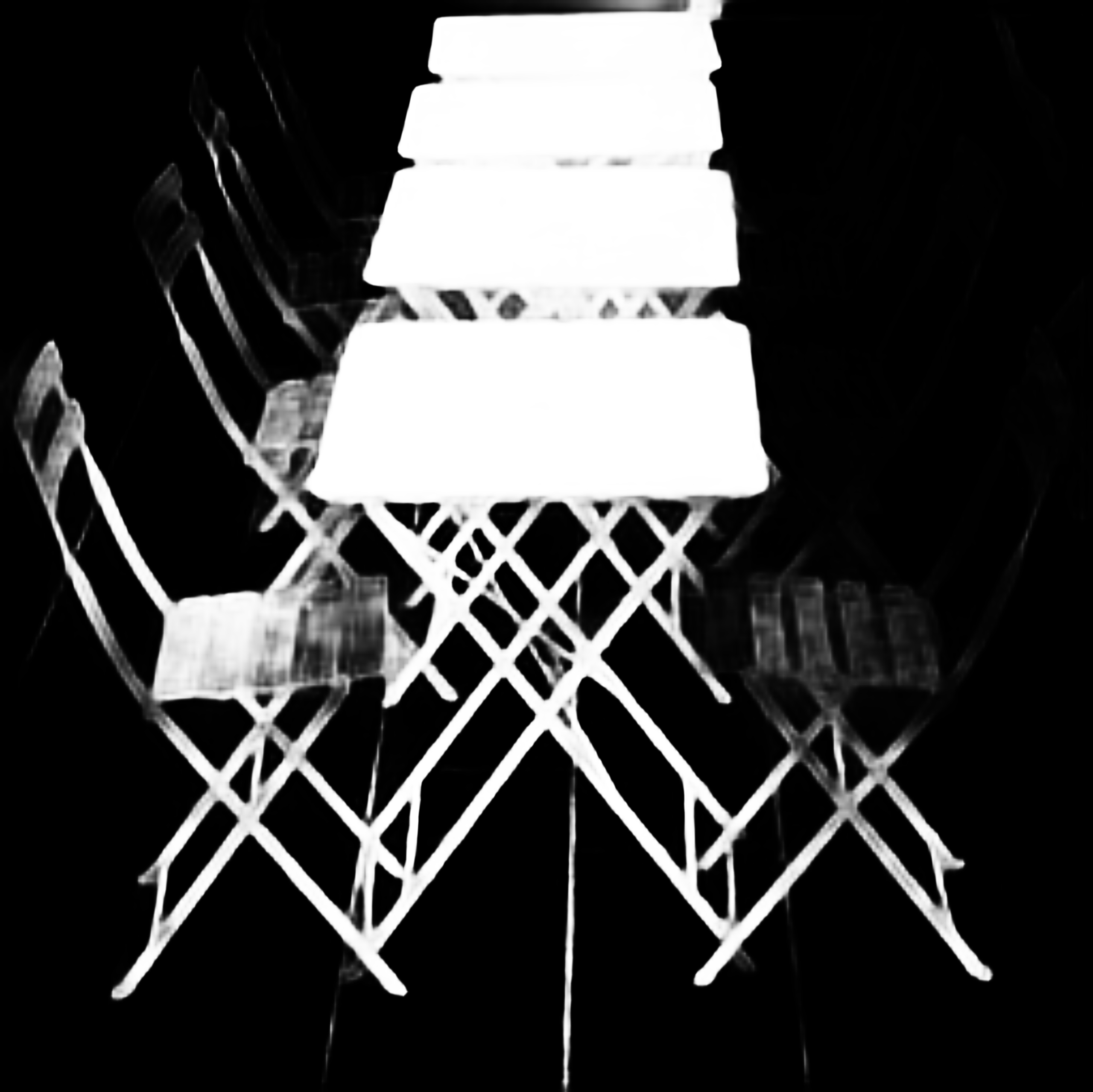}\ &
\includegraphics[width=0.1\linewidth,height=1.7cm]{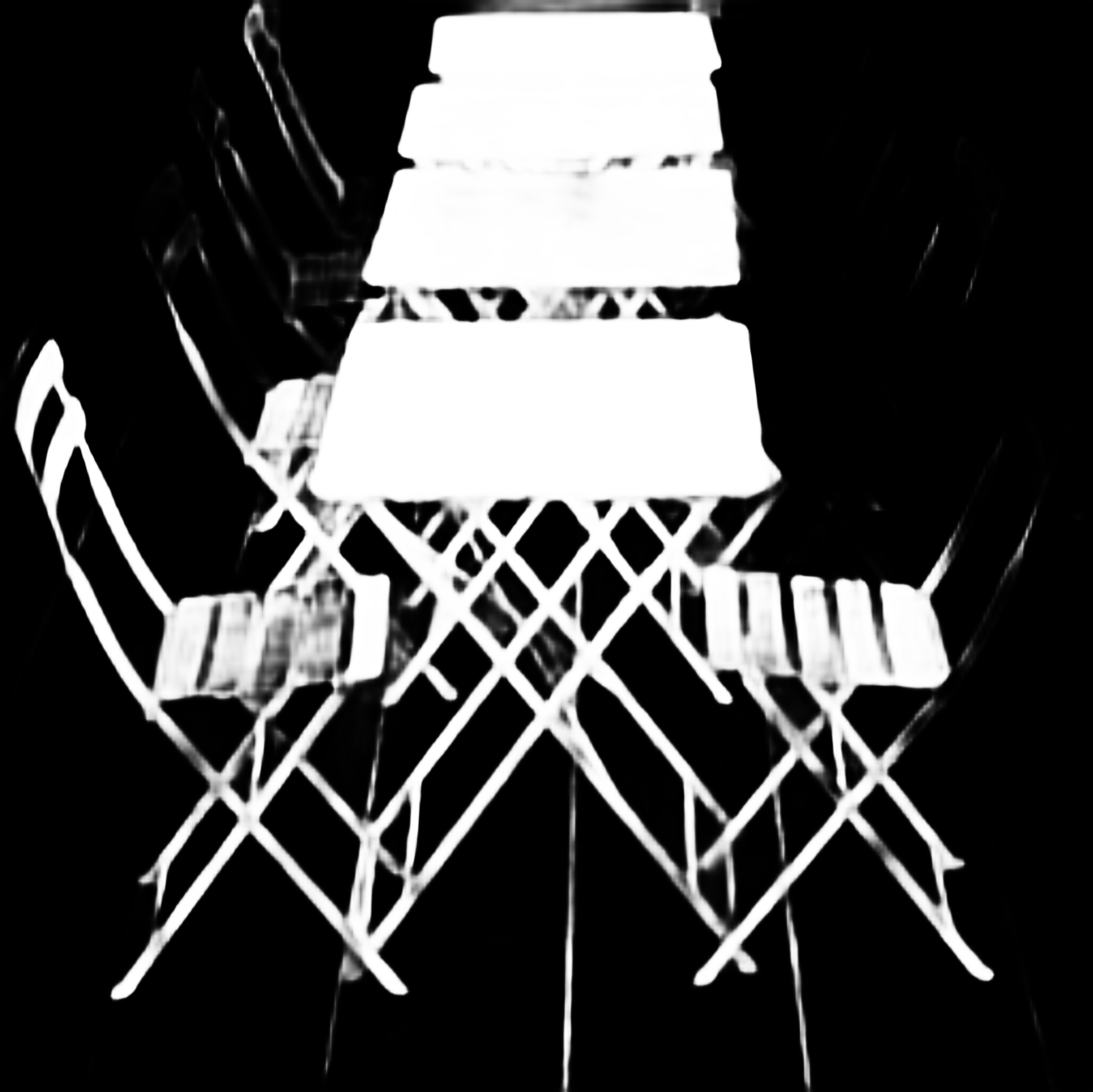}\ &
\includegraphics[width=0.1\linewidth,height=1.7cm]{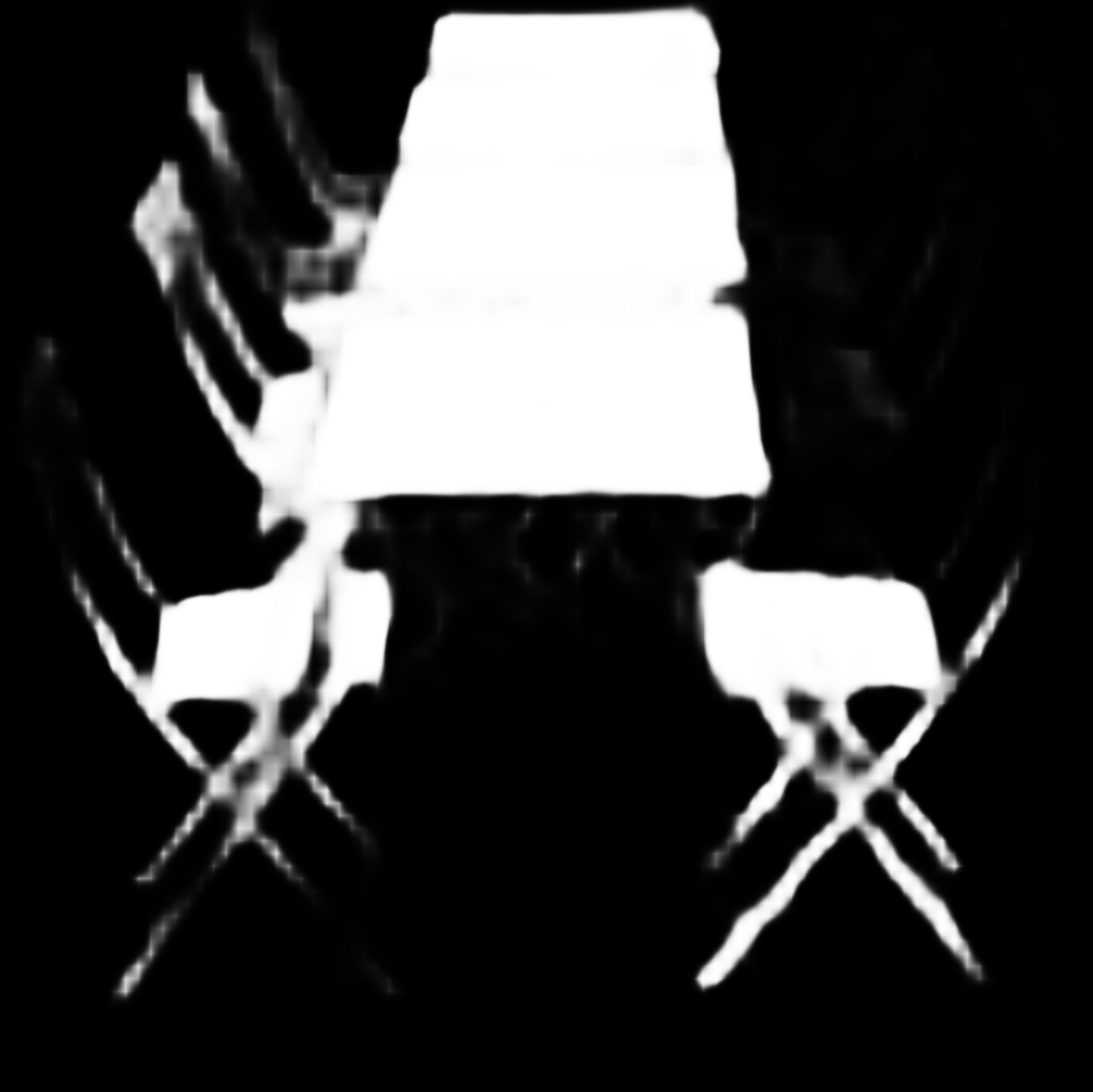}\ &
\includegraphics[width=0.1\linewidth,height=1.7cm]{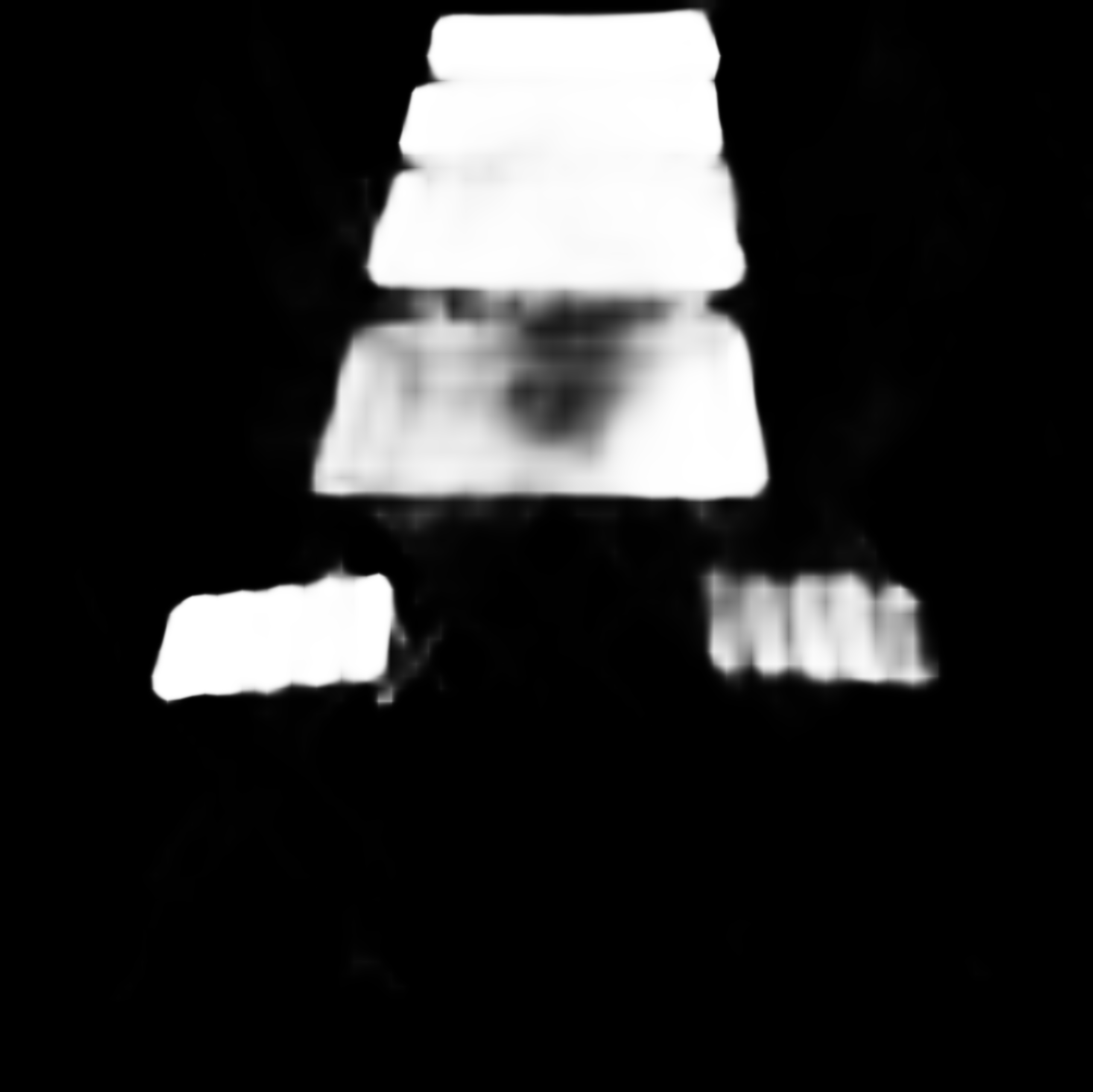}\ &
\includegraphics[width=0.1\linewidth,height=1.7cm]{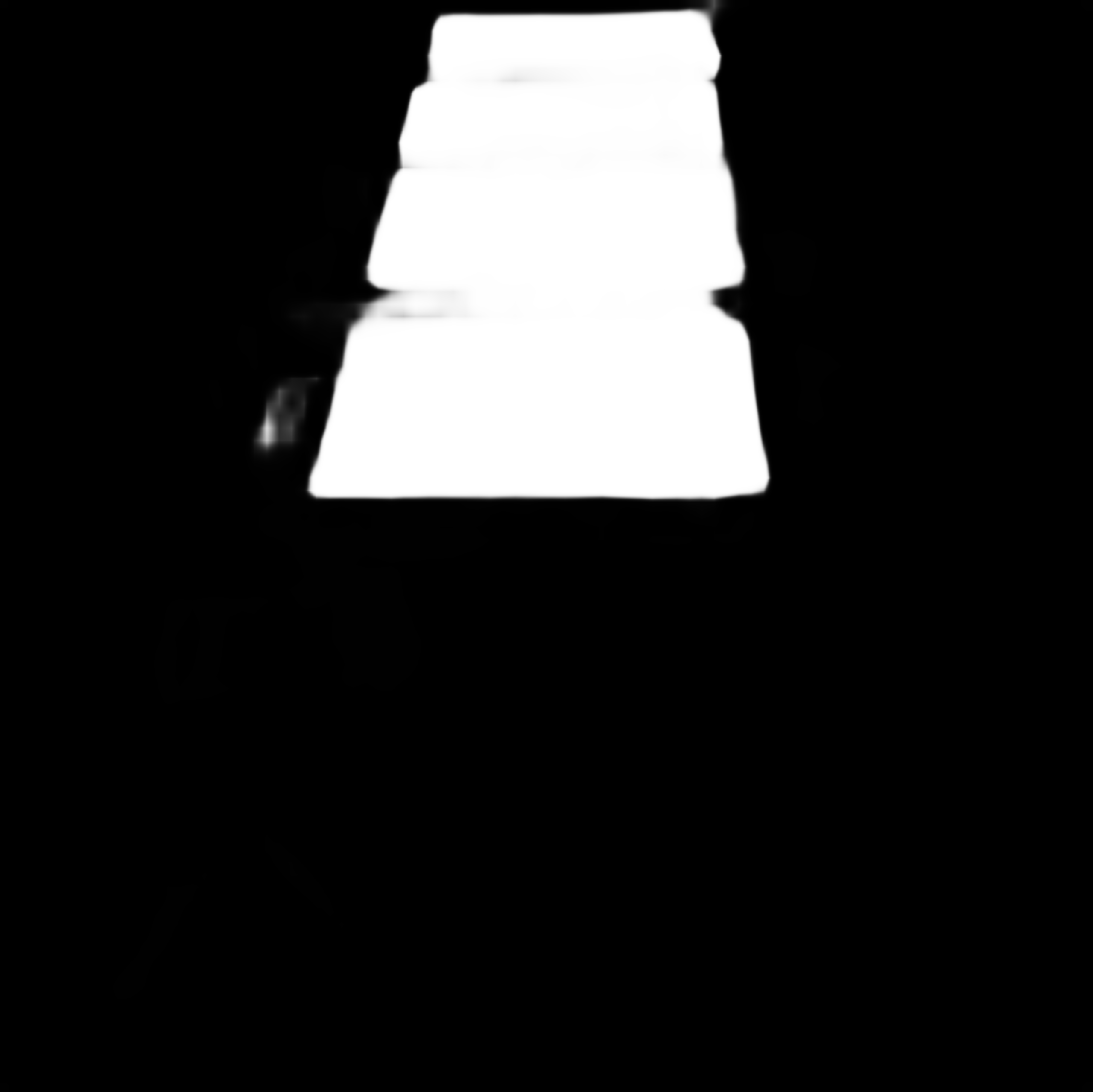}\ \\

\vspace{0.5mm}

\includegraphics[width=0.1\linewidth,height=1.7cm]{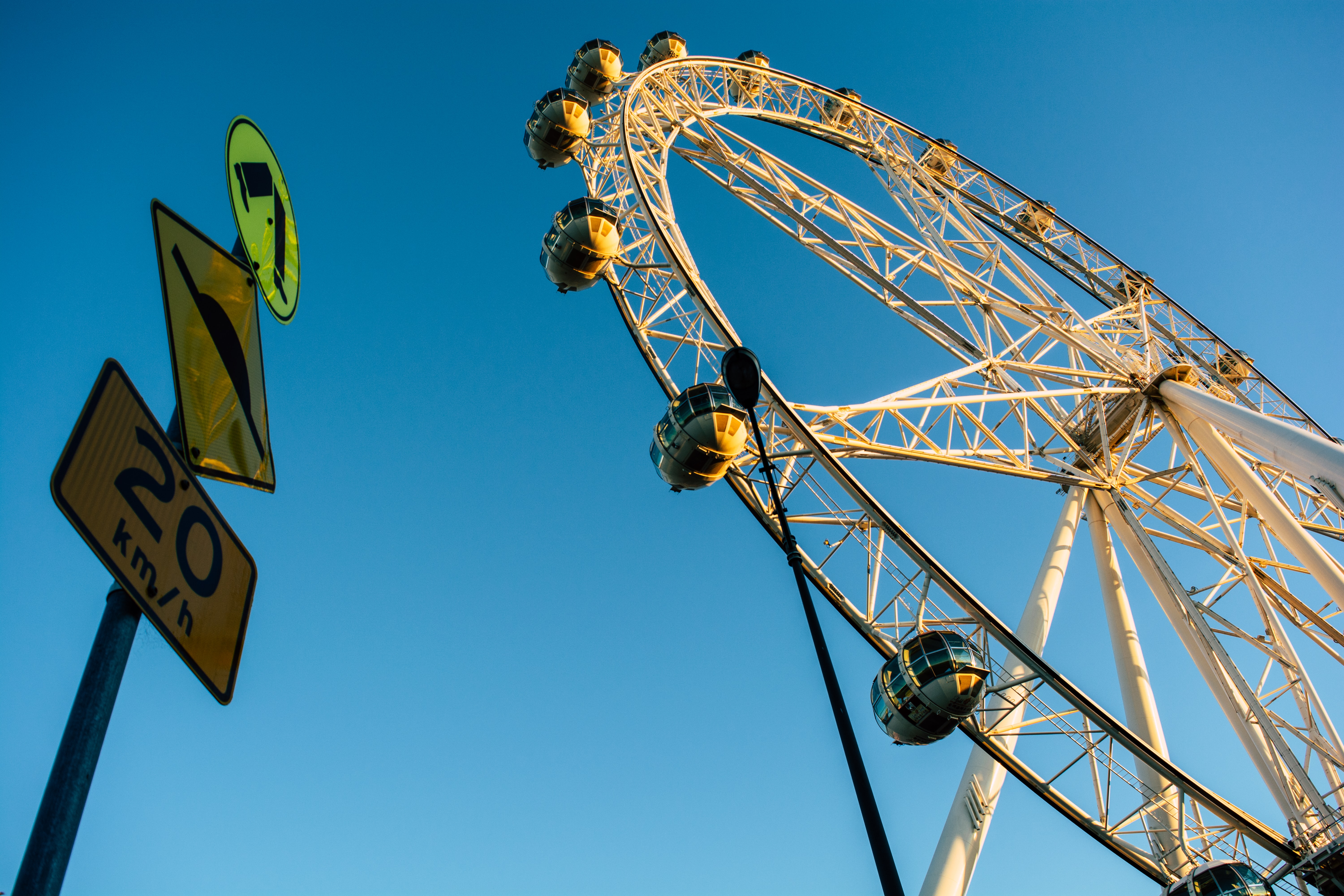}\ &
\includegraphics[width=0.1\linewidth,height=1.7cm]{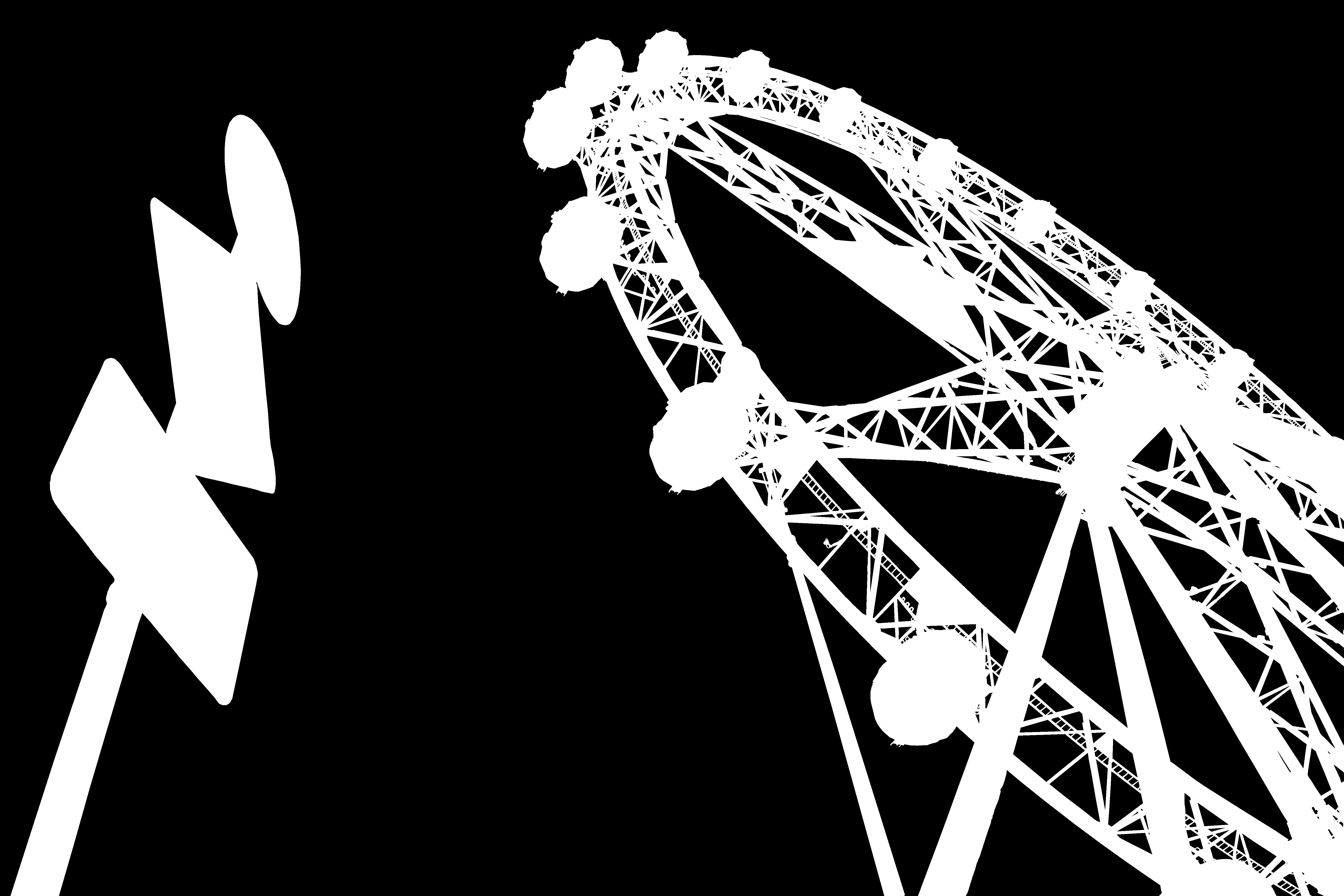}\ &
\includegraphics[width=0.1\linewidth,height=1.7cm]{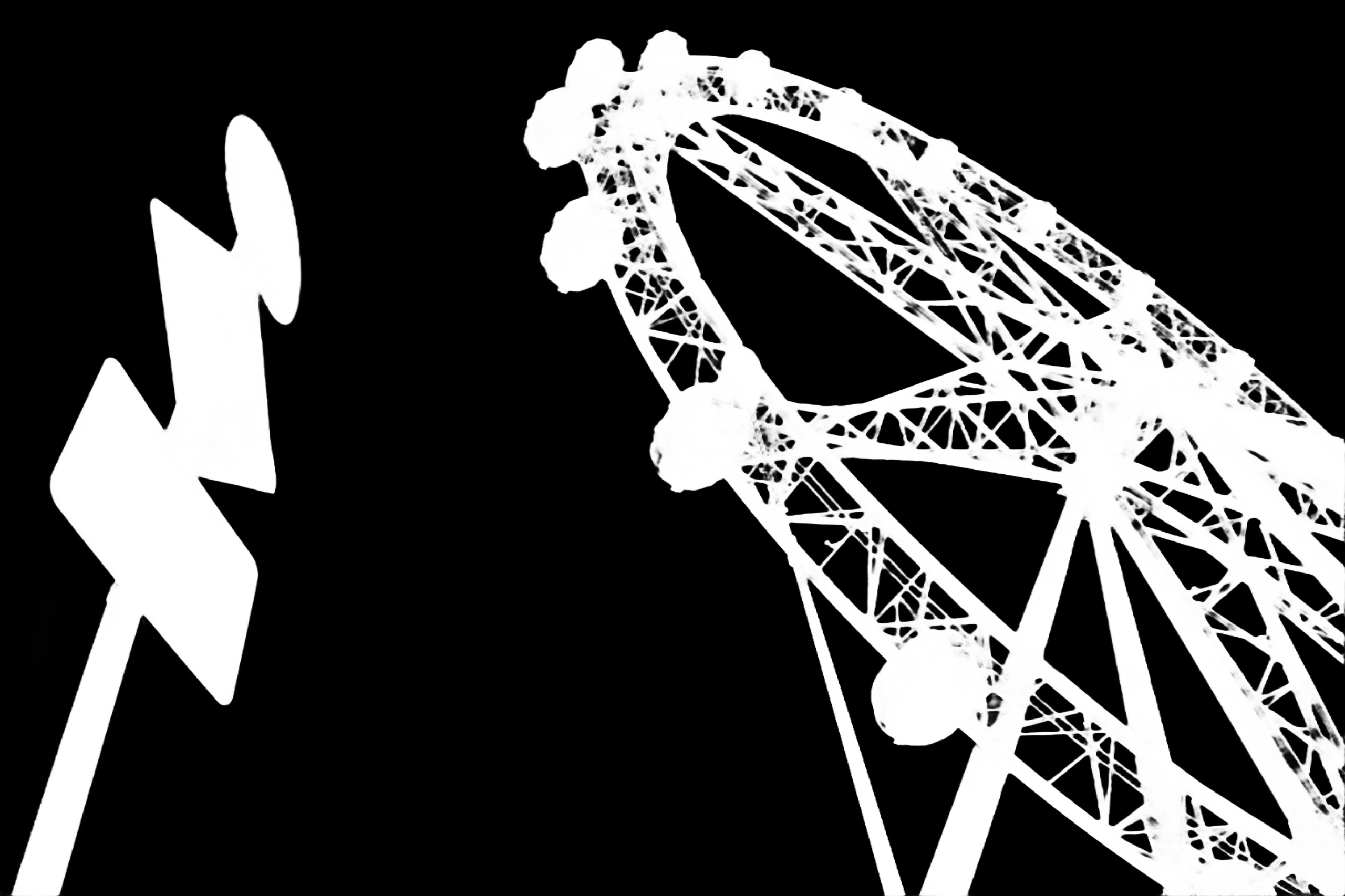}\ &
\includegraphics[width=0.1\linewidth,height=1.7cm]{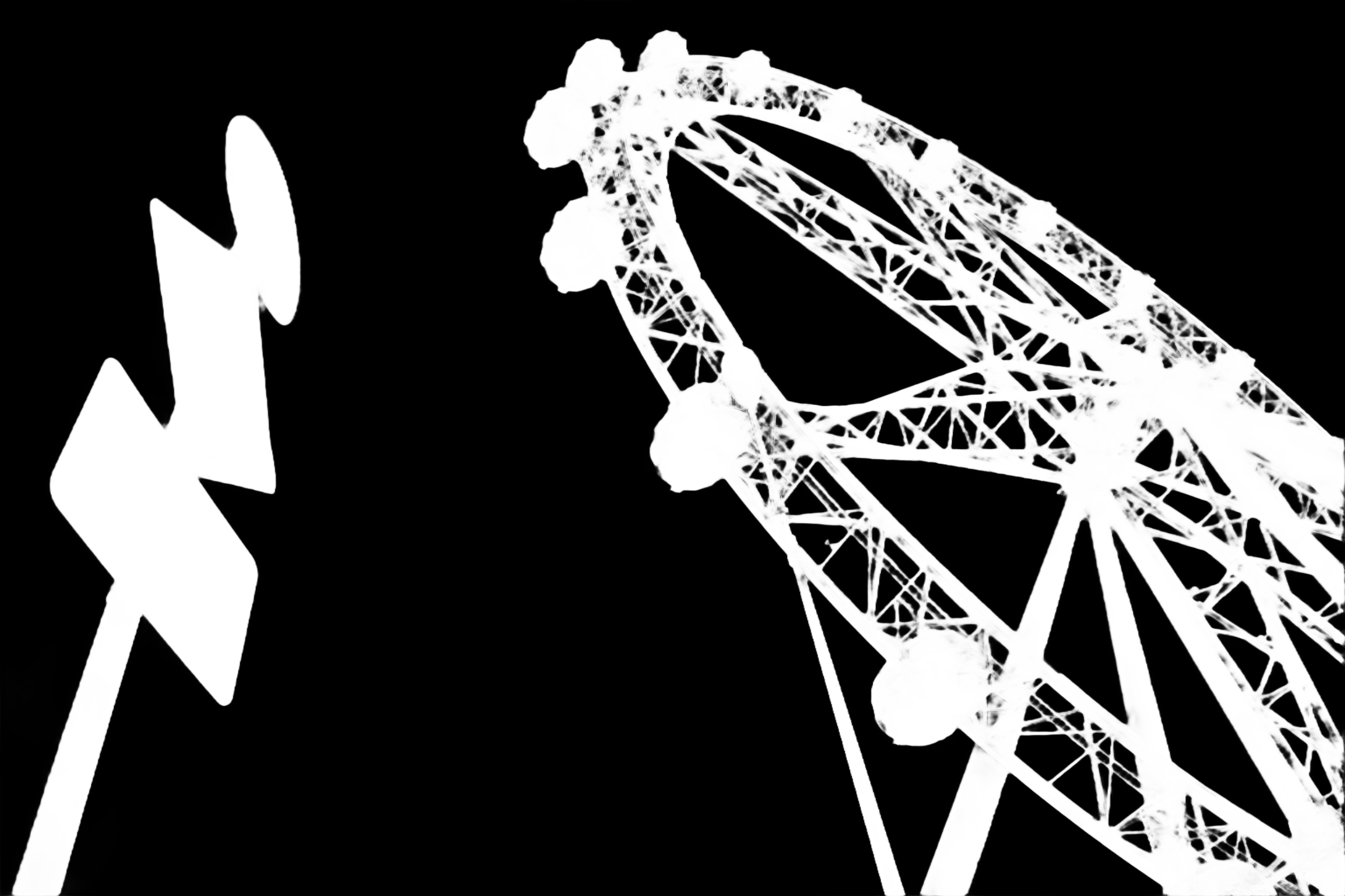}\ &
\includegraphics[width=0.1\linewidth,height=1.7cm]{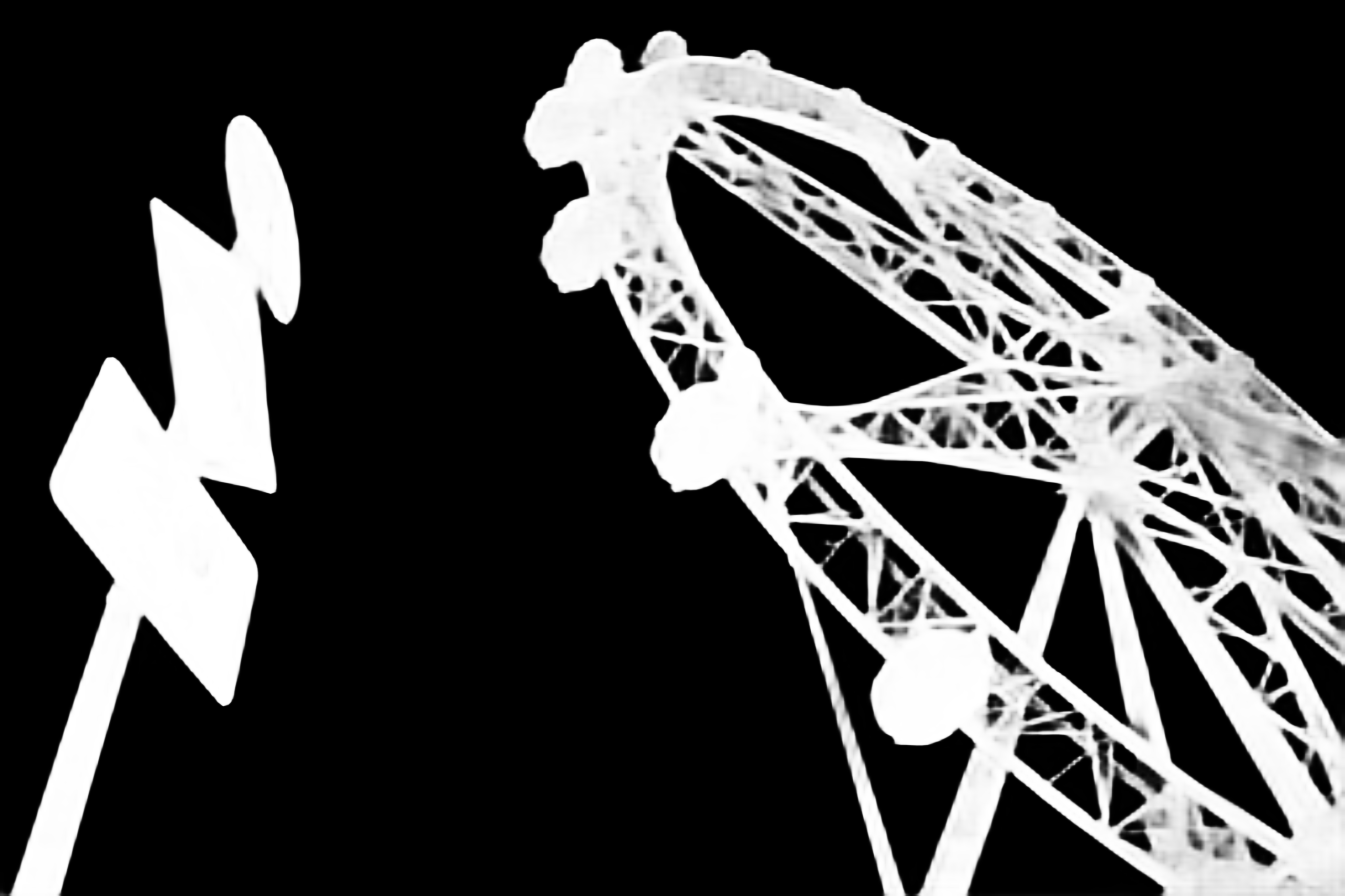}\ &
\includegraphics[width=0.1\linewidth,height=1.7cm]{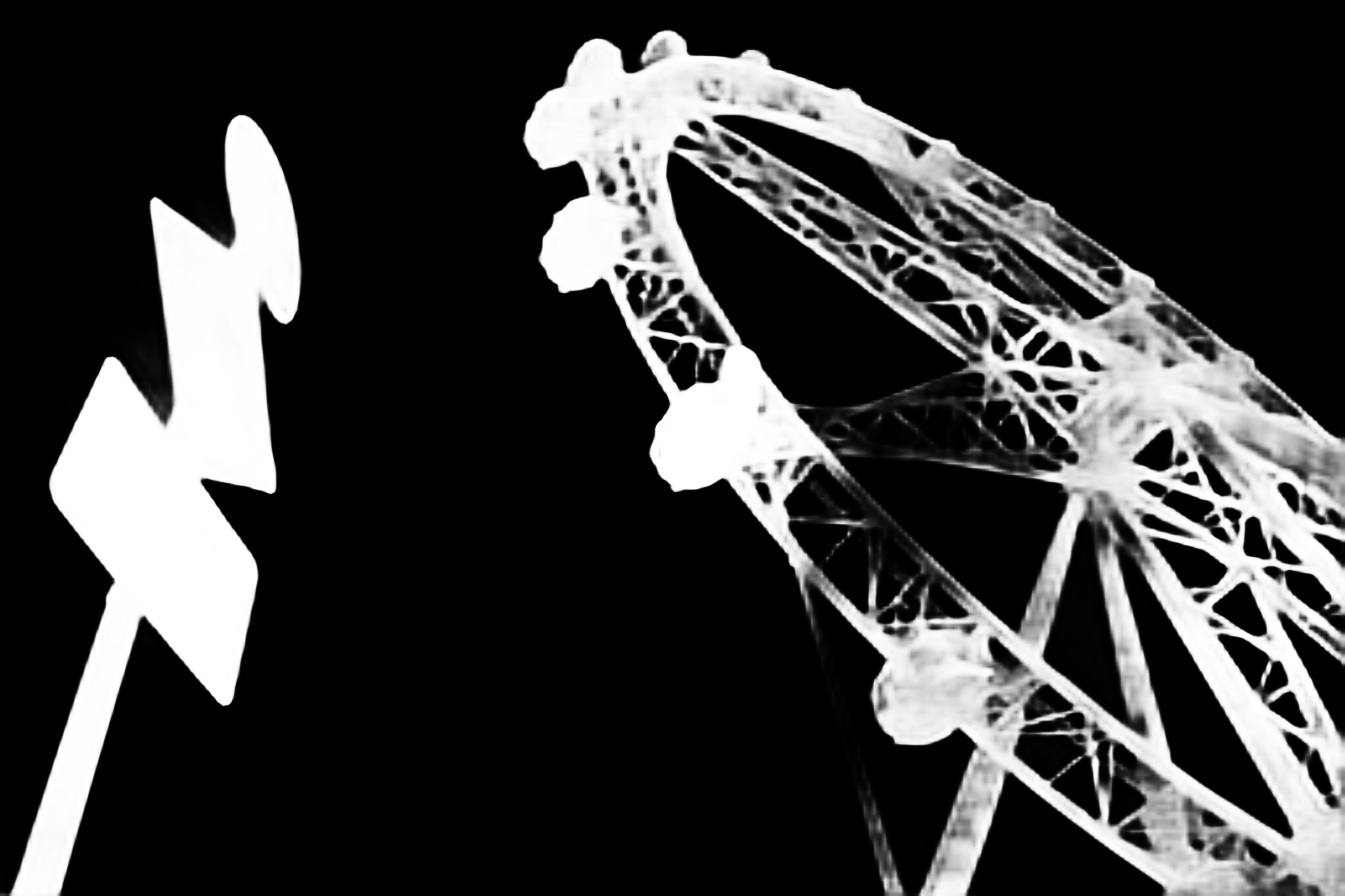}\ &
\includegraphics[width=0.1\linewidth,height=1.7cm]{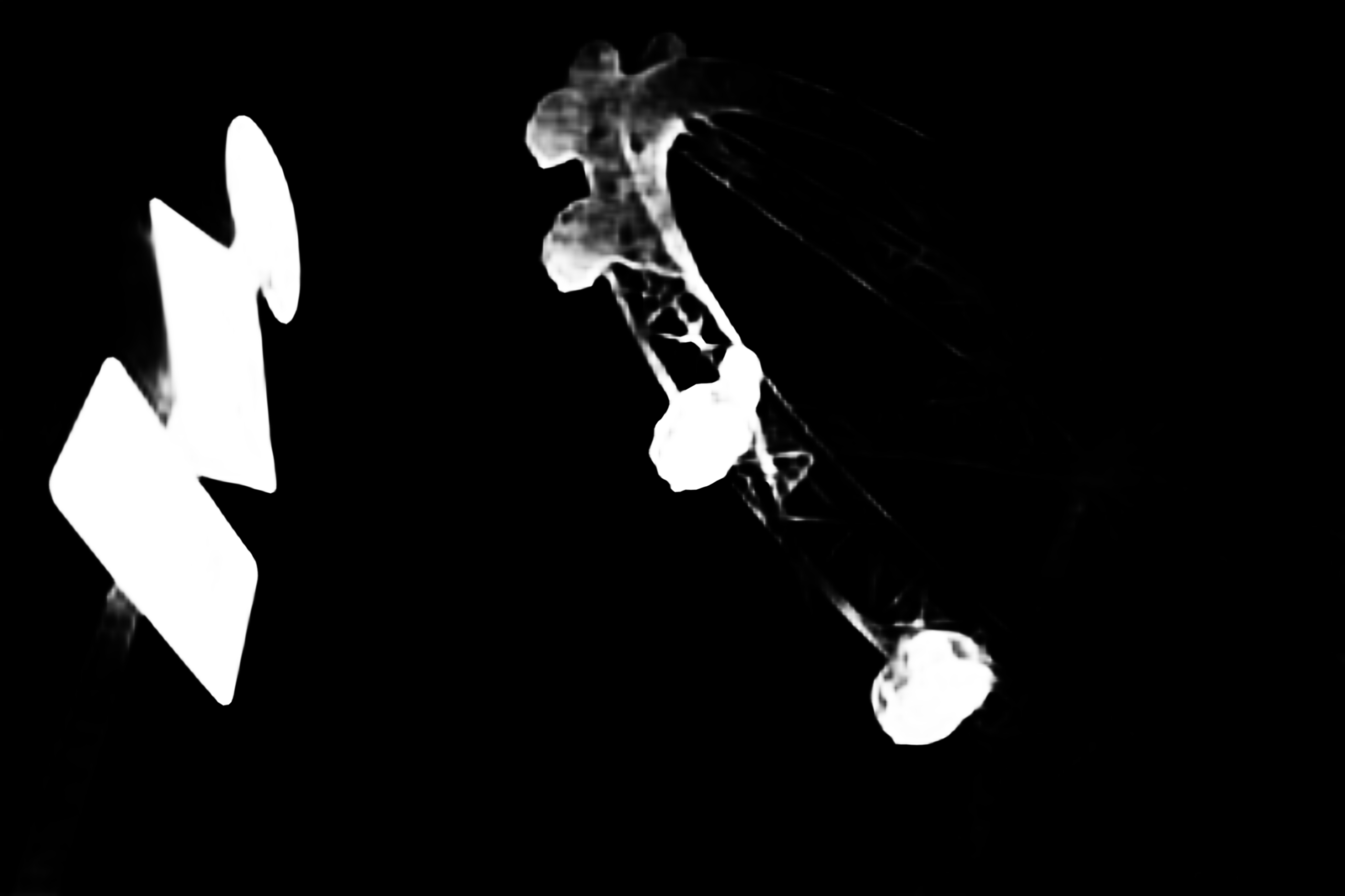}\ &
\includegraphics[width=0.1\linewidth,height=1.7cm]{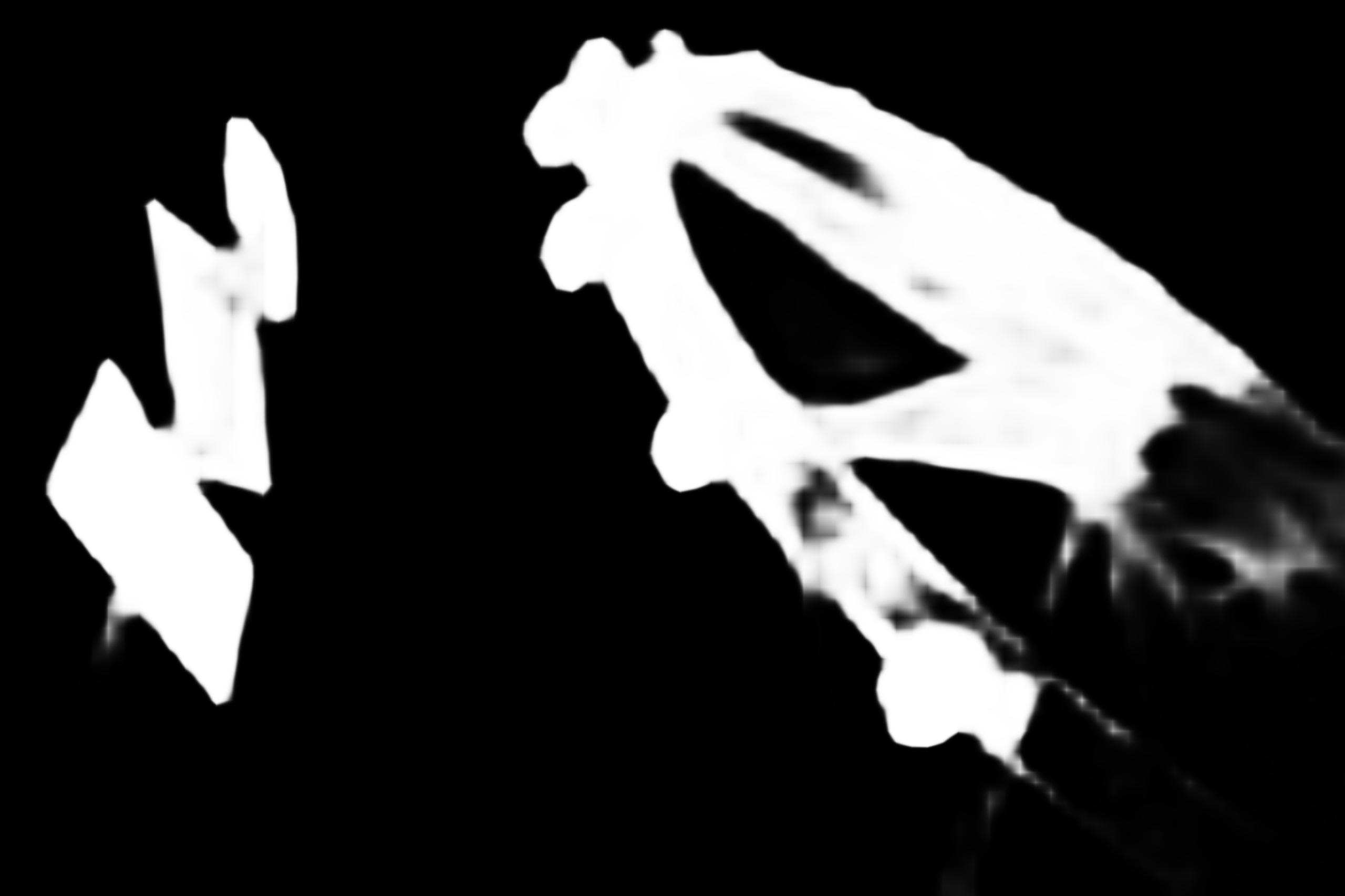}\ &
\includegraphics[width=0.1\linewidth,height=1.7cm]{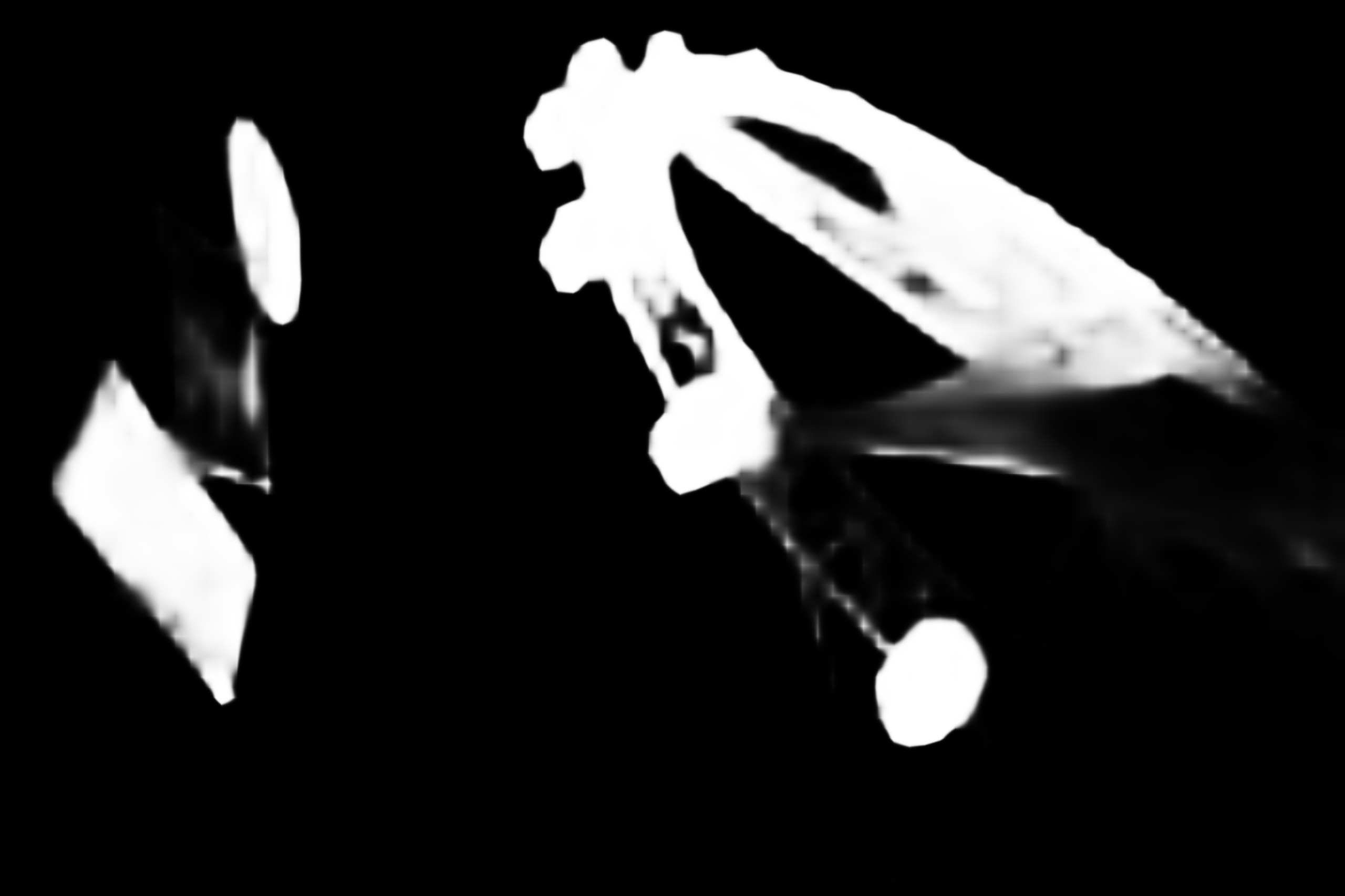}\ &
\includegraphics[width=0.1\linewidth,height=1.7cm]{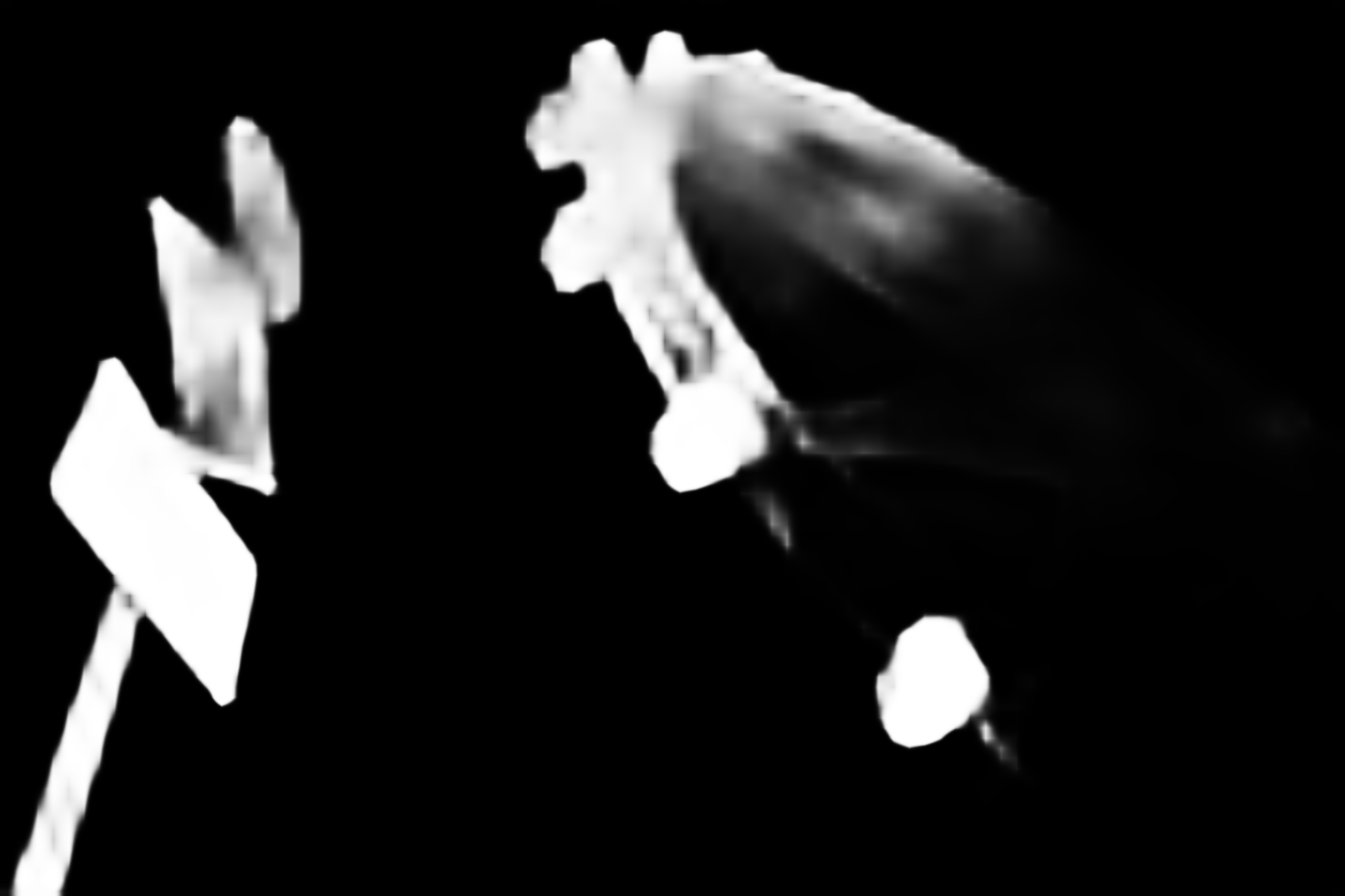}\ \\

\vspace{0.5mm}
 {\small Image} & {\small GT} & {\small Ours-KUH} & {\small Ours-UH}  & {\small PGNet-KUH} & {\small PGNet-UH} & {\small PGNet-DH} & {\small CTDNet} & {\small LDF} & {\small F3Net}\ \\
\end{tabular}
}
\vspace{-3mm}
\caption{Visual comparison of predicted saliency maps with different methods. The image in the first row is from the UHRSD-TE and the rest of them are  from our HRS10K-TE. Best view by zooming in.}
\label{fig:visual}
\end{figure*}

\begin{figure}[!t]
    \centering
    \includegraphics[width=0.9\linewidth]{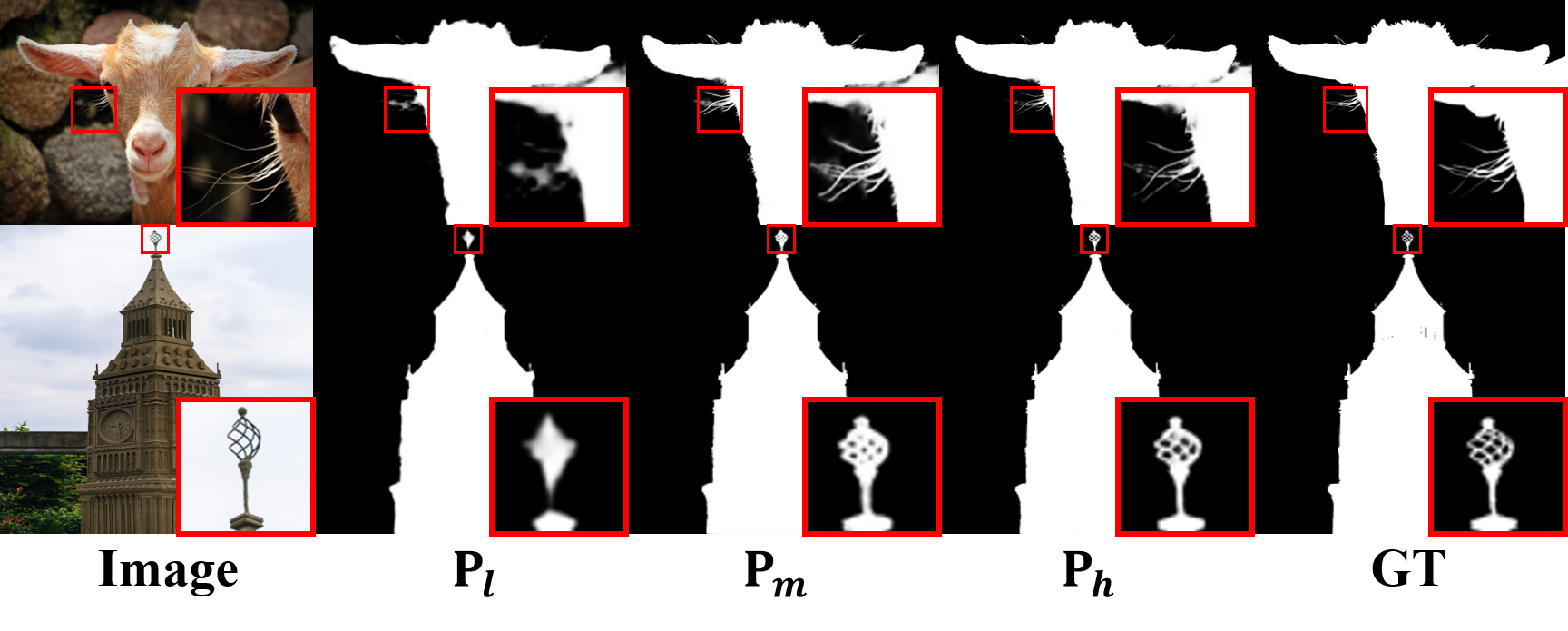}

    \vspace{-4mm}
    
    \caption{Saliency predictions at different stages. \(\textbf{P}_{l}\) is the prediction by CPS, and \(\textbf{P}_{m}\) and \(\textbf{P}_{h}\) are the prediction of two refinement stages. Best view by zooming in.}
    \label{fig:visual ablation1}
    \vspace{-2mm}
\end{figure}

\begin{figure}[!]
    \centering
    \includegraphics[width=0.9\linewidth]{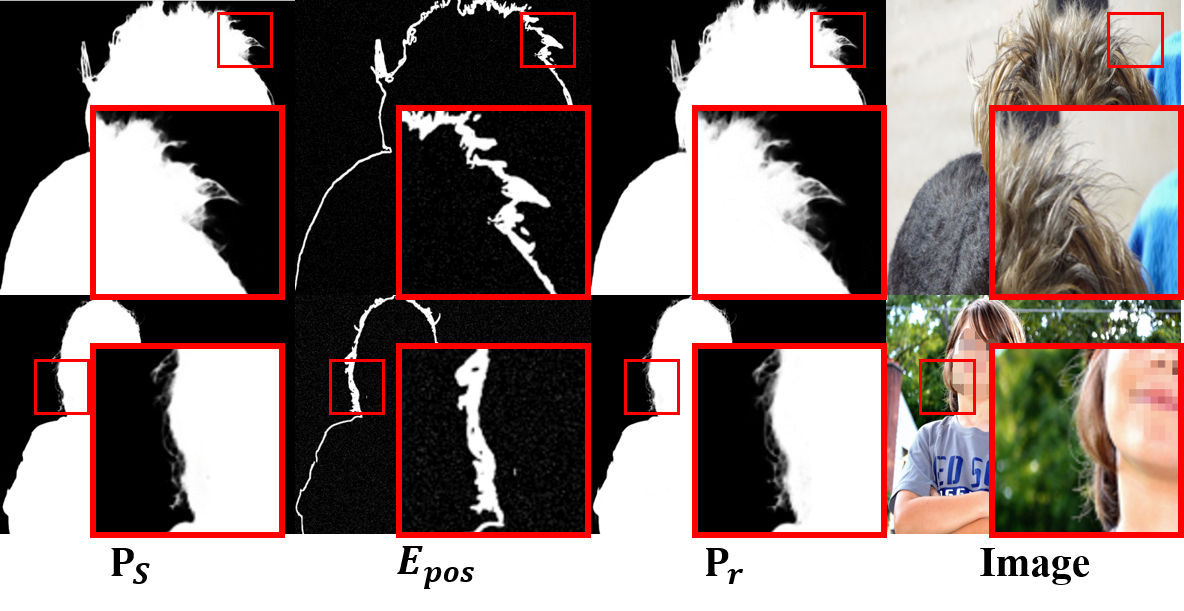}

    \vspace{-4mm}

    \caption{Visual effectiveness of the PR. \(\textbf{P}_{S}\) is the initial prediction, \(\textbf{E}_{pos}\) is the selected pixels region, and \(\textbf{P}_{r}\) is the refined prediction. Best view by zooming in.}
    \label{fig:visual ablation pr}

    \vspace{-2mm}
\end{figure}

\textbf{Qualitative Evaluation}. Figure \ref{fig:visual} shows the visual results of our method with respect to others. Our method captures more details of salient objects, especially on the fluffy object edge. Besides, by using the KUH training setting, the result shows more fine-grained structures on both our method and the previous PGNet, which also proves the usefulness of HRS10K dataset.
\subsection{Ablation Study}
To further explore the nature of our proposed method, we conduct several experiments in the UH training setting. 

\textbf{Ablation of Different Prediction Stages}. In our method, RRS refines the coarse saliency prediction to a higher resolution. We show the ablation results in Table \ref{tab:stage abl}. CPS uses the pre-trained backbone and the coarse prediction head only. The low-resolution prediction is upsampled to high-resolution for evaluation. The results show that our proposed RRS can effectively capture high-resolution information thus boost the performance. Comparing row 2 and row 3 in Table \ref{tab:stage abl}, we can conclude that the higher resolutions can bring more benefits to the final prediction.
\begin{table}[h!t]
    \renewcommand{\arraystretch}{1.2}
    \setlength{\tabcolsep}{1pt}
    \centering 
    \caption{Comparison with different stages.} 
    \vspace{-2mm}
    \resizebox{0.8\linewidth}{!}{
    \begin{tabular}{l | ccc| ccc} 
        \hline
        \multicolumn{1}{l|}{\multirow{2}*{Stages}} & \multicolumn{3}{c|}{HRSOD-TE} 
                                                    & \multicolumn{3}{c}{UHRSD-TE}\\

        \cline{2-7}
        \multicolumn{1}{l|}{} &   MAE&    \(F^{max}_{\beta}\)&  mBA&
                                MAE&    \(F^{max}_{\beta}\)&    mBA\\

        \hline  
        CPS&   
        .0250&  .927&   .651&
        .0265&  .948&   .682\\
        
        CPS + RRS1&   
        .0222&  .934&   .695&
        .0220&  .956&   .741\\

        CPS + RRS2&   
        .0203&  .940&   .717&
        .0205&  .956&   .764\\

        CPS + RRS1 + RRS2&   
        .0191&  .940&   .736&
        .0187&  .956&   .784\\

        \hline 
    \end{tabular}
    }

    \label{tab:stage abl}

    \vspace{-2mm}
\end{table}
\begin{table}[h!t]
    \renewcommand{\arraystretch}{1.2}
    \setlength{\tabcolsep}{1pt}
    \centering 
    \caption{Comparison of different components in RSS.} %
    \vspace{-2mm}
    
    \resizebox{0.8\linewidth}{!}{

    \begin{tabular}{c |cccc| ccc| ccc} 
        \hline
        \multicolumn{1}{l|}{\multirow{2}*{Methods}} 
        & \multicolumn{4}{c|}{Components} 
        & \multicolumn{3}{c|}{HRSOD-TE} 
        & \multicolumn{3}{c}{UHRSD-TE}\\

        \cline{2-11}
        \multicolumn{1}{l|}{} &   
         Inter&    Inner&  IGE&   PR&
        MAE&    \(F^{max}_{\beta}\)&  mBA&
        MAE&    \(F^{max}_{\beta}\)&  mBA\\

        \hline  
        Base&  
        -&  -&   -&  -&
        .0250&  .927&   .651&
        .0265&  .948&   .682\\
        
        1&  
        \checkmark&  \checkmark&   -&   -&
        .0198&  .942&   .724&
        .0205&  .957&   .769\\

        2&   
        \checkmark&  -&   \checkmark&   -&
        .0218&  .937&   .705&
        .0210&  .955&   .751\\

        3&   
        -&  \checkmark&   \checkmark&   -&
        .0201&  .942&   .720&
        .0206&  .955&   .766\\

        4&   
        \checkmark&  \checkmark&   \checkmark&  -&
        .0189&  .938&   .720&
        .0205&  .951&   .772\\

        5&   
        \checkmark&  \checkmark&   \checkmark&  \checkmark&
        .0191&  .940&   .736&
        .0187&  .956&   .784\\

        \hline 
    \end{tabular}
    }
    \label{tab:flow abl}
    \vspace{-4mm}
\end{table}

\textbf{Ablation of Components in RRS}. The RRS uses IGE, DGD and PR to obtain better SOD results. In Table \ref{tab:flow abl}, we remove some components of RSS to show which component contributes most. Comparing the result of methods 1-3, we can observe that the greatest performance drop happens when removing the inner-stage decoder, which suggests that most of the useful high-resolution information is from the object's boundaries. Besides, comparing the result of removing the  IGE, we can see that the performance drop is smaller. This suggests that the guidance to the decoder affects more on the final prediction results. Besides, we find that the mBA metric boosts when we add PR, which shows its strong ability for edge pixels' refinement.
\begin{table}[h!t]
    \renewcommand{\arraystretch}{1.2}
    \setlength{\tabcolsep}{1pt}
    \centering 
    \caption{Comparison of using PR at different scales.} %
    \vspace{-2mm}
    
    \resizebox{0.8\linewidth}{!}{

    \begin{tabular}{c |ccc| ccc| ccc} 
        \hline
        \multicolumn{1}{l|}{\multirow{2}*{Methods}} 
        & \multicolumn{3}{c|}{Scale} 
        & \multicolumn{3}{c|}{HRSOD-TE} 
        & \multicolumn{3}{c}{UHRSD-TE}\\

        \cline{2-10}
        \multicolumn{1}{l|}{} &   
         384&    768&  1536&
        MAE&    \(F^{max}_{\beta}\)&  mBA&
        MAE&    \(F^{max}_{\beta}\)&  mBA\\

        \hline  
        1&  
        \checkmark&  -&   -&
        .0227&  .941&  .731&
        .0204&  .957&  .780\\

        2&   
        -&  \checkmark&   -&
        .0203&  .937&   .738&
        .0193&  .955&   .782\\

        3&   
        -&  -&   \checkmark&
        .0199&  .936&   .736&
        .0190&  .955&   .786\\

        4&   
        -&  \checkmark&  \checkmark&
        .0191&  .940&   .736&
        .0187&  .956&   .784\\

        5&   
        \checkmark&  \checkmark&   \checkmark&
        .0214&  .939&   .737&
        .0188&  .956&   .788\\
        \hline 
    \end{tabular}
    }
    \vspace{-6mm}
    \label{tab:PR abl}
\end{table}

\textbf{Prediction Comparisons at Different Stages}. In Figure \ref{fig:visual ablation1}, we show some visual results at different stages. It can be observed that the coarse prediction can already find most of the salient region accurately, but lack of boundary details. With the participation of high-resolution information, the prediction from the two refinement stages restores more details.

\textbf{Ablation of PR}. PR refines the saliency map. In Table \ref{tab:PR abl}, we show the results of using PR at different scales. The results of methods 1-3 suggest that using PR at a higher resolution prediction will produce better results. However, adding more scales is not always helpful, as the methods 4-5 shows. This can be concluded that low-resolution PR will introduce additional disturbance in the latter PR. Thus, we take the method 4 as our final design.

\textbf{Visual effects of PR}. In Figure \ref{fig:visual ablation pr}, we show the predictions before and after PR. One can see that some details like hair strands can be restored in the refined prediction \(\textbf{P}_{r}\).
\section{Conclusion}
In this paper, we proposed a Recurrent Multi-scale Transformer for more precise HRSOD. It recurrently utilizes shared Transformers and multi-scale refinement architectures. Thus, high-resolution saliency maps can be generated with the guidance of lower-resolution predictions. Besides, we propose a new large-scale HRSOD dataset, which contains 10,500 high-quality annotated images. As far as we know, it is the largest dataset for the HRSOD task, which will significantly help future works in training and evaluating models. Extensive experiments show the usefulness of our dataset and superiority of the proposed framework. In the future, we will reduce the computation and improve the inference speed.
\begin{acks}
This work was supported in part by the National Natural Science Foundation of China (No. 62101092), the Open Project Program of State Key Laboratory of Virtual Reality Technology and Systems, Beihang University (No. VRLAB2022C02) and the Pujiang Program (No. 22PJ1406600).
\end{acks}

\newpage

\bibliographystyle{ACM-Reference-Format}
\balance
\bibliography{sample-base}

\newpage

\appendix
\nobalance
\section{Appendix}

\subsection{Subjects of Salient Objects}
\begin{figure}[h!t]
    \centering
    \includegraphics[width=1\linewidth]{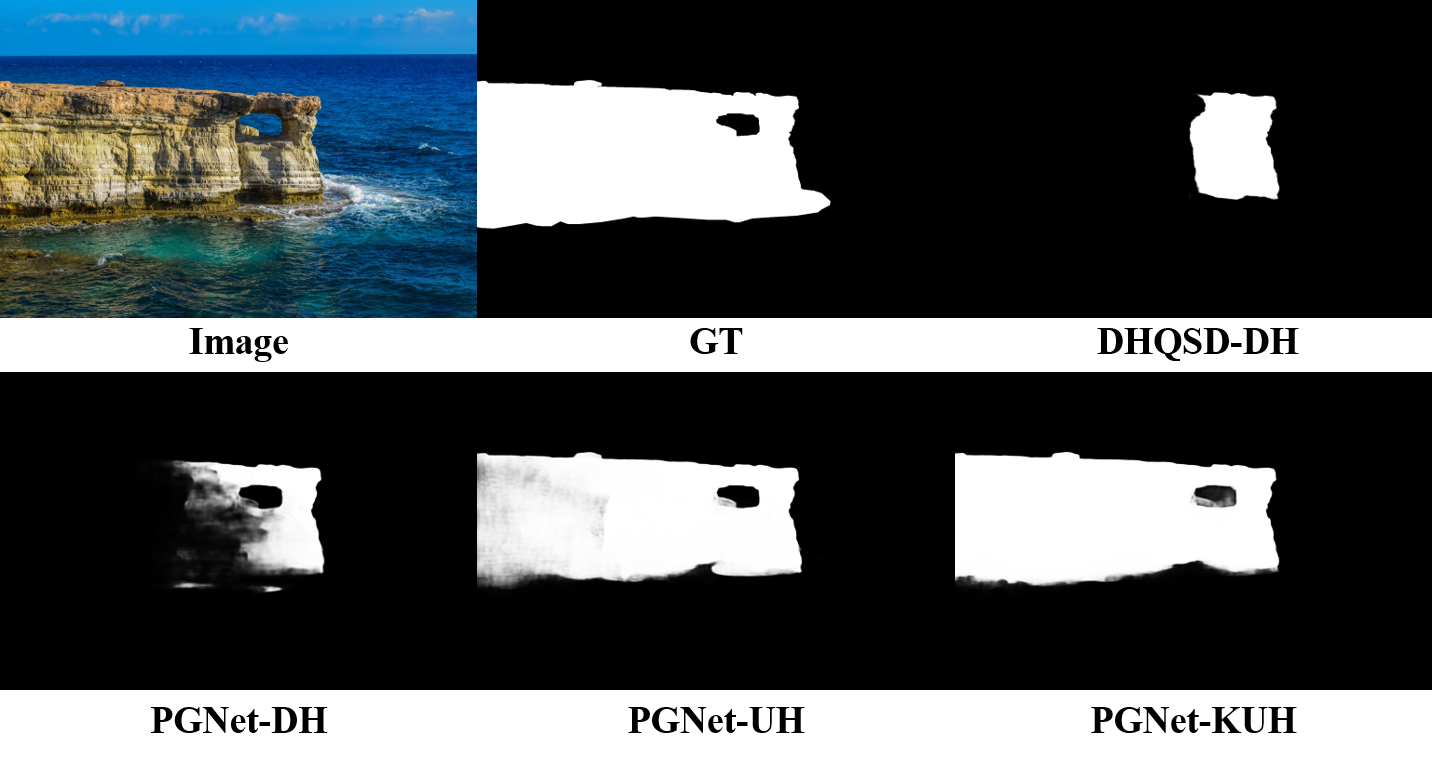}
    \caption{A typical example with rare subjects in UHRSD-TE. The PGNet trained with mixed HRSOD, UHRSD and HRS10K datasets (KUH setting) has the best prediction.}
    \label{fig:datadiff}
    \vspace{-2mm}
\end{figure}

As claimed in the main paper, there is a performance bias in the model training and evaluating when using different training settings.
Fig.~\ref{fig:datadiff} shows a typical example.
In fact, this is mainly due to the different subjects of the collected salient objects between different datasets.
During the collecting process for our dataset, we checked the subjects from previous LR SOD datasets,~\emph{e.g.}, DUTS~\cite{wang2017}, and HR SOD datasets,~\emph{e.g.}, UHRSD~\cite{xie2022pyramid} and HRSOD~\cite{zeng2019towards}.
We take their subjects as the reference and collect more salient subjects in our dataset, so as to keep the best semantic diversity.

\begin{figure}[h!t]
    \centering
    \includegraphics[width=1\linewidth]{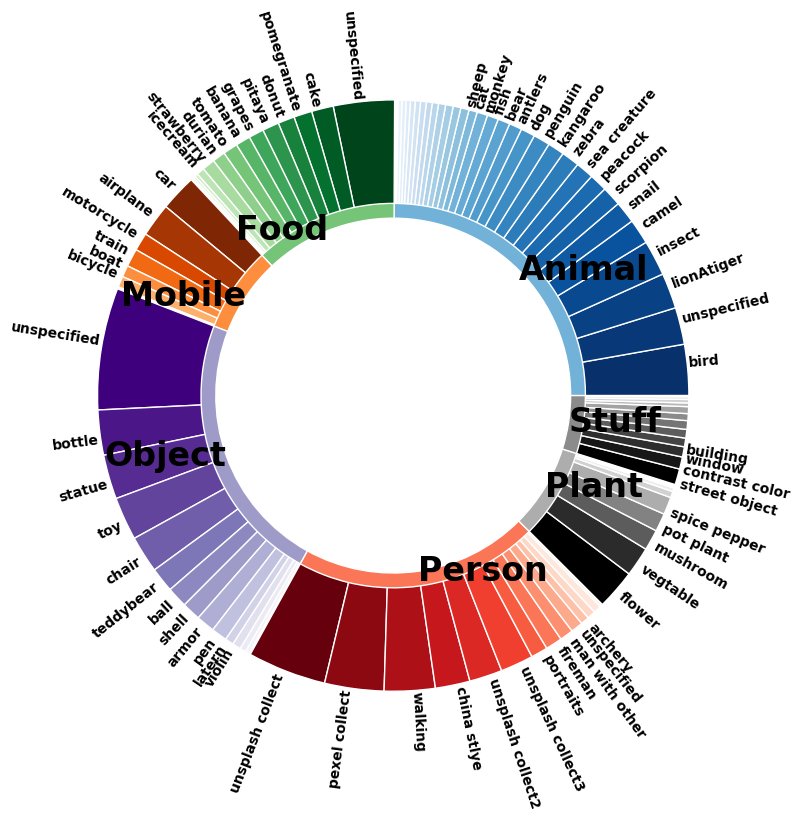}
    \caption{Visualization of different image subjects proportions in HRS10K dataset.}
    \label{fig:donut class}
    \vspace{-4mm}
\end{figure}

In fact, our proposed HRS10K dataset contains 10,500 images from 105 categories, dividing into 7 subjects.
Fig.~\ref{fig:donut class} illustrates the proportion of each subject. Specifically, these 7 subjects contains:

\textbf{Stuff}: A subject that extends rare salient categories in current SOD datasets, such as road, building, natural scene, etc.

\textbf{Plant}: A subject of salient objects which hardly appears in previous HR SOD datasets. We mainly collect plants with either attractive colors or complex shapes.

\textbf{Person}: A subject includes one or multiple human beings with diverse poses and clothes, in different situations.

\textbf{Object}: A common subject of salient objects, including different objects attracting peoples' attention. In this main subject, we also collect some categories,such as armor, teddy bear, etc., to increase the diversity of objects' types.

\textbf{Mobile}: A subject that includes man-made car, bicycle, motorbike, ship, etc. Note that salient objects in this subject usually hold clear object boundaries.

\textbf{Food}: A subject contains diverse food types and backgrounds. Highly contrastive textures may appear in the food, such as hot-dog, cake.

\textbf{Animal}: A subject includes lots of animals. Most of them have furry object boundaries. It is also the most common type of subjects in our dataset.
\subsection{More Quantitative Results on Other LR SOD Datasets}
Tab.~\ref{tab:supp lr result} shows the results on more LR SOD datasets (\emph{i.e.}, ECSSD and HKU-IS). Our method still shows better performances. These results is complementary to the corresponding quantitative results in the paper.

\begin{table}[ht]
    \renewcommand{\arraystretch}{1.5}
    \setlength{\tabcolsep}{3pt}
    \centering 
    \small
    \caption{Quantitative comparisons on more LR SOD datasets.} 
     \vspace{-2mm}
    \resizebox{1\linewidth}{!}{
    \begin{tabular}{l | cccc| cccc} 
        \hline
        \multicolumn{1}{l|}{\multirow{2}*{Method}}  & \multicolumn{4}{c|}{ECSSD}
                                                    & \multicolumn{4}{c}{HKU-IS}\\

        \cline{2-9}
        \multicolumn{1}{l|}{} & MAE&    \(F^{max}_{\beta}\)&   \(E_{\xi}\)&   \(S_{m}\)&
                                MAE&    \(F^{max}_{\beta}\)&   \(E_{\xi}\)&   \(S_{m}\)\\

        \hline  

       BASNet&
       .037&  .931&   .951&   .916&
       .032&  .919&   .951&   .908\\

       PoolNet&
       .042&  .927&   .946&   .915&
       .034&  .916&   .950&   .910\\

       EGNet&
       .039&  .931&   .951&   .921&
       .032&  .921&   .955&   .915\\

       SCRN&
       .037&  .931&   .956&   .927&
       .032&  .924&   .957&   .918\\

       F3Net&
       .033&  .935&   .954&   .924&
       .027&  .929&   .959&   .919\\

       MINet&
       .032&  .938&   .957&   .925&
       .028&  .925&   .960&   .918\\

       LDF&
       .034&  .937&   .953&   .924&
       .027&  .931&   .959&   .922\\

       GateNet&
       .037&  .936&   .955&   .924&
       .031&  .926&   .959&   .920\\

       PFSNet&
       .031&  .942&   .958&   .930&
       .025&  .934&   .962&   .924\\

       CTDNet&
       .036&  .933&   .951&   .917&
       .028&  .930&   .960&   .918\\

        \hline  

        DHQSD-DH&
        .031&  .900&   .919&   .894&
        .045&  .820&   .873&   .836\\

        PGNet-DH&
        .027&  .947&   .963&   .936&
        .023&  .939&   .966&   .930\\

        PGNet-UH&
        .032&  .936&   .958&   .926&
        .030&  .923&   .955&   .914\\

        PGNet-KUH&
        .026&  .950&   .967&   .939&
        .025&  .937&   .965&   .927\\

        \hline  

        Ours-UH&
        \textcolor{blue}{.021}&  \textcolor{blue}{.959}&   \textcolor{blue}{.974}&   \textcolor{blue}{.945}&
        \textcolor{blue}{.022}&  \textcolor{blue}{.945}&   \textcolor{blue}{.970}&   \textcolor{blue}{.933}\\

        Ours-KUH&
        \textcolor{red}{.020}&  \textcolor{red}{.962}&   \textcolor{red}{.976}&   \textcolor{red}{.948}&
        \textcolor{red}{.020}&  \textcolor{red}{.948}&   \textcolor{red}{.972}&   \textcolor{red}{.936}\\
        \hline
        
    \end{tabular}
    }
    \label{tab:supp lr result}
    \vspace{-2mm}
\end{table}
\subsection{Weights of the Loss Functions}
To clarify the effect of different loss weights, we have conducted additional experiments on the loss weights of different stages, i.e., CPS, RRS1 and RRS2. The results are shown in the Tab.~\ref{tab:loss weight ablation}. It can be observed that better results can be achieved by tuning this hyperparameter. However, there are no significant changes in the overall performance. Thus, for simplicity, we set equal weights for all stages.
\begin{table}[ht]
    \renewcommand{\arraystretch}{1.5}
    \setlength{\tabcolsep}{3pt}
    \centering 
    \small
    \caption{Comparisons with different loss weights.} 
     \vspace{-2mm}
    \resizebox{1\linewidth}{!}{
    \begin{tabular}{l | ccc| ccc| ccc} 
        \hline
        \multicolumn{1}{l|}{\multirow{2}*{Method}}  & \multicolumn{3}{c|}{Weight Config}
                                                    & \multicolumn{3}{c|}{HRSOD-TE}
                                                    & \multicolumn{3}{c}{UHRSD-TE}\\

        \cline{2-10}
        \multicolumn{1}{c|}{} & CPS&    RRS1&   RRS2&
                                MAE&    \(F^{max}_{\beta}\)&   mBA&
                                MAE&    \(F^{max}_{\beta}\)&   mBA\\

        \hline  

       1&
       0.1&  1&   1&
       .020&  .940&   .737&
       .020&  .954&   .787\\

       2&
       1&  0.1&   1&
       .020&  .938&   .738&
       .019&  .953&   .786\\

       3&
       1&  1&   0.1&
       .020&  .934&   .730&
       .020&  .951&   .777\\

       4&
       0.1&  0.5&   1&
       .017&  .934&   .730&
       .020&  .951&   .777\\

       5&
       0.5&  1&   2&
       .020&  .935&   .737&
       .019&  .957&   .787\\
        \hline
    \end{tabular}
    }
    \label{tab:loss weight ablation}
    \vspace{-2mm}
\end{table}

\subsection{Effectiveness of Edge Information}
To clarify the effectiveness of our method, we have conducted additional experiments without using edge information Ours-UH-wo/edge. The results are shown in the below Tab.~\ref{tab:no-edge results}. In this setting, we remove all edge prediction heads. For the proposed PR module, we use \(abs(pred - 0.5)\) as a replacement of edge prediction 
. It can be seen that, our method still shows better performances than PGNet, even without edge information. The result confirms the superiority of our RMFormer.

\begin{table}[ht]
    \renewcommand{\arraystretch}{1.5}
    \setlength{\tabcolsep}{3pt}
    \centering 
    \small
    \caption{Comparisons of using edge information.} 
     \vspace{-2mm}
    \resizebox{1\linewidth}{!}{
    \begin{tabular}{l | ccc| ccc| ccc} 
        \hline
        \multicolumn{1}{l|}{\multirow{2}*{Method}}  & \multicolumn{3}{c|}{HRSOD-TE}
                                                    & \multicolumn{3}{c|}{UHRSD-TE}
                                                    & \multicolumn{3}{c}{HRS10K-TE}\\

        \cline{2-10}
        \multicolumn{1}{c|}{} & MAE&    \(F^{max}_{\beta}\)&   mBA&
                                MAE&    \(F^{max}_{\beta}\)&   mBA&
                                MAE&    \(F^{max}_{\beta}\)&   mBA\\

        \hline  

       PGNet-DH&
       .020&  .931&   .726&
       .036&  .915&   .746&
       .042&  .902&   .721\\

       PGNet-UH&
       .020&  .938&   .721&
       .025&  .943&   .764&
       .034&  .923&   .728\\

       Ours-DH&
       .019&  .933&   .716&
       .027&  .938&   .744&
       .035&  .923&   .716\\

       Ours-UH&
       .019&  .939&   .763&
       .019&  .955&   .784&
       .028&  .936&   .745\\

       \hline

       Ous-UH-wo/edge	&
       .020&  .936&   .726&
       .020&  .953&   .773&
       .031&  .930&   .733\\
        \hline

    \end{tabular}
    }
    \label{tab:no-edge results}
    \vspace{-4mm}
\end{table}

\subsection{Failure Cases}
\begin{figure}[h!t]
    \centering
    \includegraphics[width=1\linewidth]{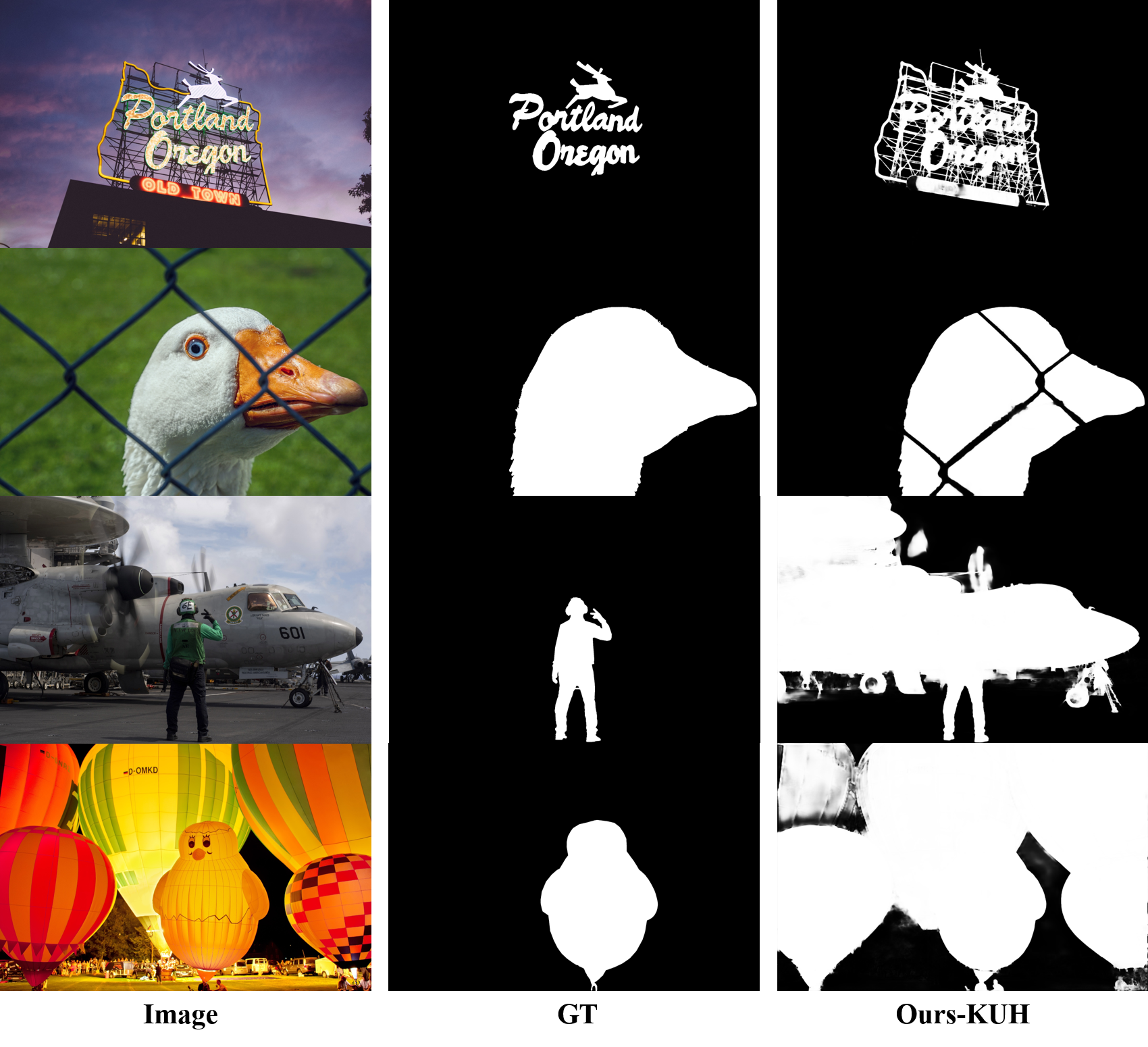}
    \vspace{-4mm}
    \caption{Failure cases with our proposed method.}
    \label{fig:fcase}
\end{figure}

Although our proposed method can produce saliency maps with better boundaries, there are still some failure cases that degenerate our method's performance.
As shown in Fig.~\ref{fig:fcase}, our method fails with a large salient region. 
We attribute this failure to two possible reasons:
\eject

\balance
First, as shown in the first and second rows of  Fig.~\ref{fig:fcase}, there are ambiguous annotations in UHRSD. This failure usually happens when high-resolution details are not correctly distinguish in the ground truth.
Then, as shown in the Fig.~\ref{fig:fcase}'s third and fourth rows, multiple salient objects may make our method hard to judge which one is the most salient object.
This failure occurs more often, especially when there are the same subject's salient objects in training images.
Thus, there is still a large room for the improvement of HR SOD tasks.

\subsection{Model Efficiency}
Tab.~\ref{tab:model eff} shows comparisons of computational efficiency between our method and other HR SOD methods.
Since our method uses a larger resolution image as input, our method holds a larger model size and more MACs.
However, we will address this limitation in our future work.
\begin{table}[h!t]
    \renewcommand{\arraystretch}{1.2}
    \setlength{\tabcolsep}{1pt}
    \centering 
    \caption{Comparisons of computational efficiency between our method and other methods.} 
    \resizebox{1\linewidth}{!}{
    \begin{tabular}{l |c| cc| cc} 
        \hline
        Methods&
        Ours&   PGNet&  DHQSD&  CTDNet&  PFSNet \\
        \hline  
        Resolution&
        1536&   1024&  1024&  352&  352 \\
        \hline
        Params (M)&
        174.19&     72&  -&  11.82&  15.57 \\
        \hline
        MACs (G)&
        563.14&     71&  27&  6.14M&  25.15M \\
        \hline
        Model Size (MB)&
        355.3&  285.8&  308.8&  96.6&  122.2 \\
        \hline 
        Speed (FPS)&
        16.1&   64.1&  8.1&  371.2&  70.4 \\
        \hline
    \end{tabular}
    }
    \label{tab:model eff}
\end{table}

\end{document}